\documentclass{article}
\usepackage{fullpage}
\usepackage{microtype}
\usepackage{graphicx}
\usepackage{subfigure}
\usepackage{booktabs} 

\usepackage{etoolbox}

\usepackage[utf8]{inputenc} 
\usepackage[T1]{fontenc}    
\usepackage{hyperref}       
\usepackage{url}            
\usepackage{booktabs}       
\usepackage{amsfonts}       
\usepackage{nicefrac}   
\usepackage[square,sort,comma,numbers]{natbib}
\usepackage{microtype}      
\usepackage{todonotes}
\usepackage{microtype,enumitem}
\usepackage{mathtools} 
\usepackage{amsmath}
\usepackage{bm}
\usepackage{algorithm2e}
\usepackage{algorithm}
\usepackage{amsthm}
\pdfoutput=1

\newtheorem{prop}{Proposition}
\newtheorem{thm}{Theorem}

\newtheorem{rmq}{Remark}

\newtheorem{prv}{Proof}

\newtheorem{lemma}{Lemma}
\newtheorem{definition}{Definition}

\newtheorem*{example*}{Example}
\newtheorem*{prop*}{Proposition}
\newtheorem*{thm*}{Theorem}
\newtheorem*{prv*}{Proof}

\newcommand{\MOT}{\textsc{EOT}\xspace}

\newcommand{\KL}{\textsc{KL}\xspace}
\newcommand{\TV}{\textsc{TV}\xspace}
\newcommand{\ent}{\textsc{H}\xspace}
\newcommand{\wass}{\textsc{W}\xspace}

\DeclareMathOperator*{\argminB}{argmin}

\usepackage{wrapfig, framed, caption}
\usepackage{multirow}

\title{Equitable and Optimal Transport with Multiple Agents}

\author{
	Meyer Scetbon$^{*1}$\qquad Laurent Meunier$^{*3,4}$ \qquad Jamal Atif$^{3}$ \qquad
	Marco Cuturi$^{1,2}$\\$^{1}$CREST, ENSAE\\
	$^{2}$ Google Brain\\
	$^{3}$LAMSADE, Université Paris-Dauphine\\
	$^{4}$ Facebook AI Research
	}
\date{}

\begin{document}

\maketitle

\begin{abstract}
We introduce an extension of the Optimal Transport problem when multiple costs are involved. Considering each cost as an agent, we aim to share equally between agents the work of transporting one distribution to another. To do so, we minimize the transportation cost of the agent who works the most. Another point of view is when the goal is to partition equitably goods between agents according to their heterogeneous preferences. Here we aim to maximize the utility of the least advantaged agent. This is a fair division problem. Like Optimal Transport, the problem can be cast as a linear optimization problem. When there is only one agent, we recover the Optimal Transport problem. When two agents are considered, we are able to recover Integral Probability Metrics defined by $\alpha$-Hölder functions, which include the widely-known Dudley metric. To the best of our knowledge, this is the first time a link is given between the Dudley metric and Optimal Transport. We provide an entropic regularization of that problem which leads to an alternative algorithm faster than the standard linear program. 
\end{abstract}

\section{Introduction}
Optimal Transport (OT) has gained interest last years in machine learning with diverse applications in neuroimaging~\citep{janati2020multi}, generative models~\citep{arjovsky2017wasserstein,salimans2018improving}, supervised learning~\citep{courty2016optimal}, word embeddings \citep{alvarez2018towards}, reconstruction cell trajectories \citep{yang2020predicting,schiebinger2019optimal} or adversarial examples~\citep{wong2019wasserstein}. The key to use OT in these applications lies in the gain of computation efficiency thanks to regularizations that smoothes the OT problem. More specifically, when one uses an entropic penalty, one recovers the so called Sinkhorn distances~\citep{cuturi2013sinkhorn}. In this paper, we introduce a new family of variational problems extending the optimal transport problem when multiple costs are involved with various applications in fair division of goods/work and operations research problems. 

Fair division~\citep{steinhaus1949fair} has been widely studied by the artificial intelligence~\citep{lattimore2015linear} and economics~\citep{moulin2004fair} communities. Fair division consists in partitioning diverse resources among agents according to some fairness criteria. One of the standard problems in fair division is the fair cake-cutting problem~\citep{dubins1961cut,brandt2016handbook}. The cake is an heterogeneous resource, such as a cake with different toppings, and the agents have heterogeneous preferences over different parts of the cake, i.e., some people prefer the chocolate toppings, some prefer the cherries, others just want a piece as large as possible. Hence, taking into account these preferences, one might share the cake equitably between the agents. A generalization of this problem, for which achieving fairness constraints is more challenging, is when the splitting involves several heterogeneous cakes, and where the agents have linked preferences over the different parts of the cakes. This problem has many variants such as the cake-cutting with two cakes \citep{cloutier2010two}, or the Multi Type Resource Allocation~\citep{mackin2015allocating,wang2019multi}. In all these models it is assumed that there is only one indivisible unit per type of resource available in each cake, and once an agent choose it, he or she has to take it all. In this setting, the cake can be seen as a set where each element of the set represents a type of resource, for instance each element of the cake represents a topping. A natural relaxation of these problems is when a divisible quantity of each type of resources is available.
We introduce $\MOT$ (\textbf{E}quitable and \textbf{O}ptimal \textbf{T}ransport), a formulation that solves
both the cake-cutting and the cake-cutting with two cakes problems in this setting. 

Our problem expresses as an optimal transportation problem. Hence, we prove duality results and provide fast computation based on Sinkhorn algorithm. As interesting properties, some Integral Probability Metrics ($\textsc{IPM}$s)~\citep{muller1997integral} as Dudley metric~\citep{dudley1966weak}, or standard Wasserstein metric~\citep{villani2003topics} are particular cases of the $\MOT$ problem.
\textbf{Contributions.} In this paper we introduce $\MOT$ an extension of Optimal Transport which aims at finding an equitable and optimal transportation strategy between multiple agents. We make the following contributions:
\begin{itemize}
    \item  In Section~\ref{sec:MOT}, we introduce the problem and show that it solves a fair division problem where heterogeneous resources have to be shared among multiple agents. We derive its dual and prove strong duality results. As a by-product, we show that $\MOT$ is related to some usual $\textsc{IPM}$s families and in particular the widely known Dudley metric.
    \item In Section~\ref{sec:entropic}, we propose an entropic regularized version of the problem, derive its dual formulation, obtain strong duality. We then provide an efficient algorithm to compute $\MOT$. Finally we propose other applications of $\MOT$ for Operations Research problems.
    
\end{itemize}

\section{Related Work}
\label{sec:background}
\paragraph{Optimal Transport.} Optimal transport aims to move a distribution towards another at lowest cost. More formally, if $c$ is a cost function on the ground space $\mathcal{X}\times\mathcal{Y}$, then the relaxed Kantorovich formulation of OT is defined for $\mu$ and $\nu$ two distributions as $$\wass_c(\mu,\nu) := \inf_{\gamma} \int_{\mathcal{X}\times\mathcal{Y}}c(x,y)d\gamma(x,y)$$
where the infimum is taken over all distributions $\gamma$ with marginals $\mu$ and $\nu$. Kantorovich theorem  states the following strong duality result under mild assumptions~\citep{villani2003topics}
$$\wass_c(\mu,\nu) = \sup_{f,g} \int_{\mathcal{X}}f(x)d\mu(x) + \int_{\mathcal{Y}}g(y)d\nu(y)$$
where the supremum is taken over continuous bounded functions satisfying for all $x,y$,  $f(x)+g(y)\leq c(x,y)$.
The question of considering an optimal transport problem when multiple costs are involved has already been raised in recent works. For instance,~\citep{paty2019subspace} proposed a robust Wasserstein distance where the distributions are projected on a $k$-dimensional subspace that maximizes their transport cost. In that sense, they aim to choose the most expensive cost among Mahalanobis square distances with kernels of rank $k$. In articles~\citep{li2019learning,sun2020learning}, the authors aim to learn a cost given observed matchings by inversing the optimal transport problem~\citep{dupuy2016estimating}. In~\citep{petrovich2020feature} the authors study ``feature-robust'' optimal transport, which can be also seen as a robust cost selection for optimal transport. In articles~\citep{genevay2017learning,scetbon2020linear}, the authors learn an adversarial cost to train a generative adversarial network. Here, we do not aim to consider a worst case scenario among the available costs but rather consider that the costs work together in order to split equitably the transportation problem among them at lowest cost.

\paragraph{Entropic relaxation of OT.} Computing exactly the optimal transport cost requires solving a linear program with a supercubic complexity $(n^3 \log n)$~\citep{Tarjan1997} that results in an output that is \textit{not} differentiable with respect to the measures' locations or weights~\citep{bertsimas1997introduction}. Moreover, OT suffers from the curse of dimensionality~\citep{dudley1969speed,fournier2015rate} and is therefore likely to be meaningless when used on samples from high-dimensional densities. Following the line of work introduced by~\cite{cuturi2013sinkhorn}, we propose an approximated computation of our problem by regularizing it with an entropic term. Such regularization in OT accelerates the computation, makes the problem differentiable with regards to the distributions~\citep{feydy2018interpolating} and reduces the curse of dimensionality~\citep{genevay2018sample}. Taking the dual of the approximation, we obtain a smooth and convex optimization problem under a simplicial constraint.

\paragraph{Fair Division.} Fair division of goods has a long standing history in economics and computational choice. A classical problem is the fair cake-cutting that consists in splitting the cake between $N$ individuals according to their heterogeneous preferences. The cake $\mathcal{X}$, viewed as a set, is divided in $\mathcal{X}_1,\dots,\mathcal{X}_N$ disjoint sets among the $N$ individuals. The utility for a single individual $i$ for a slice $S$ is denoted $V_i(S)$. It is often assumed that $V_i(\mathcal{X}) = 1$ and that $V_i$ is additive for disjoint sets. There exists many criteria to assess fairness for a partition $\mathcal{X}_1,\dots,\mathcal{X}_N$ such as proportionality ($V_i(\mathcal{X}_i)\geq 1/N$), envy-freeness ($V_i(\mathcal{X}_i)\geq V_i(\mathcal{X}_j)$) or equitability ($V_i(\mathcal{X}_i)= V_j(\mathcal{X}_j)$).
The cake-cutting problem has applications in many fields such as dividing land estates, advertisement space or broadcast time. An extension of the cake-cutting problem is the cake-cutting with two cakes problem~\citep{cloutier2010two} where two heterogeneous cakes are involved. In this problem, preferences of the agents can be coupled over the two cakes. The slice of one cake that an agent prefers might be influenced by the slice of the other cake that he or she might also obtain. The goal is to find a partition of the cakes that satisfies fairness conditions for the agents sharing the cakes.~\citet{cloutier2010two} studied the envy-freeness partitioning. Both the cake-cutting and the cake-cutting with two cakes problems assume that there is only one indivisible unit of supply per element $x\in\mathcal{X}$ of the cake(s). Therefore sharing the cake(s) consists in obtaining a paritition of the set(s). In this paper, we show that $\MOT$ is a relaxation of the cutting cake and the cake-cutting with two cakes problems, when there is a divisible amount of each element of the cake(s). In that case, cakes are no more sets but distributions that we aim to divide between the agents according to their coupled preferences.

\paragraph{Integral Probability Metrics. }
In our work, we make links with some integral probability metrics. IPMs are (semi-)metrics on the space of probability measures. For a set of functions $\mathcal{F}$ and two probability distributions $\mu$ and $\nu$, they are defined as $$\textsc{IPM}_\mathcal{F}(\mu,\nu)=\sup_{f\in\mathcal{F}}\int fd\mu-\int fd\nu.$$ For instance, when $\mathcal{F}$ is chosen to be the set of bounded functions with uniform norm less or equal than 1, we recover the Total Variation distance~\citep{steerneman1983total} (TV). They recently regained interest in the Machine Learning community thanks to their application to Generative Adversarial Networks (GANs)~\citep{goodfellow2014generative} where  $\textsc{IPM}$s are natural metrics  for  the discriminator~\citep{dziugaite2015training,arjovsky2017wasserstein,mroueh2017fisher,husain2019primal}. They also helped to build consistent two-sample tests~\citep{gretton2012kernel,scetbon2019comparing}. However when a closed form of the IPM is not available, exact computation of $\textsc{IPM}$s between discrete distributions may not be possible or can be costful. For instance, the Dudley metric can be written as a Linear Program~\citep{sriperumbudur2012empirical} which has at least the same complexity as standard OT. Here, we show that the Dudley metric is in fact a particular case of our problem and obtain a faster approximation thanks to the entropic regularization.

\begin{figure}[h!]
\centering
\includegraphics[width=0.3\linewidth]{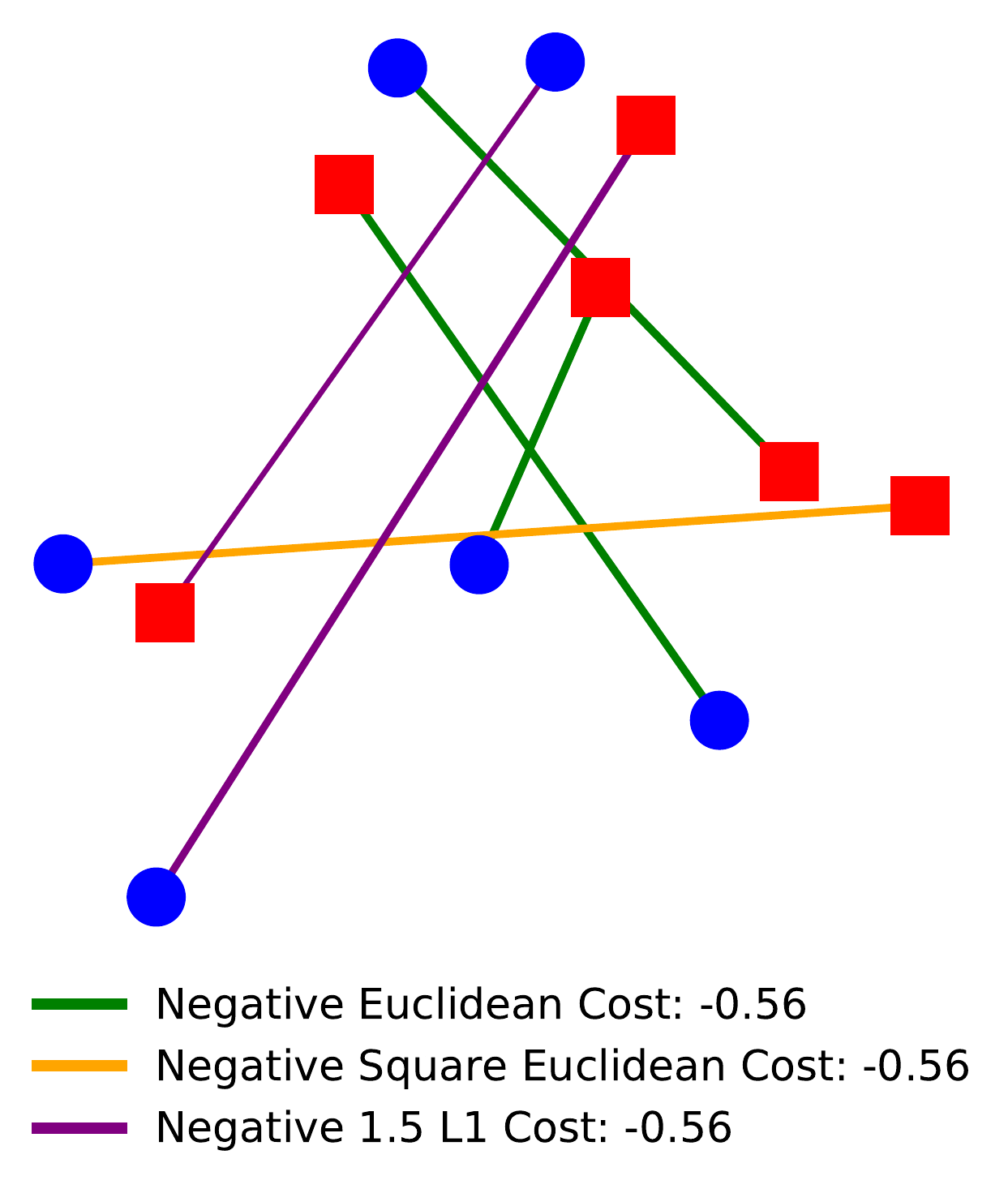}
\caption{Equitable and optimal division of the resources between $N=3$ different negative costs (i.e. utilities) given by $\MOT$. Utilities have been normalized. Blue dots and red squares represent the different elements of resources available in each cake. We consider the case where there is exactly one unit of supply per element in the cakes, which means that we consider uniform distributions. Note that the partition between the agents is equitable (i.e. utilities are equal) and proportional (i.e. utilities are larger than $1/N$).}
\label{fig:transport-map}
\end{figure}

\section{Equitable and Optimal Transport}
\label{sec:MOT}
\paragraph{Notations.} Let $\mathcal{Z}$ be a Polish space, we denote $\mathcal{M}(\mathcal{Z})$ the set of Radon measures on $\mathcal{Z}$. We call $\mathcal{M}_+(\mathcal{Z})$ the sets of positive Radon measures, and  $\mathcal{M}_+^1(\mathcal{Z})$ the set of probability measures. We denote $\mathcal{C}^b(\mathcal{Z})$ the vector space of bounded continuous functions on $\mathcal{Z}$. Let $\mathcal{X}$ and $\mathcal{Y}$ be two Polish spaces.  We denote for $\mu\in\mathcal{M}(\mathcal{X})$ and $\nu\in\mathcal{M}(\mathcal{Y})$, $\mu\otimes\nu$ the tensor product of the measures $\mu$ and $\nu$, and $\mu\ll\nu$ means that $\nu$ dominates $\mu$. We denote $\Pi_1:(x,y)\in\mathcal{X}\times\mathcal{Y}\mapsto x$ and $\Pi_2:(x,y)\in\mathcal{X}\times\mathcal{Y}\mapsto y$ respectively the projections on $\mathcal{X}$ and  $\mathcal{Y}$, which are continuous applications. For an application $g$ and a measure $\mu$, we denote $g_\sharp\mu$ the pushforward measure of $\mu$ by $g$.  For  $\mathcal{X}$ and $\mathcal{Y}$ two Polish spaces, we denote $\mathrm{LSC}(\mathcal{X}\times\mathcal{Y})$ the space of lower semi-continuous functions on $\mathcal{X}\times\mathcal{Y}$,  $\mathrm{LSC}^+(\mathcal{X}\times\mathcal{Y})$ the space of non-negative lower semi-continuous functions on $\mathcal{X}\times\mathcal{Y}$ and $\mathrm{LSC}^-_{*}(\mathcal{X}\times\mathcal{Y})$ the set of negative bounded below lower semi-continuous functions on $\mathcal{X}\times\mathcal{Y}$ . We also denote $\mathrm{C}^+(\mathcal{X}\times\mathcal{Y})$ the space of non-negative continuous functions on $\mathcal{X}\times\mathcal{Y}$ and $\mathrm{C}^-_{*}(\mathcal{X}\times\mathcal{Y})$ the set of negative continuous functions on $\mathcal{X}\times\mathcal{Y}$. Let $N\geq 1$ be an integer and denote $\Delta_N^{+}: = \{\lambda\in\mathbb{R}_+^N~\mathrm{s.t.}~\sum_{i=1}^N\lambda_i=1\}$, the probability simplex of $\mathbb{R}^N$.
For two positive measures of same mass $\mu\in\mathcal{M}_+(\mathcal{X})$ and $\nu\in\mathcal{M}_+(\mathcal{Y})$, we define the set of couplings with marginals $\mu$ and $\nu$:
\begin{align*}
    \Pi_{\mu,\nu}:=\left\{\gamma~\mathrm{s.t.}~ \Pi_{1\sharp}\gamma=\mu ~,~ \Pi_{2\sharp}\gamma=\nu\right\}\; .
\end{align*}
We introduce the subset of $(\mathcal{M}_+^1(\mathcal{X})\times \mathcal{M}_+^1(\mathcal{Y}))^N$ representing marginal decomposition: 
\begin{align*}
 \textstyle\Upsilon_{\mu,\nu}^N:=&\Big\{(\mu_i,\nu_i)_{i=1}^N ~\mathrm{ s.t. }~ \sum_i \mu_i = \mu \mathrm{, } \sum_i \nu_i = \nu \\ 
 &~\mathrm{ and }~\forall i,~ \mu_i(\mathcal{X}) = \nu_i(\mathcal{Y})\;   \Big\}.
\end{align*}
We also define the following subset of $\mathcal{M}_+(\mathcal{X}\times \mathcal{Y})^N$ corresponding to the coupling decomposition:
\begin{align*}
    \Gamma^N_{\mu,\nu}:=\left\{(\gamma_i)_{i=1}^N~\mathrm{s.t.}~ \Pi_{1\sharp}\sum\gamma_i=\mu ~,~ \Pi_{2\sharp}\sum\gamma_i=\nu\right\} .
\end{align*}

\subsection{Primal Formulation}
\label{sec:primal-dual}
Consider a fair division problem where several agents aim to share two sets of resources, $\mathcal{X}$ and $\mathcal{Y}$, and assume that there is a divisible amount of each resource $x\in\mathcal{X}$ (resp. $y\in\mathcal{Y}$) that is available. Formally, we consider the case where resources are no more sets but rather distributions on these sets. Denote $\mu$ and $\nu$ the distribution of resources on respectively $\mathcal{X}$ and $\mathcal{Y}$. For example, one might think about a situation where agents want to share fruit juices and ice creams and there is a certain volume of each type of fruit juices and a certain mass of each type of ice creams available. Moreover each agent defines his or her paired preferences for each couple $(x,y)\in\mathcal{X}\times\mathcal{Y}$. Formally, each person $i$ is associated to an upper semi-continuous mapping  $u_i:\mathcal{X}\times\mathcal{Y}\xrightarrow{} \mathbb{R}^{+}$ corresponding to his or her preference for any given pair $(x,y)$. For example, one may prefer to eat chocolate ice cream with apple juice, but may prefer pineapple juice when it comes with vanilla ice cream. The total utility for an individual $i$ and a pairing $\gamma_i\in\mathcal{M}_+(\mathcal{X}\times\mathcal{Y})$ is then given by $V_i(\gamma_i) := \int u_i d\gamma_i$. To partition fairly among individuals, we maximize the minimum of individual utilities.

From a transport point of view, let assume that there are $N$ workers available to transport a distribution $\mu$ to another one $\nu$. The cost of a worker $i$ to transport a unit mass from location $x$ to the location $y$ is $c_i(x,y)$. To partition the work among the $N$ workers fairly, we minimize the maximum of individual costs. 

These problems are in fact the same where the utility $u_i$, defined in the fair division problem, might be interpreted as the opposite of the cost $c_i$ defined in the transportation problem, i.e. for all $i$, $c_i = -u_i$. The two above problem motivate the introduction of $\MOT$ defined as follows.
\begin{definition}[Equitable and Optimal Transport]
\label{def-mot}
Let $\mathcal{X}$ and $\mathcal{Y}$ be Polish spaces. Let $\mathbf{c}:=(c_i)_{1\leq i\leq N}$ be a family of bounded below lower semi-continuous cost functions on $\mathcal{X}\times \mathcal{Y}$, and $\mu\in\mathcal{M}^1_+(\mathcal{X})$ and  $\nu\in\mathcal{M}^1_+(\mathcal{Y})$. We define the equitable and optimal transport primal problem:
\begin{align}
\label{eq-primal}
\MOT_{\mathbf{c}}(\mu,\nu) := \inf_{\substack{(\gamma_i)_{i=1}^N\in\Gamma^N_{\mu,\nu}\
}}   \max_i \int c_i d\gamma_i\; .
\end{align}
\end{definition}

We prove along with Theorem~\ref{thm:duality-GOT} that the problem is well defined and the infimum is attained. Lower-semi continuity is a standard assumption in OT. In fact, it is the weakest condition to prove Kantorovich duality~\citep[Chap. 1]{villani2003topics}. Note that the problem defined here is a linear optimization problem and when $N=1$ we recover standard optimal transport. Figure~\ref{fig:transport-map} illustrates the equitable and optimal transport problem we consider. Figure~\ref{fig:transport-map-ot-view} in Appendix~\ref{appendix-illustrations} shows an illustration with respect to the transport viewpoint in the exact same setting, i.e. $c_i = - u_i$. As expected, the couplings obtained in the two situations are not the same. 

We now show that in fact, $\MOT$ optimum satisfies equality constraints in case of constant sign costs, i.e. total utility/cost of each individual are equal in the optimal partition. See Appendix~\ref{prv:mot-equality} for the proof.
\begin{prop}[$\MOT$ solves the problem under equality constraints]
\label{prop:mot-equality}
Let $\mathcal{X}$ and $\mathcal{Y}$ be Polish spaces. Let $\mathbf{c}:=(c_i)_{1\leq i\leq N}\in\mathrm{LSC}^+(\mathcal{X}\times\mathcal{Y})^N\cup\mathrm{LSC}^-_{*}(\mathcal{X}\times\mathcal{Y})^N$, $\mu\in\mathcal{M}^1_+(\mathcal{X})$ and  $\nu\in\mathcal{M}^1_+(\mathcal{Y})$. Then the following are equivalent:
\vspace{-0.3cm}
\begin{itemize}
    \item $(\gamma_i^{*})_{i=1}^N\in\Gamma^N_{\mu,\nu}$ is solution of Eq.~(\ref{eq-primal}),
    \item $(\gamma_i^{*})_{i=1}^N\in \argminB\limits_{\substack{(\gamma_i)_{i=1}^N\in\Gamma^N_{\mu,\nu}
}}   \left\{t~\mathrm{s.t.}~ \forall i~\int c_i d\gamma_i=t\right\}$.
\end{itemize}
\vspace{-0.4cm}
Moreover,
\vspace{-0.3cm}
\begin{align*}
  \MOT_{\mathbf{c}}(\mu,\nu) =  \min_{\substack{(\gamma_i)_{i=1}^N\in\Gamma^N_{\mu,\nu}
}}   \left\{t~\mathrm{s.t.}~ \forall i~\int c_i d\gamma_i=t\right\}\; .
\end{align*}
\end{prop}

This property highly relies on the sign of the costs. For instance if two costs are considered, one always positive and the other always negative, then the constraints cannot be satisfied. When the cost functions are non-negatives, $\MOT$ refers to a transportation problem while when the costs are all negatives, costs become utilities and $\MOT$ refers to a fair division problem. The two points of view are concordant, but proofs and interpretations rely on the sign of the costs.

\subsection{An Equitable and Proportional Division } 
When the cost functions considered $c_i$ are all negatives, $\MOT$ become a fair division problem where the utility functions are defined as $u_i:=-c_i$. Indeed according to Proposition~\ref{prop:mot-equality}, $\MOT$ solves
\begin{align*}
\max_{\substack{(\gamma_i)_{i=1}^N\in\Gamma^N_{\mu,\nu}
}}   \left\{t~\mathrm{s.t.}~ \forall i,~\int u_i d\gamma_i=t\right\}\; .
\end{align*}
Recall that in our model, the total utility of the agent $i$ is given by $V_i(\gamma_i):=\int u_i d\gamma_i$. Therefore $\MOT$ aims to maximize the total utility of each agent $i$ while ensuring that they are all equal. Let us now analyze which fairness conditions the partition induced by $\MOT$ verifies. Assume that the utilities are normalized, i.e., $\forall i$, there exists $\gamma_i\in\mathcal{M}_+^1(\mathcal{X}\times\mathcal{Y})$ such that $V_i(\gamma_i)=1$. For example one might consider the cases where $\forall i$,  $\gamma_i=\mu\otimes\nu$ or $\gamma_i\in\argminB_{\gamma\in\Pi_{\mu,\nu}}  \int c_i d\gamma$. Then any solution $(\gamma_i^{*})_{i=1}^N\in\Gamma^N_{\mu,\nu}$ of $\MOT$ satisfies:
\begin{itemize}
    \item \textbf{Proportionality}: for all $i$, $V_i(\gamma_i^{*})\geq 1/N$,
    \item \textbf{Equitablity}: for all $i,j$, $V_i(\gamma_i^{*})=V_j(\gamma_j^{*})$.

\end{itemize}
Proportionality is a standard fair division criterion for which a resource is divided among $N$ agents, giving each agent at least $1/N$ of the heterogeneous resource by his/her own subjective valuation. Therefore here, this situation corresponds to the case where the normalized utility of each agent is at least $1/N$. Moreover, an equitable division is a division of an heterogeneous resource, in which each partner is equally happy with his/her share. Here this corresponds to the case where the utility of each agent are all equal.

The problem solved by $\MOT$ is a fair division problem where heterogeneous resources have to be shared among multiple agents according to their preferences. This problem is a relaxation of the two cake-cutting problem when there are a divisible amount of each item of the cakes. In that case, cakes are distributions and $\MOT$ makes a proportional and equitable partition of them. Details are left in Appendix~\ref{prv:mot-equality}.

\paragraph{Fair Cake-cutting.} Consider the case where the cake is an heterogeneous resource and there is a certain divisible quantity of each type of resource available. For example chocolate and vanilla are two types of resource present in the cake for which a certain mass is available. In that case, each type of resource in the cake is pondered by the actual quantity present in the cake. Up to a normalization, the cake is no more the set $\mathcal{X}$ but rather a distribution on this set. Note that for the two points of view to coincide, it suffices to assume that there is exactly the same amount of mass for each type of resources available in the cake. In that case, the cake can be represented by the uniform distribution over the set $\mathcal{X}$, or equivalently the set $\mathcal{X}$ itself. When cakes are distributions, the fair cutting cake problem can be interpreted as a particular case of $\MOT$ when the utilities of the agents do not depend on the variable $y\in\mathcal{Y}$. In short, we consider that utilities are functions of the form $u_i(x,y)=v_i(x)$ for all $(x,y)\in\mathcal{X}\times\mathcal{Y}$. The normalization of utilities can be cast as follows: $\forall i$, $V_i(\mu) = \int v_i(x) d\mu(x) = 1$. Then Proposition~\ref{prop:mot-equality} shows that the partition of the cake made by $\MOT$ is proportional and equitable. Note that for $\MOT$ to coincide with the classical cake-cutting problem, one needs to consider that the uniform masses of the cake associated to each type of resource cannot be splitted. This can be interpreted as a Monge formulation~\citep{villani2003topics} of $\MOT$ which is out of the scope of this paper. 

\subsection{Optimality of $\MOT$} 
We next investigate the coupling obtained by solving $\MOT$. In the next proposition, we show that under the same assumptions of Proposition~\ref{prop:mot-equality}, $\MOT$ solutions are optimal transportation plans. See Appendix~\ref{prv:mot-otplans} for the proof.
\begin{prop}[$\MOT$ realizes optimal plans]
\label{prop:mot-otplans}
Under the same conditions of Proposition~\ref{prop:mot-equality}, for any $(\gamma_i^{*})_{i=1}^N\in\Gamma^N_{\mu,\nu}$ solution of Eq.~(\ref{eq-primal}), we have for all $i\in\{1,\dots,N\}$
\begin{equation}
  \begin{aligned}
  \label{eq-optimal-coupling}
    \gamma_i^{*}&\in\argminB_{\gamma\in\Pi_{\mu_i^{*},\nu_i^{*}}}  \int c_i d\gamma\\
    \text{where\quad} & \mu_i^{*}:=\Pi_{1\sharp}\gamma_i^{*} ~,~ \nu_i^{*}:=\Pi_{2\sharp}\gamma_i^{*}\; ,
\end{aligned}
\end{equation}
and
\begin{equation}
\begin{aligned}
\label{eq-mot-optimality}
  \MOT_{\mathbf{c}}(\mu,\nu)=  &\min_{\substack{(\mu_i,\nu_i)_{i=1}^N\in\Upsilon^N_{\mu,\nu}}} t\\
  &\mathrm{s.t.}~ \forall i~ \wass_{c_i}(\mu_i,\nu_i) =t \; .
\end{aligned}
\end{equation}
\end{prop}

Given the optimal matchings $(\gamma_i^{*})_{i=1}^N\in\Gamma^N_{\mu,\nu}$, one can easily obtain the partition of the agents of each marginals. Indeed for all $i$, $\mu_i^{*}:=\Pi_{1\sharp}\gamma_i^{*}$ and $ \nu_i^{*}:=\Pi_{2\sharp}\gamma_i^{*}$ represent respectively the portion of the agent $i$ from distributions $\mu$ and $\nu$.

\begin{rmq}[Utilitarian and Optimal Transport]
To contrast with $\MOT$, an alternative problem is to maximize the sum of the total utilities of agents, or equivalently minimize the sum of the total costs of agents. This problem can be cast as follows:
\begin{align}
\label{eq-UOT}
  \inf_{\substack{(\gamma_i)_{i=1}^N\in\Gamma^N_{\mu,\nu}\
}}  \sum_i \int c_i d\gamma_i
\end{align}
Here one aims to maximize the total utility of all the agents, while in $\MOT$ we aim to maximize the total utility per agent under egalitarian constraint. The solution of~(\ref{eq-UOT}) is not fair among agents and one can show that this problem is actually equal to $\wass_{\min_i (c_i)}(\mu,\nu)$. Details can be found in Appendix~\ref{res:min-sum}.
\end{rmq}

\begin{figure*}[t]
\begin{tabular}{@{}c@{}c@{}c@{}c@{}}
\includegraphics[width=0.24\textwidth]{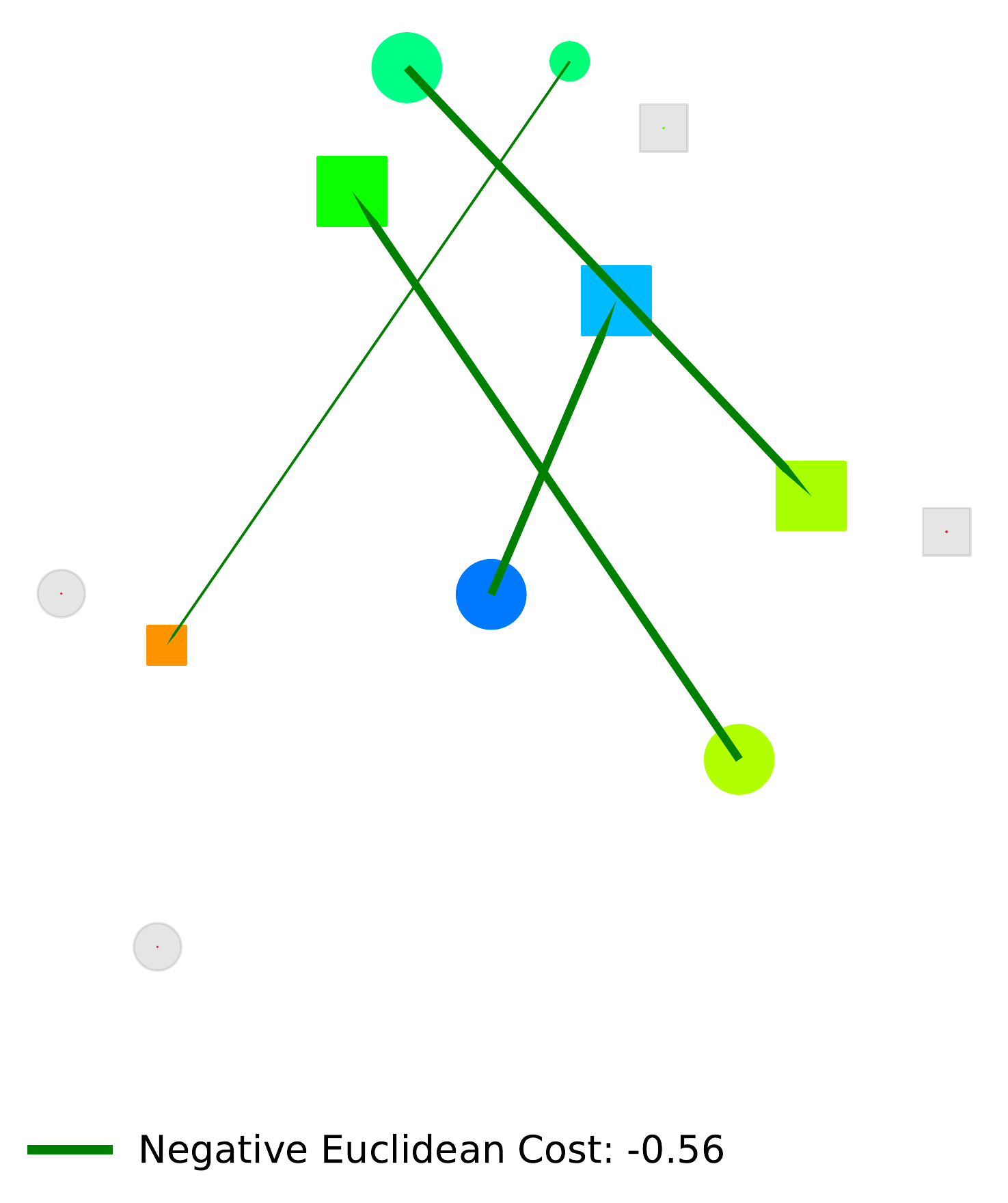}&
\includegraphics[width=0.24\textwidth]{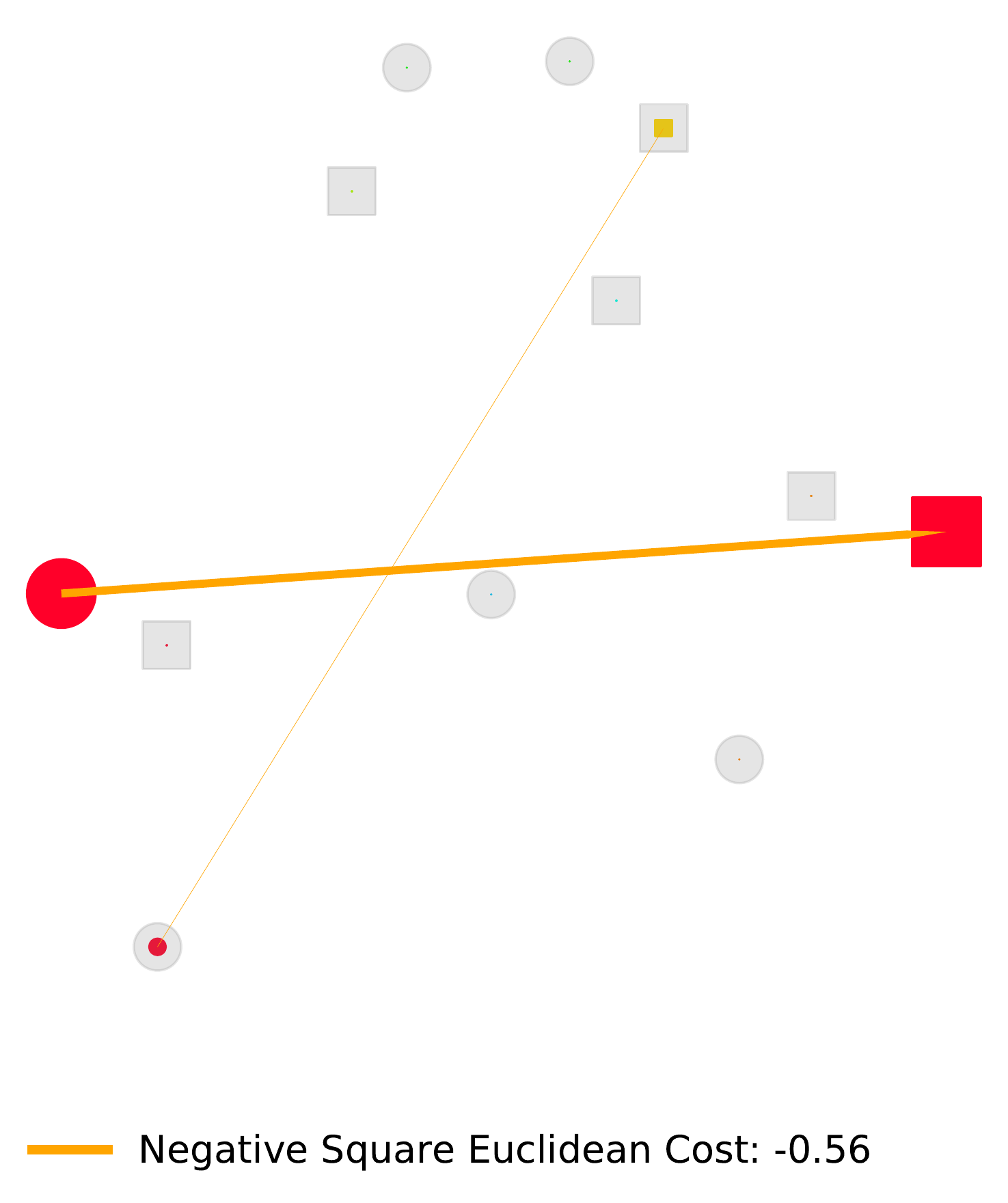}&
\includegraphics[width=0.24\textwidth]{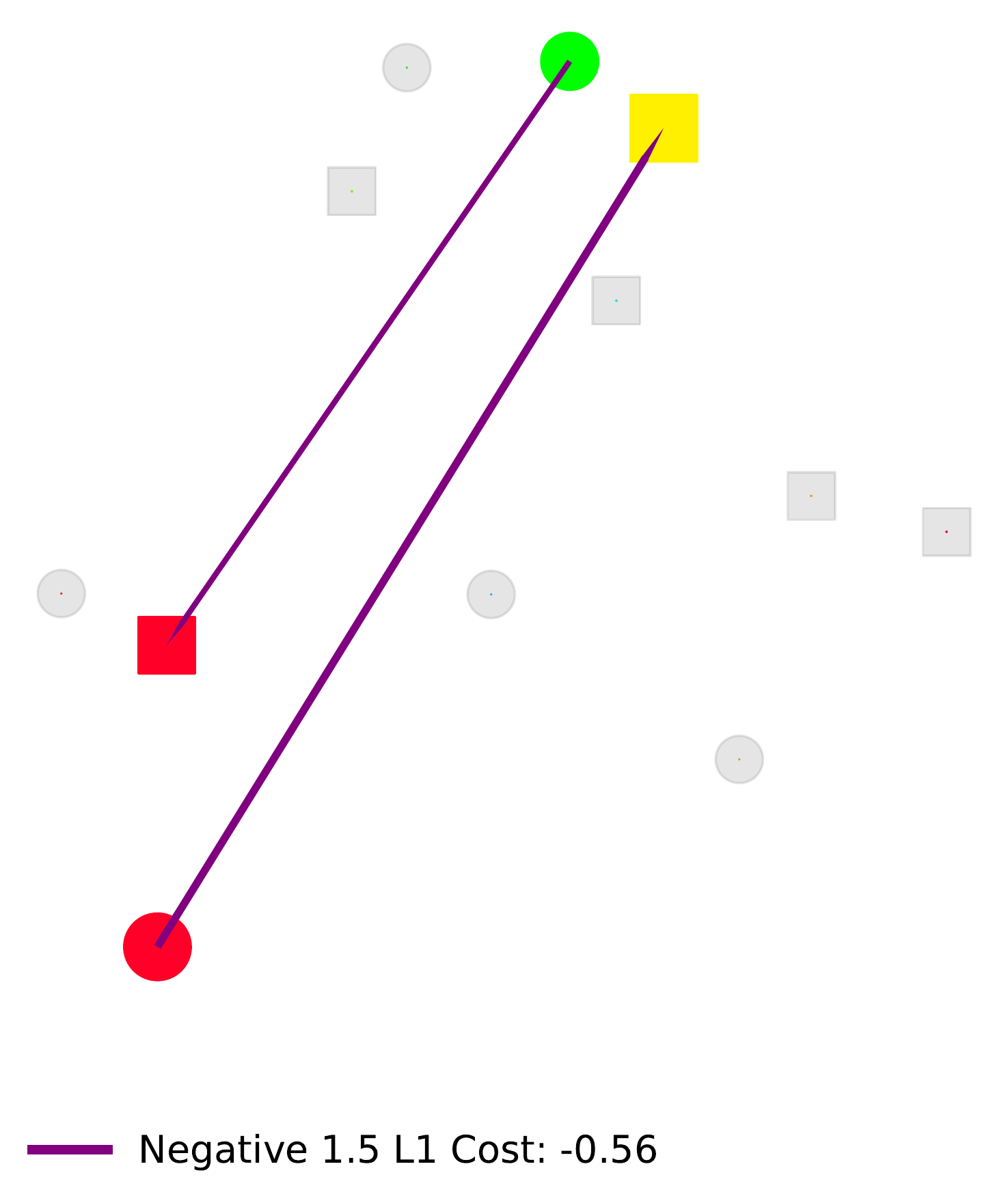}&
\includegraphics[width=0.275\textwidth]{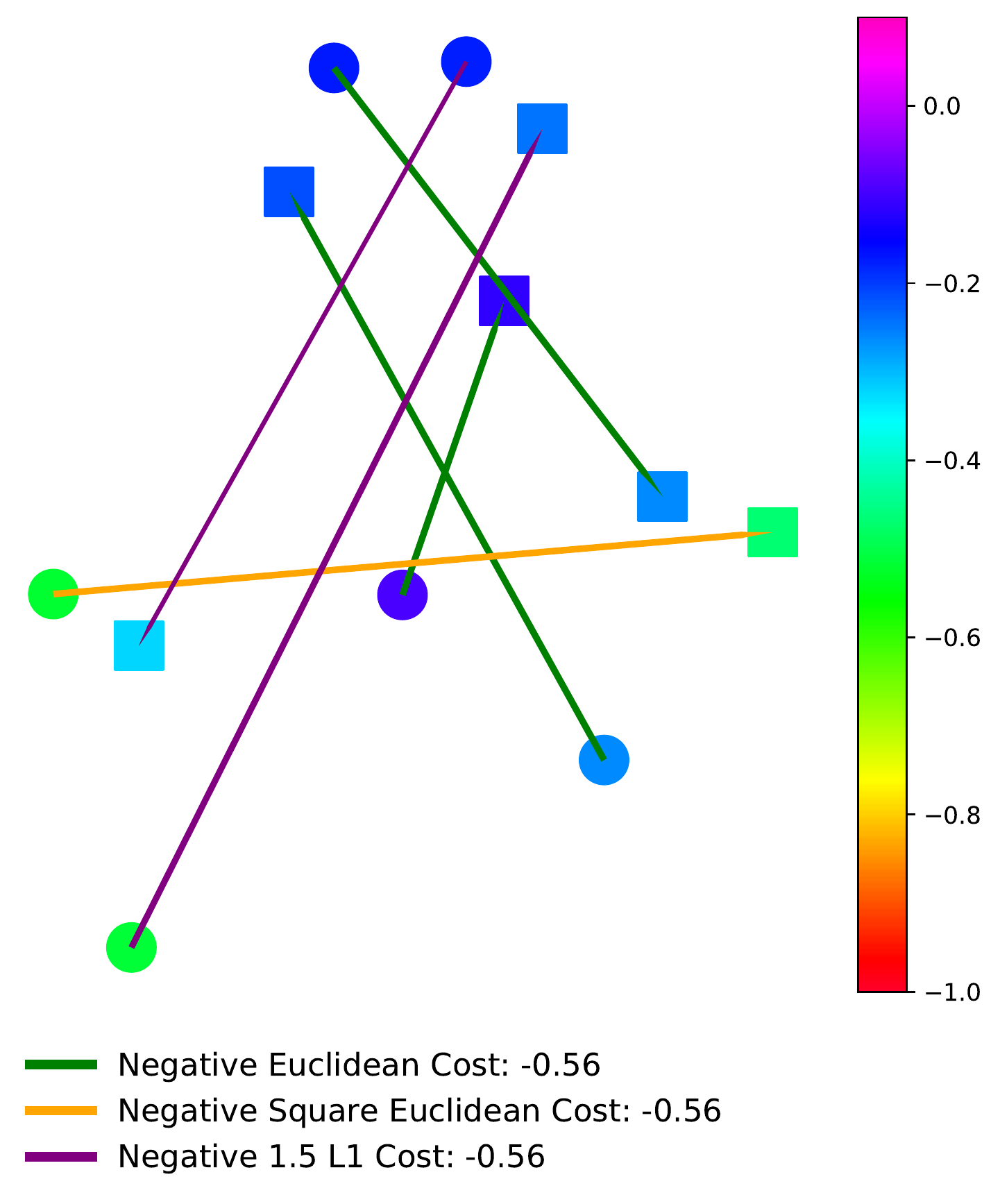}
\end{tabular}
\caption{\emph{Left, middle left, middle right}: the size of dots and squares is proportional to the weight of their representing atom in the distributions $\mu_k^{*}$ and $\nu_k^{*}$ respectively. 
The utilities $f_k^{*}$ and $g_k^{*}$ for each point in respectively $\mu_k^{*}$ and $\nu_k^{*}$ are represented by the color of dots and squares according to the color scale on the right hand side. The gray dots and squares correspond to the points that are ignored by agent $k$ in the sense that there is no mass or almost no mass in distributions $\mu^*_k$ or $\nu^*_k$. \emph{Right}: the size of dots and squares are uniform since they correspond to the weights of uniform distributions $\mu$ and $\nu$ respectively. The values of $f^*$ and $g^*$ are given also by the color at each point. Note that each agent gets exactly the same total utility, corresponding exactly to $\MOT$. This value can be computed using dual formulation~\eqref{eq-dual} and for each figure it equals the sum of the values (encoded with colors) multiplied by the weight of each point (encoded with sizes).\label{fig:potential-dual}}
\end{figure*}

\subsection{Dual Formulation}

Let us now introduce the dual formulation of the problem and show that strong duality holds under some mild assumptions. See Appendix~\ref{prv:duality-GOT} for the proof.
\begin{thm}[Strong Duality] 
\label{thm:duality-GOT}
Let $\mathcal{X}$ and $\mathcal{Y}$ be Polish spaces. Let $\mathbf{c}:=(c_i)_{i=1}^{N}$ be bounded below lower semi-continuous costs. Then \emph{strong duality holds}, i.e. for $(\mu,\nu)\in\mathcal{M}_{+}^{1}(\mathcal{X})\times\mathcal{M}_{+}^{1}(\mathcal{Y})$:
\begin{align}
\label{eq-dual}
   \MOT_{\mathbf{c}}(\mu,\nu)= \sup\limits_{\substack{\lambda\in\Delta^+_N\\(f,g)\in\mathcal{F}_{\mathbf{c}}^{\lambda}}}\int fd\mu+ \int gd\nu
\end{align}
where  $\mathcal{F}^{\lambda}_{\mathbf{c}}:=\{(f,g)\in\mathcal{C}^b(\mathcal{X})\times\mathcal{C}^b(\mathcal{Y})~\mathrm{ s.t. }~\forall i\in\{1,...,N\},~ f\oplus g\leq\lambda_ic_i\}$.
\end{thm}

This theorem holds under the same hypothesis and follows the same reasoning as the one in~\citep[Theorem 1.3]{villani2003topics}. While the primal formulation of the problem is easy to understand, we want to analyse situations where the dual variables also play a role. For that purpose we
show in the next proposition a simple characterisation
of the primal-dual optimality in case of constant sign
cost functions.  See Appendix~\ref{prv:optimality-cond} for the proof.
\begin{prop}
\label{prop:optimality-cond}
Let $\mathcal{X}$ and $\mathcal{Y}$ be compact Polish spaces. Let $\mathbf{c}:=(c_i)_{1\leq i\leq N}\in\mathrm{C}^+(\mathcal{X}\times\mathcal{Y})^N\cup\mathrm{C}^-_{*}(\mathcal{X}\times\mathcal{Y})^N$, $\mu\in\mathcal{M}^1_+(\mathcal{X})$ and  $\nu\in\mathcal{M}^1_+(\mathcal{Y})$. Let also $(\gamma_k)_{k=1}^N\in\Gamma^N_{\mu,\nu}$ and $(\lambda,f,g)\in\Delta_n^{+}\times\mathcal{C}^b(\mathcal{X})\times\mathcal{C}^b(\mathcal{Y})$. Then Eq.~(\ref{eq-dual}) admits a solution and the following are equivalent:
\begin{itemize}
    \item $(\gamma_k)_{k=1}^N$ is a solution of Eq.~(\ref{eq-primal}) and $(\lambda,f,g)$ is a solution of Eq.~(\ref{eq-dual}).
    \item \begin{enumerate}
        \item $\forall i\in\{1,...,N\},~ f\oplus g\leq\lambda_i c_i$
        \item $\forall i,j\in\{1,...,N\}~\int c_i d\gamma_i=\int c_j d\gamma_j$
        \item $f \oplus g= \lambda_i c_i ~~\gamma_{i}\text{-a.e.}$
    \end{enumerate}
\end{itemize}
\end{prop}
\begin{rmq}
\label{rk:lambdanonzero}
It is worth noting that when we assume that $\mathbf{c}:=(c_i)_{1\leq i\leq N}\in\mathrm{C}^+_{*}(\mathcal{X}\times\mathcal{Y})^N\cup\mathrm{C}^-_{*}(\mathcal{X}\times\mathcal{Y})^N$, then we can refine the second point of the equivalence presented in Proposition~\ref{prop:optimality-cond} by adding the following condition: $\forall i\in\{1,...,N\}~\lambda_i\neq 0$.
\end{rmq}
Given two distributions of resources represented by the measures $\mu$ and $\nu$, and $N$ utility functions denoted $(u_i)_{i=1}^N$, we want to find an \emph{equitable} and \emph{stable} partition among the agents in case of \emph{transferable utilities}. Let $k$ be an agent. We say that his or her utility is transferable when once $x\in\mathcal{X}$ and $y\in\mathcal{Y}$ get matched, he or she has to decide how to split his or her associated utility $u_k(x,y)$ . She or he divides $u_k(x,y)$ into a quantity $f_k(x)$ which can be seen as the utility of having $x$ and $g_k(y)$ for having $y$. Therefore in that problem we ask for $(\gamma_k,f_k,g_k)_{k=1}^N$ such that
\begin{align}
\label{eq-1-dual}
    u_k(x,y)=f_k(x)+g_k(y) ~~\gamma_{k}\text{-a.e.} 
\end{align}
Moreover, for the partition to be \emph{stable}~\citep{rothm}, we want to ensure that, for every agent $k$, none of the resources $x\in\mathcal{X}$ and $y\in\mathcal{Y}$ that have not been matched together for this agent would increase their utilities, $f_k(x)$ and $g_k(y)$, if there were matched together in the current matching instead. Formally we ask that for $k\in\{1,\dots,N\}$ and all $(x,y)\in\mathcal{X}\times\mathcal{Y}$,
\begin{align}
\label{eq-2-dual}
    f_k(x)+g_k(y)\geq u_k(x,y)\; .
\end{align}
Indeed if there exist $k$, $x$ and $y$ such that $u_k(x,y)>f_k(x)+g_k(y)$, then $x$ and $y$ will not be matched together in the share of the agent $k$ and he can improve his utility for both $x$ and $y$ by matching $x$ with $y$. 

Finally we aim to share equitably the resources among the agents which boils down to ask 
\begin{align}
    \label{eq-3-dual}
    \forall i,j\in\{1,...,N\}~\int u_i d\gamma_i=\int u_j d\gamma_j
\end{align}
Thanks to Proposition~\ref{prop:optimality-cond}, finding  $(\gamma_k,f_k,g_k)_{k=1}^N$ satisfying \eqref{eq-1-dual}, \eqref{eq-2-dual} and \eqref{eq-3-dual} can be done by solving Eq.~\eqref{eq-primal} and Eq.~\eqref{eq-dual}. Indeed let $(\gamma_k)_{k=1}^N$ an optimal solution of Eq.~(\ref{eq-primal}) and $(\lambda,f,g)$ an optimal solution of Eq.~(\ref{eq-dual}). Then by denoting for all $k=1,\dots,N$, 
   $f_k = \frac{f}{\lambda_k}$ and $g_k = \frac{g}{\lambda_k}$,
we obtain that $(\gamma_k,f_k,g_k)_{k=1}^N$ solves the \emph{equitable} and \emph{stable} partition problem in case of \emph{transferable utilities}. Note that again, we end up with equality constraints for the optimal dual variables. Indeed, for all $i,j\in\{1,\dots,N\}$, at optimality we have 
$\int f_i + g_i d\gamma_i = \int f_j + g_j d\gamma_j  \; $. Figure~\ref{fig:potential-dual} illustrates this formulation of the problem with dual potentials. Figure~\ref{fig:potential-dual-ot-viewpoint} in Appendix~\ref{appendix-illustrations} shows the dual solutions with respect to the transport viewpoint in the exact same setting, i.e. $c_i = - u_i$. Once again, the obtained solutions differ.

\subsection{Link with other Probability Metrics}
\label{sec:properties}
In this section, we provide some topological properties on the object defined by the $\MOT$ problem. In particular, we make links with other known probability metrics, such as Dudley and Wasserstein metrics and give a tight upper bound. 
    
When $N=1$, recall from the definition~\eqref{eq-primal} that the problem considered is exactly the standard OT problem. Moreover any $\MOT$ problem with $k\leq N$ costs can always be rewritten as a $\MOT$ problem with $N$ costs. See Appendix~\ref{res:MOT-gene} for the proof. From this property, it is interesting to note that, for any $N\geq 1$, $\MOT$ generalizes standard Optimal Transport.
\paragraph{Optimal Transport.} Given a cost function $c$, if we consider the problem $\MOT$ with $N$ costs such that, for all $i$, $c_i=N\times c$ then, the problem $\MOT_\mathbf{c}$ is exactly $\wass_c$. See Appendix~\ref{res:MOT-gene} for the proof.

Now we have seen that all standard OT problems are sub-cases of the $\MOT$ problem, one may ask whether $\MOT$ can recover other families of metrics different from standard OT. Indeed we show that the $\MOT$ problem recovers an important family of $\textsc{IPM}$s with supremum taken over the space of $\alpha$-Hölder functions with $\alpha\in (0,1]$. See Appendix~\ref{prv:GOT-holder} for the proof.
\begin{prop} 
\label{prop:GOT-holder}
Let $\mathcal{X}$ be a Polish space. Let $d$ be a metric on $\mathcal{X}^2$ and $\alpha\in (0,1]$. Denote $c_1= 2\times\mathbf{1}_{x\neq y}$, $c_2=d^{\alpha}$ and $\mathbf{c}:=(c_1,(N-1)\times c_2,...,(N-1)\times c_2)\in \mathrm{LSC}(\mathcal{X}\times\mathcal{X})^N$
then for any $(\mu,\nu)\in\mathcal{M}_+^{1}(\mathcal{X})\times\mathcal{M}_+^{1}(\mathcal{X})$  
\begin{align}
\label{eq-holder}
    \MOT_{\mathbf{c}}(\mu,\nu) =\sup_{f\in B_{d^{\alpha}}(\mathcal{X})} \int_{\mathcal{X}} f d\mu - \int_{\mathcal{X}} f d\nu
\end{align}
where $B_{d^{\alpha}}(\mathcal{X}):=\left\{f\in C^{b}(\mathcal{X})\mathrm{:}~\Vert f\Vert_{\infty}+\lVert f\rVert_\alpha\leq 1 \right\}$ and $\lVert f\rVert_\alpha := \sup_{x\neq y}\frac{|f(x)-f(y)|}{d^{\alpha}(x,y)}$.
\end{prop}

\paragraph{Dudley Metric.} When $\alpha=1$, then for $(\mu,\nu)\in\mathcal{M}_+^{1}(\mathcal{X})\times\mathcal{M}_+^{1}(\mathcal{X})$, we have 
$$\MOT_{\mathbf{c}}(\mu,\nu) = \MOT_{(c_1,d)}(\mu,\nu) = \beta_d(\mu,\nu)$$
where  $\beta_d$ is the \textit{Dudley Metric}~\citep{dudley1966weak}. In other words, the Dudley metric can be interpreted as an equitable and optimal transport between the measures with the trivial cost and a metric $d$. We acknowledge that~\cite{chizat2018unbalanced} made a link between Unbalanced Optimal Transport and the ``flat metric'', an IPM close to the Dudley metric, defined on the space $\left\{f\mathrm{:}~\Vert f\Vert_{\infty}\leq 1,~\lVert f\rVert_1\leq 1 \right\}$. 

\paragraph{Weak Convergence.} When $d$ is an unbounded metric on $\mathcal{X}$, it is well known that $\wass_{d^{p}}$ with $p\in(0,+\infty)$ metrizes a convergence a bit stronger than weak convergence~\citep[Chap. 7]{villani2003topics}.  A sufficient condition for Wasserstein distances to metrize weak convergence on the space of distributions is that the metric $d$ is bounded. In contrast, metrics defined by Eq.~\eqref{eq-holder} do not require such assumptions and $\MOT_{(\mathbf{1}_{x\neq y},d^{\alpha})}$  metrizes the weak convergence of probability measures~\citep[Chap. 1-7]{villani2003topics}. 

For an arbitrary choice of costs $(c_i)_{1\leq i\leq N}$, we obtain a tight upper control of $\MOT$ and show how it is related to the OT problem associated to each cost involved. See Appendix~\ref{prv:ineqharmonic} for the proof. 
\begin{prop}

\label{prop:ineqharmonic}
Let  $\mathcal{X}$ and $\mathcal{Y}$ be Polish spaces. Let $\mathbf{c}:=(c_i)_{1\leq i\leq N}$ be a family of nonnegative lower semi-continuous costs. For any $(\mu,\nu)\in\mathcal{M}_+^{1}(\mathcal{X})\times\mathcal{M}_+^{1}(\mathcal{Y})$
\begin{align}
\label{eq:harmonic}
    \MOT_{\mathbf{c}}(\mu,\nu)\leq \left(\sum\limits_{i=1}^N\frac{1}{\wass_{c_i}(\mu,\nu)}\right)^{-1} 
\end{align}
\end{prop}
Proposition~\ref{prop:ineqharmonic} means that 
the minimal cost to transport all goods under the constraint that all workers contribute equally is lower than the case where agents share equitably and optimally the transport with distributions $\mu_i$ and $\nu_i$ respectively proportional to $\mu$ and $\nu$, which equals the harmonic sum written in Equation~\eqref{eq:harmonic}.

\begin{example*}
Applying the above result in the case of the  Dudley metric recovers the following inequality~\citep[Proposition 5.1]{sriperumbudur2012empirical} \begin{align*}
    \beta_d(\mu,\nu)\leq \frac{\TV(\mu,\nu)\wass_d(\mu,\nu)}{\TV(\mu,\nu)+\wass_d(\mu,\nu)}.
\end{align*}
\end{example*}

\section{Entropic Relaxation}
\label{sec:entropic}
In their original form, as proposed by Kantorovich~\cite{Kantorovich42}, Optimal Transport distances are not a natural fit for applied problems: they minimize a network flow problem, with a supercubic complexity $(n^3 \log n)$~\citep{Tarjan1997}. Following the work of~\cite{cuturi2013sinkhorn}, we propose an entropic relaxation of $\MOT$, obtain its dual formulation and derive an efficient algorithm to compute an approximation of $\MOT$. 

\subsection{Primal-Dual Formulation}
Let us first extend the notion of Kullback-Leibler divergence for positive Radon measures. Let $\mathcal{Z}$ be a Polish space, for $\mu,\nu\in\mathcal{M}_+(\mathcal{Z})$, we define the generalized Kullback-Leibler divergence as $\KL(\mu||\nu) = \int\log{\frac{d\mu}{d\nu}}d\mu+\int  d\nu - \int d\mu$  if $\mu\ll \nu$, and $+\infty$ otherwise.  We introduce the following regularized version of $\MOT$.

\begin{definition}[Entropic relaxed primal problem]
Let $\mathcal{X}$ and $\mathcal{Y}$ be two Polish spaces, $\mathbf{c}:=(c_i)_{1\leq i\leq N}$ a family of bounded below lower semi-continuous costs lower semi-continuous costs on $\mathcal{X}\times\mathcal{Y}$ and $\bm{\varepsilon}:=(\varepsilon_i)_{1\leq i\leq N}$ be non negative real numbers. For $(\mu,\nu)\in\mathcal{M}_+^{1}(\mathcal{X})\times\mathcal{M}_+^{1}(\mathcal{Y})$, we define the \emph{$\MOT$ regularized primal problem}:
\begin{equation}
\vspace{-0.4cm}
\begin{aligned}
\MOT_{\mathbf{c}}^{\bm{\varepsilon}}(\mu,\nu):= & \inf\limits_{\gamma\in\Gamma_{\mu,\nu}^N}   \max_i\int c_id\gamma_i\label{eq-primal-entrop} 
+\sum_{j=1}^N \varepsilon_j\KL(\gamma_j||\mu\otimes \nu)\notag
\end{aligned}
\end{equation}
\end{definition}
Note that here we sum the generalized Kullback-Leibler divergences since our objective is function of $N$ measures in $\mathcal{M}_+(\mathcal{X}\times\mathcal{Y})$. This problem can be compared with the one from standard regularized OT. In the case where $N=1$, we recover the standard regularized OT. For $N\geq 1$, the underlying problem  is $\sum_{i=1}^{N}\varepsilon_i-$strongly convex.
Moreover, we prove the essential property that as $\bm{\varepsilon}\to 0$, the regularized problem converges to the standard problem. See Appendix~\ref{res:epsto0} for the full statement and the proof. As a consequence, entropic regularization is a consistent approximation of the original problem we introduced in Section~\ref{sec:primal-dual}.
Next theorem shows that strong duality holds for lower semi-continuous costs and compact spaces. This is the basis of the algorithm we will propose in Section~\ref{sec:algorithms}. See Appendix~\ref{prv:duality-entropic} for the proof.
\begin{thm}[Duality for the regularized problem]
\label{thm:duality-entropic}
Let $\mathcal{X}$ and $\mathcal{Y}$ be two compact Polish spaces, $\mathbf{c}:=(c_i)_{1\leq i\leq N}$ a family of bounded below lower semi-continuous costs on $\mathcal{X}\times\mathcal{Y}$ and $\bm{\varepsilon}:=(\varepsilon_i)_{1\leq i\leq N}$ be non negative numbers. For $(\mu,\nu)\in\mathcal{M}^1_{+}(\mathcal{X})\times\mathcal{M}_+^{1}(\mathcal{Y})$, strong duality holds:
\begin{align}
    \MOT_{\mathbf{c}}^{\bm{\varepsilon}}&(\mu,\nu)=\sup_{\lambda\in\Delta^+_N} \sup\limits_{\substack{f\in\mathcal{C}_b(\mathcal{X})\\g\in\mathcal{C}_b(\mathcal{Y})}} \int fd\mu+ \int gd\nu\label{eq-dual-entrop}\\
 &-\sum_{i=1}^N\varepsilon_i\left( \int e^{\frac{f(x)+g(y)-\lambda_ic_i(x,y)}{\varepsilon_i}} d\mu(x)d\nu(y)-1\right)\notag
\end{align}
and the infimum of the primal problem is attained. 
\end{thm}
As in standard regularized optimal transport there is a link between primal and dual variables at optimum. Let $\gamma^*$ solving the reguralized primal problem and $(f^*,g^*,\lambda^*)$ solving the dual one: $$\forall i,~\gamma_i^* = \exp\left(\frac{f^*+g^*-\lambda^*_i c_i}{\varepsilon_i}\right)\cdot\mu\otimes\nu$$.
\vspace{-0.6cm}
\subsection{Proposed Algorithms}
\label{sec:algorithms}


\begin{algorithm}[ht!]
\SetAlgoLined
\textbf{Input:} $\mathbf{C}=(C_i)_{1\leq i\leq N}$, $a$, $b$, $\varepsilon$, $L_{\lambda}$\\
\textbf{Init:} $f^0\leftarrow \mathbf{1}_n\text{;  }$ $g^0 \leftarrow \mathbf{1}_m\text{;  }$ $\lambda^0 \leftarrow (1/N,...,1/N)\in\mathbb{R}^N$\\
\For{$k=1,2,...$}{$
K^k \leftarrow \sum_{i=1}^N K_i^{\lambda_i^{k-1}},\\
c_k \leftarrow \langle f^{k-1}, K^k g^{k-1}\rangle,~f^{k} \leftarrow \frac{c_k a}{K^{k}g^{k-1}},\\
d_k \leftarrow\langle f^{k}, K^k g^{k-1}\rangle,~g^{k} \leftarrow \frac{d_k b}{ (K^k)^Tf^{k}}, \\
\lambda^k \leftarrow \mathrm{Proj}_{\Delta^+_N}\left( \lambda^{k-1} + \frac{1}{L_\lambda}\nabla_{\lambda} F_{\mathbf{C}}^{\varepsilon}(\lambda^{k-1},f^{k},g^{k})\right).$}
\caption{Projected Alternating Maximization \label{algo:Proj-Sinkhorn}}
\textbf{Result}: $\lambda,f,g$
\end{algorithm}

We can now present algorithms obtained from entropic relaxation to approximately compute the solution of $\MOT$. Let $\mu=\sum_{i=1}^n a_i\delta_{x_i}$ and $\nu=\sum_{j=1}^m b_j\delta_{y_j}$ be discrete probability measures where $a\in\Delta_n^{+}$, $b\in\Delta^+_m$, $\{x_1,...,x_n\}\subset \mathcal{X}$ and $\{y_1,...,y_m\}\subset \mathcal{Y}$. Moreover for all $i\in\{1,\dots,N\}$ and $\lambda>0$, define $\mathbf{C}:=(C_i)_{1\leq i\leq N}\in\left(\mathbb{R}^{n\times m}\right)^N$ with $C_i:=(c_i(x_k,y_l))_{k,l}$ the $N$ cost matrices and $K_i^{\lambda}:=\exp\left(-\lambda C_{i}/\varepsilon\right)$. Assume that  $\varepsilon_1=\dots=\varepsilon_N=\varepsilon$.
Compared to the standard regularized OT,  the main difference here is that the problem contains an additional variable $\lambda\in\Delta_N^{+}$. 
When $N=1$, one can use Sinkhorn algorithm. However when $N\geq 2$, we do not have a closed form for updating $\lambda$ when the other variables of the problem are fixed. In order to enjoy from the strong convexity of the primal formulation, we consider instead the dual associated with the equivalent primal problem given when the additional trivial constraint $\mathbf{1}_n^T\left(\sum_i P_i\right)\mathbf{1}_m = 1$  is considered. In that the dual obtained is
\begin{align*}
\label{eq-obj-proj-sin}
    \widehat{\MOT}_{\mathbf{C}}^{\bm{\varepsilon}}(a,b)=&\sup\limits_{\substack{\lambda\in\Delta_N^{+}\\f\in \mathbb{R}^n,~g\in \mathbb{R}^m}} \langle f, a\rangle+\langle g,b\rangle
    -\varepsilon\left[\log\left(\sum_i\langle e^{\mathbf{f}/\varepsilon},K_i^{\lambda_i} e^{\mathbf{g}/\varepsilon}\rangle \right) + 1\right]
\end{align*}
We show that the new objective obtained above is smooth w.r.t $(\lambda,f,g)$. See Appendix~\ref{res:pgd} for the proof. One can  apply the accelerated projected gradient ascent~\citep{beck2009fast,tseng2008accelerated} which enjoys an optimal convergence rate for first order methods of $\mathcal{O}(k^{-2})$ for $k$ iterations.%
\begin{figure*}[ht!]
\vspace{-0.3cm}
\begin{tabular}{ c  c  c  c}
\includegraphics[width=0.28\textwidth
]{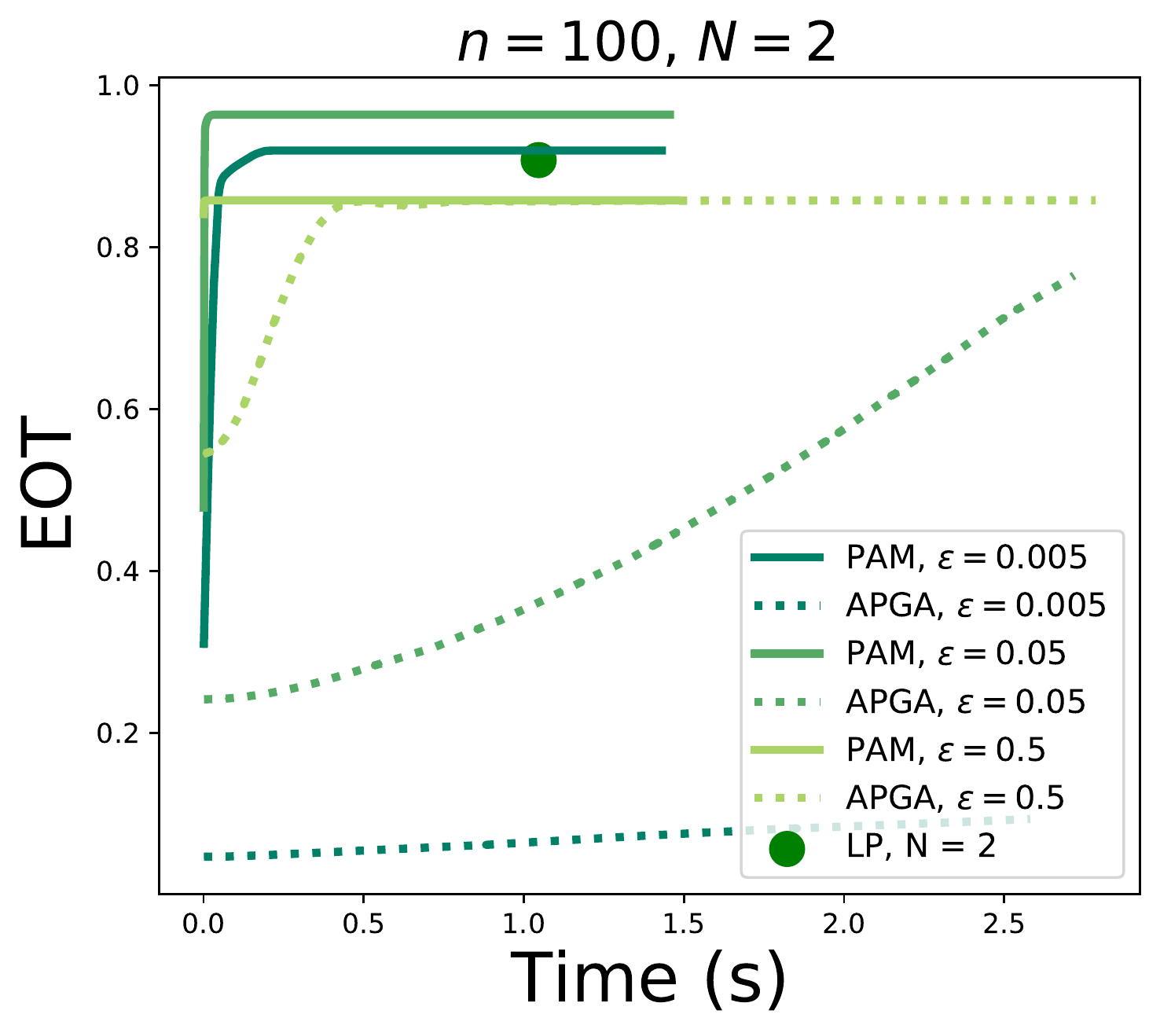} &
\includegraphics[width=0.28\textwidth
]{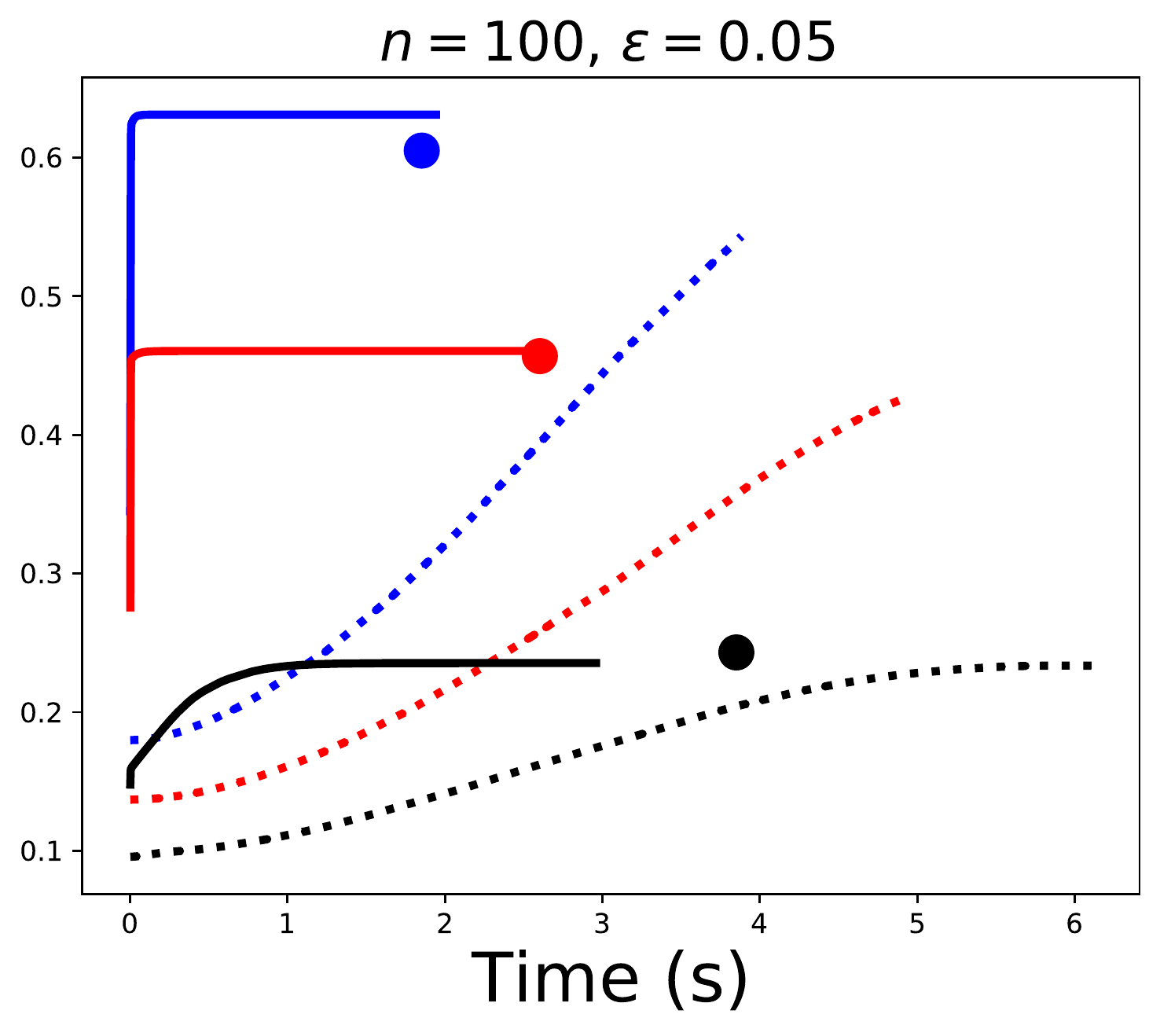}
&
\includegraphics[width=0.28\textwidth
]{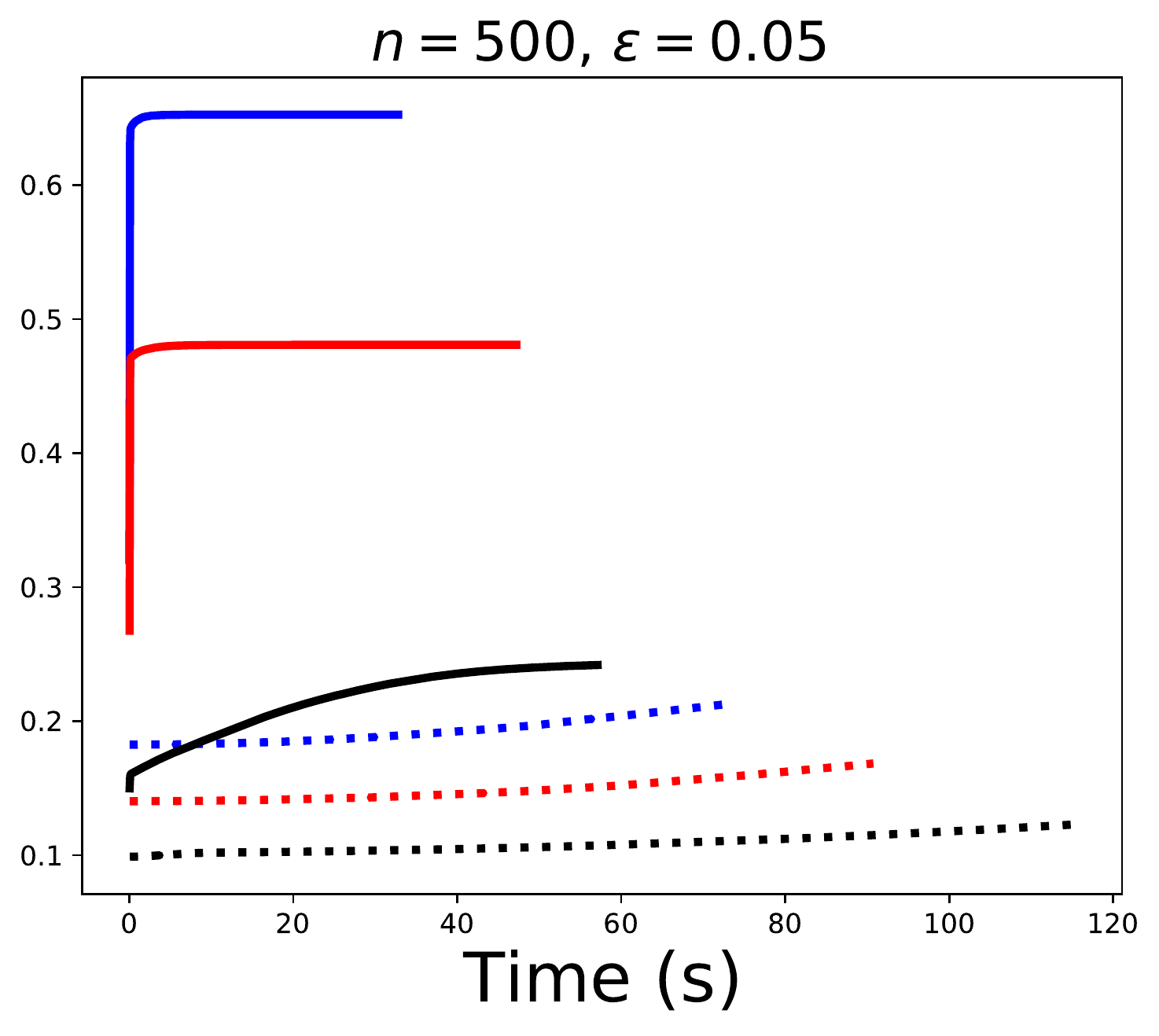} &
\hspace*{-0.6cm}
\raisebox{.9\height}{\includegraphics[width=0.08\textwidth
]{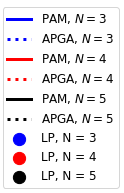}}
\end{tabular}
\caption{Comparison of the time-accuracy tradeoffs between the different proposed algorithms. \emph{Left:} we consider the case where the number of days is $N=2$, the size of support for both measures is $n=m=100$ and we vary $\varepsilon$ from $0.005$ to $0.5$. \emph{Middle:} we fix $n=m=100$ and the regularization $\varepsilon=0.05$ and we vary the number of days $N$ from 3 to 5. \emph{Right:} the setting considered is the same as in the figure in the middle, however we increase the sample size such that $n=m=500$. Note that in that case, $\textbf{LP}$ is too costly to be computed. \label{fig-seq-OT}}
\end{figure*}

It is also possible to adapt Sinkhorn algorithm to our problem. See Algorithm~\ref{algo:Proj-Sinkhorn}. We denoted by $\mathrm{Proj}_{\Delta_N^+}$ the orthogonal projection on $\Delta_N^+$~\citep{shalev2006efficient}, whose complexity is in $\mathcal{O}(N\log{N})$.  The smoothness constant in $\lambda$ in the algorithm is $L_\lambda = \max_{i}\Vert C_i\Vert_{\infty}^2/\varepsilon$. In practice  Alg.~\ref{algo:Proj-Sinkhorn} gives better results than the accelerated gradient descent. Note that the proposed algorithm differs from the Sinkhorn algorithm in many points and therefore the convergence rates cannot be applied here. Analyzing the rates of a \emph{projected} alternating maximization method is, to the best of our knowledge, an unsolved problem. Further work will be devoted to study the convergence of this algorithm. We illustrate Algorithm~\ref{algo:Proj-Sinkhorn} by showing the convergence of the regularized version of $\MOT$ towards the ground truth when $\varepsilon\to 0$ in the case of the Dudley Metric. See Figure~\ref{fig:result_acc} in Appendix~\ref{appendix-illustrations}.

\section{Other applications of EOT}

\paragraph{Minimal Transportation Time.} Assume there are $N$ internet service providers who propose different debits to transport data across locations, and one needs to transfer data from multiple servers to others, the fastest as possible.   We assume that $c_i(x,y)\geq 0$ corresponds to the transportation time needed by provider $i$ to transport one unit of data from a server $x$ to a server $y$. For instance, the unit of data can be one Megabit. Then $\int c_i d\gamma_i$ corresponds the time taken by provider $i$ to transport $\mu_i=\Pi_{1\sharp}\gamma_i$ to $\nu_i=\Pi_{2\sharp}\gamma_i$. Assuming the transportation can be made in parallel and given a partition of the transportation task $(\gamma_i)_{i=1}^N$, $\max_i\int c_i d\gamma_i$ corresponds to the total time of transport the data $\mu=\Pi_{1\sharp}\sum\gamma_i $ to the locations $\nu=\Pi_{2\sharp}\sum\gamma_i$ according to this partition. Then $\MOT$, which minimizes $\max_i\int c_i d\gamma_i$, is finding the fastest way to transport the data from $\mu$ to $\nu$ by splitting the task among the $N$ internet service providers. Note that at optimality, all the internet service providers finish their transportation task at the same time (see Proposition~\ref{prop:mot-equality}).
 
\paragraph{Sequential Optimal Transport.} Consider the situation where an agent aims to transport goods from some stocks to some stores in the next $N$ days. The cost to transport one unit of good from a stock located at $x$ to a store located at $y$ may vary across the days. For example the cost of transportation may depend on the price of gas, or the daily weather conditions. Assuming that he or she has a good knowledge of the daily costs of the $N$ coming days, he or she may want a transportation strategy such that his or her daily cost is as low as possible. By denoting $c_i$ the cost of transportation the $i$-th day, and given a strategy $(\gamma_i)_{i}^N$, the maximum daily cost is then $\max_i\int c_i d\gamma_i$, and $\MOT$ therefore finds the cheapest strategy to spread the transport task in the next $N$ days such that the maximum daily cost is minimized. Note that at optimality he or she has to spend the exact same amount everyday.

In Figure~\ref{fig-seq-OT} we aim to simulate the Sequential OT problem and compare the time-accuracy trade-offs of the proposed algorithms. Let us consider a situation where one wants to transport merchandises from $\mu = \frac{1}{n}\sum_{i=1}^n  \delta_{x_i}$ to $\nu =\frac{1}{m} \sum_{j=1}^m \delta_{y_j}$ in $N$ days. Here we model the locations  $\{x_i\}$ and $\{y_j\}$ by drawing them independently from two Gaussian distributions in $\mathbb{R}^2$: $\forall i,~x_i\sim \mathcal{N}\left(\begin{psmallmatrix}
3\\
3\\
\end{psmallmatrix},\begin{psmallmatrix}
0 & 1\\
1 & 0 \\
\end{psmallmatrix}\right)$ and $\forall j,~y_j\sim \mathcal{N}\left(\begin{psmallmatrix}
4\\
4\\
\end{psmallmatrix},\begin{psmallmatrix}
1 & -.2\\
-.2 & 1\\
\end{psmallmatrix}\right).$
We assume that everyday there is wind modeled by a vector $w\sim \mathcal{U}(B(0,1))$ where $B(0,1)$ is the unit ball in $\mathbb {R}^2$ that is perfectly known in advance. We define the cost of transportation on day $i$ as $c_i(x,y) = \lVert y-x\rVert -0.7 \langle w_i,y-x\rangle$ to model the effect of the wind on the transportation cost. In the following figures we plot the estimates of EOT obtained from the proposed algorithms in function of the runtime for various sample sizes $n$, number of days $N$ and regularizations $\varepsilon$. \textbf{PAM} denotes Alg.~\ref{algo:Proj-Sinkhorn}, \textbf{APGA} denotes Alg.~\ref{algo:Proj-grad} (See Appendix C.4), \textbf{LP} denotes the linear program which solves exactly the primal formulation of the EOT problem. Note that when $\textbf{LP}$ is computable (i.e. $n\leq 100$), it is therefore the ground truth. We show that in all the settings, \textbf{PAM} performs better than $\textbf{APGA}$ and provides very high accuracy with order of magnitude faster than $\text{LP}$.

\newpage
\bibliography{biblio}
\bibliographystyle{plainnat}

\clearpage
\appendix

\onecolumn
\section*{Supplementary material}

\section{Proofs}
\label{sec:proofs}

\subsection{Notations}

Let $\mathcal{Z}$ be a Polish space, we denote $\mathcal{M}(\mathcal{Z})$ the set of Radon measures on $\mathcal{Z}$ endowed with total variation norm: $\lVert\mu\rVert_{\TV}=\mu_+(\mathcal{Z})+\mu_-(\mathcal{Z})$ with $(\mu_+,\mu_-)$ is the Dunford decomposition of the signed measure $\mu$. We call $\mathcal{M}_+(\mathcal{Z})$ the sets of positive Radon measures, and  $\mathcal{M}^1_+(\mathcal{Z})$ the set of probability measures. We denote $\mathcal{C}^b(\mathcal{Z})$ the vector space of bounded continuous functions on $\mathcal{Z}$ endowed with $\lVert\cdot \rVert_\infty$ norm. We recall the \textit{Riesz-Markov theorem}: if $\mathcal{Z}$ is compact, $\mathcal{M}(\mathcal{Z})$ is the topological dual of $\mathcal{C}^b(\mathcal{Z})$. Let $\mathcal{X}$ and $\mathcal{Y}$ be two Polish spaces. It is immediate  that \textit{$\mathcal{X}\times\mathcal{Y}$ is a Polish space}.  We denote for $\mu\in\mathcal{M}(\mathcal{X})$ and $\nu\in\mathcal{M}(\mathcal{Y})$, $\mu\otimes\nu$ the tensor product of the measures $\mu$ and $\nu$, and $\mu\ll\nu$ means that $\nu$ dominates $\mu$.  We denote $\Pi_1:(x,y)\in\mathcal{X}\times\mathcal{Y}\mapsto x$ and $\Pi_2:(x,y)\in\mathcal{X}\times\mathcal{Y}\mapsto y$ respectively the projections on $\mathcal{X}$ and  $\mathcal{Y}$, which are continuous applications. For an application $g$ and a measure $\mu$, we denote $g_\sharp\mu$ the pushforward measure of $\mu$ by $g$. For $f:\mathcal{X}\rightarrow\mathbb{R}$ and $g:\mathcal{Y}\rightarrow\mathbb{R}$, we denote $f\oplus g:(x,y)\in\mathcal{X}\times\mathcal{Y}\mapsto f(x)+g(y)$ the tensor sum of $f$ and $g$. For  $\mathcal{X}$ and $\mathcal{Y}$ two Polish spaces, we denote $\text{LSC}(\mathcal{X}\times\mathcal{Y})$ the space of lower semi-continuous functions on $\mathcal{X}\times\mathcal{Y}$,  $\mathrm{LSC}^+(\mathcal{X}\times\mathcal{Y})$ the space of non-negative lower semi-continuous functions on $\mathcal{X}\times\mathcal{Y}$ and $\mathrm{LSC}^-_{*}(\mathcal{X}\times\mathcal{Y})$ the set of negative bounded below lower semi-continuous functions on $\mathcal{X}\times\mathcal{Y}$ .  Let $N\geq 1$ be an integer and denote $\Delta_N^{+}: = \{\lambda\in\mathbb{R}_+^N~\mathrm{s.t.}~\sum_{i=1}^N\lambda_i=1\}$, the probability simplex of $\mathbb{R}^N$.
For two positive measures of same mass $\mu\in\mathcal{M}_+(\mathcal{X})$ and $\nu\in\mathcal{M}_+(\mathcal{Y})$, we define the set of couplings with marginals $\mu$ and $\nu$:
\begin{align*}
    \Pi_{\mu,\nu}:=\left\{\gamma~\mathrm{s.t.}~ \Pi_{1\sharp}\gamma=\mu ~,~ \Pi_{2\sharp}\gamma=\nu\right\}\; .
\end{align*}
For $\mu\in\mathcal{M}_+^1(\mathcal{X})$ and $\nu\in\mathcal{M}_+^1(\mathcal{Y})$, we introduce the subset of $(\mathcal{M}_+^1(\mathcal{X})\times \mathcal{M}_+^1(\mathcal{Y}))^N$ representing marginal decomposition: 
\begin{align*}
 \textstyle\Upsilon_{\mu,\nu}^N:=\left\{(\mu_i,\nu_i)_{i=1}^N ~\mathrm{ s.t. }~ \sum_i \mu_i = \mu \mathrm{, } \sum_i \nu_i = \nu ~\mathrm{ and }~\forall i,~ \mu_i(\mathcal{X}) = \nu_i(\mathcal{Y})  \right\}.
\end{align*}
We also define the following subset of $\mathcal{M}_+(\mathcal{X}\times \mathcal{Y})^N$ corresponding to the coupling decomposition:
\begin{align*}
    \Gamma^N_{\mu,\nu}:=\left\{(\gamma_i)_{i=1}^N~\mathrm{s.t.}~ \Pi_{1\sharp}\sum_i\gamma_i=\mu ~,~ \Pi_{2\sharp}\sum_i\gamma_i=\nu\right\} .
\end{align*}

\subsection{Proof of Proposition~\ref{prop:mot-equality}}
\label{prv:mot-equality}
\begin{prv*}

First, it is clear that $\MOT_{\mathbf{c}}(\mu,\nu) \geq  \inf_{\gamma\in\Gamma^N_{\mu,\nu}}\{t\text{ s.t. }\forall i,~t=\int c_id\gamma_i\}$. Let us now show that in fact it is an equality. Thanks to Theorem~\ref{thm:duality-GOT}, the infimum is attained for $\inf_{\gamma\in\Gamma_{\mu,\nu}}\max_i\int c_id\gamma_i$. Indeed recall that $\Gamma^N_{\mu,\nu}$ is compact and that the objective is lower semi-continuous. Let $\gamma^*$ be such a minimizer. Let $I$ be the set of indices $i$ such that $\int c_id\gamma^*_i=\MOT_{\mathbf{c}}(\mu,\nu)$. Assume that there exists $j$ such that, $\MOT_{\mathbf{c}}(\mu,\nu)>\int c_jd\gamma^*_j$. 

In case of costs of $\mathrm{LSC}^+(\mathcal{X}\times\mathcal{Y})$, for all $i\in I$, there exists $(x_i,y_i)\in \text{Supp}(\gamma^*_i)$ such that $c_i(x_i,y_i)>0$. Let us denote $A_{(x_i,y_i)}$ measurable sets such that $(x_i,y_i)\in A_{(x_i,y_i)}$ and let us denote $\tilde{\gamma}$ defined as for all $k\notin I\cup\{j\}$, $\tilde{\gamma}_k = \gamma_k^*$, for $i\in I$, $\tilde{\gamma}_i = \gamma^*_i-\epsilon \mathbf{1}_{A_{(x_i,y_i)}}\gamma^*_i$ and $\tilde{\gamma}_j = \gamma^*_j+\sum_{i\in I}\epsilon \mathbf{1}_{A_{(x_i,y_i)}}\gamma^*_i$ for $\epsilon$ sufficiently small so that $\tilde{\gamma}\in\Gamma^N_{\mu,\nu}$. Now, $\max_k \int c_k d\gamma^*_k>\max_k \int c_k d\tilde{\gamma}_k$, which contradicts that $\gamma^*$ is a minimizer. Then for $i,j$, $\int c_id\gamma^*_i=\int c_jd\gamma^*_j$. And then:
$\MOT_{\mathbf{c}}(\mu,\nu) = \inf_{\gamma\in\Gamma^N_{\mu,\nu}}\max_i\int c_id\gamma_i$.

In case of costs in $\mathrm{LSC}^-_{*}(\mathcal{X}\times\mathcal{Y})$, there exists $(x_0,y_0)\in \text{Supp}(\gamma^*_j)$ such that $c_j(x_0,y_0)<0$. Let us denote $A_{(x_0,y_0)}$ a measurable set such that $(x_0,y_0)\in A_{(x_0,y_0)}$ and let us denote $\tilde{\gamma}$ defined as for all $k\notin I\cup\{j\}$, $\tilde{\gamma}_k = \gamma_k^*$ and for all $i\in I$, $\tilde{\gamma}_i = \gamma^*_i+\frac{\epsilon}{\lvert I\rvert} \mathbf{1}_{A_{(x_0,y_0)}}\gamma^*_j$ and $\tilde{\gamma}_j = \gamma^*_j-\epsilon\mathbf{1}_{A_{(x_0,y_0)}}\gamma^*_j$ for $\epsilon$ sufficiently small so that $\tilde{\gamma}\in\Gamma^N_{\mu,\nu}$. Now, $\max_k \int c_k d\gamma^*_k>\max_k \int c_k d\tilde{\gamma}_i$, which contradicts that $\gamma^*$ is a minimizer. Then for $i,j$, $\int c_id\gamma^*_i=\int c_jd\gamma^*_j$. And then:
$\MOT_{\mathbf{c}}(\mu,\nu) = \inf_{\gamma\in\Gamma^N_{\mu,\nu}}\max_i\int c_id\gamma_i$.

It is clear that equitability is verified thanks to the previous proof. For proportionality, assume the normalization:  $\forall i$, there exists $\gamma_i\in\mathcal{M}_+^1(\mathcal{X}\times\mathcal{Y})$ such that $V_i(\gamma_i)=1$. Then for each $i$, $V_i(\gamma_i/N)=1/N$ and $(\gamma_i)_i\in\Gamma^N_{\mu,\nu}$. Then at optimum: $\forall i$, $V_i(\gamma_i^*)\geq 1/N$ and proportionality is verified.

\end{prv*}

\subsection{Proof of Proposition~\ref{prop:mot-otplans}}
\label{prv:mot-otplans}

\begin{prv*}

We prove along with Theorem~\ref{thm:duality-GOT} that the infimum defining $\MOT_\mathbf{c}(\mu,\nu)$ is attained. Let $\gamma^*$ be this infimum. Then at optimum we have shown that for all $i,j$, $\int c_id\gamma^*_i=\int c_jd\gamma^*_j = t$. Let denote for all $i$, $\mu_i=\Pi_{1\sharp}\gamma^*_i$ and $\nu_i=\Pi_{2\sharp}\gamma^*_i$. 

Let assume there exists $i$ such that $\int c_i d\gamma^*_i>\wass_{c_i}(\mu_i,\nu_i)$. Let $\gamma'_i$ realising the infimum of $\wass_{c_i}(\mu_i,\nu_i)$. Let $\epsilon>0$ be sufficiently small, then let define $\tilde{\gamma}$ as follows: for all $j\neq i$, $\tilde{\gamma}_j=(1-\epsilon)\gamma^*_j$. and $\tilde{\gamma}_i = \gamma'_i+\epsilon \sum_{j\neq i}\gamma^*_j$.
Then for all $j\neq i$, $\int c_j d \tilde{\gamma}_j = (1-\epsilon) t$ and $\int c_i d \tilde{\gamma}_i = \wass_{c_i}(\mu_i,\nu_i)+\epsilon \sum_{j\neq i} \int c_id\gamma^*_j$. It is clear that $\tilde{\gamma}\in \Gamma^N_{\mu,\nu}$. For $\epsilon>0$ sufficiently small, $\max_i\int c_id\tilde{\gamma}_i = (1-\epsilon) t<t$, which contradicts the optimality of $\gamma^*$.

A possible reformulation for $\MOT$ is:
\begin{align*}
\MOT_\mathbf{c}(\mu,\nu) = \min_{\substack{(\mu_i,\nu_i)_{i=1}^N\in\Upsilon^N_{\mu,\nu}\\ \forall i,~\gamma_i\in\Pi_{\mu,\nu}}}\left\{t~\mathrm{s.t.}~\int c_id\gamma_i=t\right\}
\end{align*}
We previously show that at optimum the couplings are optimal transport plans, then:
\begin{align*}
\MOT_\mathbf{c}(\mu,\nu) =\min_{\substack{(\mu_i,\nu_i)_{i=1}^N\in\Upsilon^N_{\mu,\nu}}} \left\{t~\mathrm{s.t.}~ \forall i,~ \wass_{c_i}(\mu_i,\nu_i) =t\right\}
\end{align*}
which concludes the proof.
\end{prv*}

\subsection{Proof of Theorem~\ref{thm:duality-GOT}}
\label{prv:duality-GOT}

To prove this theorem, one need to prove the three following technical lemmas. The first one shows the weak compacity of $\Gamma^N_{\mu,\nu}$.

\begin{lemma}
\label{lem:compact-weak}
Let $\mathcal{X}$ and $\mathcal{Y}$ be Polish spaces, and $\mu$ and $\nu$ two probability measures respectively on  $\mathcal{X}$ and $\mathcal{Y}$. Then $\Gamma^N_{\mu,\nu}$  is sequentially compact for the weak topology induced by $\Vert \gamma \Vert = \max\limits_{i=1,..,N} \Vert \gamma_i\Vert_{\TV}$. 
\end{lemma}

\begin{prv*}
Let $(\gamma^n)_{n\geq 0}$ a sequence in $\Gamma^N_{\mu,\nu}$, and let us denote for all $n\geq 0$, $\gamma^n=(\gamma^n_i)_{i=1}^N$. We first remark that for all $i\in\{1,...,N\}$ and $n\geq 0$, $\Vert \gamma_i^n\Vert_{\TV}\leq 1$ therefore for all $i\in\{1,...,N\}$, $(\gamma^n_i)_{n\geq 0}$ is uniformly bounded. Moreover as $\{\mu\}$ and $\{\nu\}$ are tight, for any $\delta>0$, there exist $K\subset \mathcal{X} $ and $L\subset \mathcal{Y}$ compact sets such that 
\begin{align}
    \mu(K^c)\leq \frac{\delta}{2} \text{\quad and\quad }  \nu(L^c)\leq \frac{\delta}{2}.
\end{align}
Therefore, we obtain that for any for all $i\in\{1,...,N\}$,
\begin{align}
    \gamma_i^n(K^c\times L^c)&\leq \sum_{k=1}^N \gamma_k^n(K^c\times L^c)\\
    &\leq  \sum_{k=1}^N \gamma_k^n(K^c\times\mathcal{Y})+\gamma_k^n(\mathcal{X}\times L^c)\\
    &\leq  \mu(K^c) + \nu(L^c) = \delta.
\end{align}
Therefore, for all $i\in\{1,...,N\}$,  $(\gamma_i^n)_{n\geq 0}$ is tight and uniformly bounded and Prokhorov's theorem~\citep[Theorem A.3.15]{dupuis2011weak} guarantees for all $i\in\{1,...,N\}$,  $(\gamma_i^n)_{n\geq 0}$ admits a weakly convergent subsequence. By extracting a common convergent subsequence, we obtain that $(\gamma^n)_{n\geq 0}$ admits a weakly convergent subsequence. By continuity of the projection, the limit also lives in $\Gamma
^N_{\mu,\nu}$ and the result follows.
\end{prv*}

Next lemma generalizes Rockafellar-Fenchel duality to our case.
\begin{lemma}
\label{lem:rockafellar-gene}
Let $V$ be a normed vector space and $V^*$ its topological dual. Let $V_1,...,V_N$ be convex functions and lower semi-continuous on $V$ and $E$ a convex function on $V$. Let $V^*_1,...V^*_N,E^*$ be the Fenchel-Legendre transforms of $V_1,...V_N,E$. Assume there exists $z_0\in V$ such that for all $i$, $V_i(z_0)<\infty$, $E(z_0)<\infty$, and for all $i$, $V_i$ is continuous at $z_0$. Then:
\begin{align*}
\inf_{u\in V} \sum_i V_i(u) + E(u) = \sup\limits_{\substack{\gamma_1...,\gamma_N,\gamma\in V^*\\\sum_i \gamma_i = \gamma}}-\sum_i V^*_i(-\gamma_i)-E^*(\gamma)
\end{align*}
\end{lemma}
\begin{prv*}
This Lemma is an immediate application of Rockafellar-Fenchel duality theorem~\citep[Theorem 1.12]{brezis2010functional} and of Fenchel-Moreau theorem~\citep[Theorem 1.11]{brezis2010functional}. 
Indeed, $V = \sum\limits_{i=1}^N V_i(u)$ is a convex function, lower semi-continuous and its Legendre-Fenchel transform is given by:
\begin{align}
    V^{*}(\gamma^*)=\inf_{\sum\limits_{i=1}^N \gamma_{i}^*=\gamma^*}\sum_{i=1}^N V_i^{*}(\gamma_{i}^*).
\end{align}
\end{prv*}




Last lemma is an application of Sion's Theorem to this problem.
\begin{lemma}
\label{lem:technical-lemma-primal}
Let $\mathcal{X}$ and $\mathcal{Y}$ be Polish spaces. Let $\mathbf{c}=(c_i)_{1\leq i\leq N}$ be a family of bounded below lower semi-continuous costs on $\mathcal{X}\times \mathcal{Y}$, then for $\mu\in\mathcal{M}^1_+(\mathcal{X})$ and  $\nu\in\mathcal{M}^1_+(\mathcal{Y})$, we have
\begin{align}
\label{eq:supinf}
\MOT_{\mathbf{c}}(\mu,\nu)= \sup_{\lambda\in\Delta_N^{+}}\inf_{\gamma\in\Gamma_{\mu,\nu}^N} \sum_{i=1}^N\lambda_i\int_{\mathcal{X}\times\mathcal{Y}}c_i(x,y)d\gamma_i(x,y)
\end{align}
and the infimum is attained.
\end{lemma}
\begin{prv*}
Taking for granted that a minmax principle can be invoked, we have
\begin{align*}
    \sup_{\lambda\in\Delta_N^{+}}\inf_{\gamma\in\Gamma_{\mu,\nu}^N} \sum_{i=1}^N\lambda_i\int_{\mathcal{X}\times\mathcal{Y}}c_i(x,y)d\gamma_i(x,y) &= \inf_{\gamma\in\Gamma_{\mu,\nu}^N}\sup_{\lambda\in \Delta_N^{+}} \sum_{i=1}^N\lambda_i\int_{\mathcal{X}\times\mathcal{Y}}c_i(x,y)d\gamma_i(x,y)\\
    &=\MOT_{\mathbf{c}}(\mu,\nu)
\end{align*}
But thanks to Lemma \ref{lem:compact-weak}, we have that $\Gamma_{\mu,\nu}^N$ is compact for the weak topology. And $\Delta^+_N$ is convex. Moreover the objective function $f:(\lambda,\gamma)\in\Delta^+_N\times\Gamma^N_{\mu,\nu} \mapsto \sum_{i=1}^N\lambda_i \int_{\mathcal{X}\times \mathcal{Y}} c^n_id\gamma_i$ is bilinear, hence convex and concave  in its variables, and continuous with respect to $\lambda$. Moreover, let $(c^n_i)_n$ be non-decreasing sequences of bounded cost functions such that $c_i=\sup_n c^n_i$. By monotone convergence, we get $f(\lambda,\gamma) = \sup_n \sum_i\lambda_i \int c^n_i d\gamma_i$, $f(\lambda,.)$. So $f$ the supremum of continuous functions, then $f$ is lower semi-continuous with respect to $\gamma$, therefore Sion's minimax theorem
~\citep{sion1958} holds.

\end{prv*}

We are now able to prove Theorem~\ref{thm:duality-GOT}.

\begin{prv*}
Let $\mathcal{X}$ and $\mathcal{Y}$ be two Polish spaces. For all $i\in \{1,..,N\}$, we define $c_i:\mathcal{X}\times\mathcal{Y}\rightarrow \mathbb{R}$ a bounded below lower-semi cost function. The proof follows the exact same steps as those in the proof of~\citep[Theorem 1.3]{villani2003topics}. First we suppose that $\mathcal{X}$ and $\mathcal{Y}$ are compact and that for all $i$, $c_i$ is continuous, then we show that it can be extended to $X$ and $Y$ non compact and finally to $c_i$ only lower semi continuous.

\medskip
First, let assume $\mathcal{X}$ and $\mathcal{Y}$ are compact and that for all $i$, $c_i$ is continuous. Let fix $\lambda \in\Delta^+_N$. We recall the topological dual of the space of bounded continuous functions $\mathcal{C}^b(\mathcal{X}\times\mathcal{Y})$ endowed with $\lVert.\rVert_\infty$ norm, is the space of Radon measures $\mathcal{M}(\mathcal{X}\times\mathcal{Y})$ endowed with total variation norm. We define, for $u\in \mathcal{C}^b(\mathcal{X}\times\mathcal{Y})$:
\begin{align*}
V^\lambda_i(u) =
\left\{\begin{matrix} 0 &\quad\text{if}\quad& u\geq -\lambda_i c_i\\
+\infty &\quad\text{else}\quad&\end{matrix}\right.
\end{align*}
and:
\begin{align*}
E(u)=\left\{\begin{matrix} \int fd\mu+\int gd\nu &\quad\text{if}\quad& \exists (f,g)\in \mathcal{C}^b(\mathcal{X})\times \mathcal{C}^b(\mathcal{Y}),~ u = f+g\\
+\infty &\quad\text{else}\quad&\end{matrix}\right.
\end{align*}
One can show that for all $i$, $V^\lambda_i$ is convex and lower semi-continuous (as the sublevel sets are closed) and $E^\lambda$ is convex. More over for all $i$, these functions continuous in $u_0\equiv 1$ the hypothesis of Lemma~\ref{lem:rockafellar-gene} are satisfied.

Let now compute the Fenchel-Legendre transform of these function.  Let $\gamma\in \mathcal{M}(\mathcal{X}\times\mathcal{Y})$ :

\begin{align*}
V^{\lambda*}_i(-\gamma) &= \sup_{u\in \mathcal{C}^b(\mathcal{X}\times\mathcal{Y})}\left\{-\int ud\gamma;\quad u\geq-\lambda_i c_i\right\}
\\
& = \left\{\begin{matrix}\int \lambda_i c_i d\gamma &\quad\text{if} \quad& \gamma\in\mathcal{M}_+(\mathcal{X}\times\mathcal{Y}) \\
+\infty &\quad\text{otherwise}\quad& \end{matrix}\right.
\end{align*}

On  the other hand:
\begin{align*}
E^{\lambda*}(\gamma)=\left\{\begin{matrix} 0 &\quad\text{if}\quad& \forall (f,g)\in \mathcal{C}^b(\mathcal{X})\times \mathcal{C}^b(\mathcal{Y}),~ \int fd\mu+\int gd\nu  = \int (f+g)d\gamma\\
+\infty &\quad\text{else}\quad&\end{matrix}\right.
\end{align*}
This dual function is finite and equals $0$ if and only if that the marginals of the dual variable $\gamma$ are $\mu$ and $\nu$. 

Applying Lemma~\ref{lem:rockafellar-gene}, we get:
\begin{align*}
\inf_{u\in \mathcal{C}^b(\mathcal{X}\times\mathcal{Y})} \sum_i V^{\lambda}_i(u)+E(u) = \sup\limits_{\substack{\gamma_1,...,\gamma_N,\gamma\in \mathcal{M}(\mathcal{X}\times\mathcal{Y})\\\sum\gamma_i=\gamma}}\sum -V^{\lambda*}_i(\gamma_i)-E^{\lambda*}(-\gamma)
\end{align*}

Hence, we  have shown that, when $\mathcal{X}$ and $\mathcal{Y}$ are compact sets, and  the costs $(c_i)_i$ are continuous:
\begin{align*}
\sup\limits_{(f,g)\in\mathcal{F}^\lambda_\mathbf{c}}\int fd\mu+\int gd\nu = \inf\limits_{\gamma\in\Gamma^N_{\mu,\nu}}\sum_i\lambda_i \int c_id\gamma_i
\end{align*}


\medskip

Let now prove  the result holds when the spaces $\mathcal{X}$ and $\mathcal{Y}$ are not compact. We still suppose that for all $i$, $c_i$ is uniformly continuous and bounded. We denote $\Vert\mathbf{c}\rVert_\infty := \sup_i \sup_{(x,y)\in\mathcal{X}\times\mathcal{Y}} \lvert c_i(x,y)\rvert$. Let define $I^\lambda(\gamma):=\sum_i\lambda_i\int_{\mathcal{X}\times\mathcal{Y}}c_id\gamma_i$

Let $\gamma^*\in \Gamma^N_{\mu,\nu}$ such that $I^\lambda(\gamma^*) =\min_{\gamma\in\Gamma^N_{\mu,\nu}}I^\lambda(\gamma)$.  The existence of the minimum comes from the lower-semi continuity of $I^\lambda$ and  the compacity of $\Gamma^N_{\mu,\nu}$ for weak topology.

Let fix $\delta\in (0,1)$. $\mathcal{X}$ and $\mathcal{Y}$ are Polish spaces then $\exists  \mathcal{X}_0\subset \mathcal{X},  \mathcal{Y}_0\subset \mathcal{Y}$ compacts such that $\mu(\mathcal{X}_0^c) \leq  \delta$ and $\mu(\mathcal{Y}_0^c) \leq  \delta$.  It follows that $\forall i$, $\gamma^*_i((\mathcal{X}_0\times\mathcal{Y}_0)^c)\leq 2\delta$. Let define $\gamma^{*0}$ such that for all $i$, $\gamma^{*0}_{i}=\frac{\mathbf{1}_{\mathcal{X}_0\times\mathcal{Y}_0}}{\sum_i\gamma_i
^*(\mathcal{X}_0\times\mathcal{Y}_0)}\gamma^*_i$. We define $\mu_0 = \Pi_{1\sharp}\sum_i\gamma_{i}^{*0}$ and  $\nu_0 = \Pi_{2\sharp}\sum_i\gamma_{i}^{*0}$. We then naturally define $\Gamma^{N}_{0,\mu_0,\nu_0} :=\left\{(\gamma_i)_{1\leq i\leq N}\in \mathcal{M}_+(\mathcal{X}_0\times \mathcal{Y}_0)^N\text{ s.t. } \Pi_{1\sharp}\sum_i\gamma_i=\mu_0 \text{ and } \Pi_{2\sharp}\sum_i\gamma_i=\nu_0\right\}$ and $I^\lambda_0(\gamma_0) := \sum_i\lambda_i\int_{\mathcal{X}_0\times\mathcal{Y}_0}c_id\gamma_{0,i}$ for $\gamma_0\in\Gamma^{N}_{0,\mu_0,\nu_0}$. 

Let $\tilde{\gamma}_0$ verifying  $I^\lambda_0(\tilde{\gamma}_0) = \min_{\gamma_0\in\Pi^N_{0,\mu_0,\nu_0}}I^\lambda_0(\gamma_0)$. Let $\tilde{\gamma} = \left(\sum_i\gamma_i^*(\mathcal{X}_0\times\mathcal{Y}_0)\right)\tilde{\gamma}_0+\mathbf{1}_{(\mathcal{X}_0\times\mathcal{Y}_0)^c}\gamma^*\in \Gamma^N_{\mu,\nu}$. Then we get 
\begin{align*}
I^\lambda(\tilde{\gamma})\leq \min_{\gamma_0\in\Gamma^N_{0,\mu_0,\nu_0}}I^\lambda_0(\gamma_0)+2\sum|\lambda_i|\lVert \mathbf{c}\rVert_\infty\delta
\end{align*}

We have already proved that:

\begin{align*}
\sup\limits_{(f,g)\in\mathcal{F}^\lambda_{0,\mathbf{c}}}J_0^\lambda(f,g) = \inf\limits_{\gamma_0\in\Gamma^N_{0,\mu_0,\nu_0}}I_0^\lambda(\gamma_0)
\end{align*}

with $J_0^\lambda(f,g) = \int fd\mu_0+\int gd\nu_0$ and $\mathcal{F}^\lambda_{0,\mathbf{c}}$ is the set of $(f,g)\in\mathcal{C}^b(\mathcal{X}_0)\times \mathcal{C}^b(\mathcal{Y}_0)$ satisfying, for every $i$, $f\oplus g\leq\min_i\lambda_i c_i$. Let $(\tilde{f}_0,\tilde{g}_0)\in \mathcal{F}^\lambda_\mathbf{0,c}$ such that :
\begin{align*}
    J_0^\lambda(\tilde{f}_0,\tilde{g}_0)\geq \sup\limits_{(f,g)\in\mathcal{F}^\lambda_{0,\mathbf{c}}}J_0^\lambda(f,g)-\delta
\end{align*}
Since $J_0^\lambda(0,0)=0$, we get $\sup J_0^\lambda\geq 0$ and then,  $J_0^\lambda(\tilde{f}_0,\tilde{g}_0)\geq \delta\geq-1$. For every $\gamma_0 \in \Gamma^N_{0,\mu_0,\nu_0}$:

\begin{align*}
J_0^\lambda(\tilde{f}_0,\tilde{g}_0) = \int (\tilde{f}_0(x)+\tilde{g}_0(y))d\gamma_0(x,y)
\end{align*}
then we have the existence of $(x_0,y_0)\in\mathcal{X}_0\times\mathcal{Y}_0$ such that : $\tilde{f}_0(x_0)+\tilde{g}_0(y_0)\geq -1$. If we replace $(\tilde{f}_0,\tilde{g}_0)$ by $(\tilde{f}_0-s,\tilde{g}_0+s)$ for an accurate $s$, we get that: $\tilde{f}_0(x_0)\geq \frac12$
and $\tilde{g}_0(y_0)\geq \frac12$, and then $\forall(x,y)\in\mathcal{X}_0\times\mathcal{Y}_0$:
\begin{align*}
    \tilde{f}_0(x)\leq c'(x,y_0)-\tilde{g}_0(y_0)\leq c'(x,y_0)+\frac12\\
    \tilde{g}_0(y)\leq c'(x_0,y)-\tilde{f}_0(x_0)\leq c'(x_0,y)+\frac12
\end{align*}
where $c':=\min_i\lambda_ic_i$.  Let define $\bar{f}_0(x) = \inf_{y\in\mathcal{Y}_0}c'(x,y)-\tilde{g}_0(y)$ for $x\in\mathcal{X}$.  Then $\tilde{f}_0\leq \bar{f}_0$ on $\mathcal{X}_0$. We then get $J_0^\lambda(\bar{f}_0,\tilde{g}_0)\geq J_0^\lambda(\tilde{f}_0,\tilde{g}_0)$ and $\bar{f}_0\leq c'(.,y_0)+\frac12$ on $\mathcal{X}$. Let define $\bar{g}_0(y)= \inf_{x\in\mathcal{X}}c'(x,y)-\bar{f}_0(y)$. By construction $(f_0,g_0)\in \mathcal{F}
^\lambda_\mathbf{c}$ since the costs are uniformly continuous and bounded and $J_0^\lambda(\bar{f}_0,\bar{g}_0)\geq J_0^\lambda(\bar{f}_0,\tilde{g}_0)\geq J_0^\lambda(\tilde{f}_0,\tilde{g}_0)$. We also have  $\bar{g}_0\geq c'(x_0,.)+\frac12$ on $\mathcal{Y}$. Then we have in particular: $\bar{g}_0\geq -\lVert\mathbf{c}\rVert_\infty-\frac12$ on $\mathcal{X}$ and $\bar{f}_0\geq -\lVert\mathbf{c}\rVert_\infty-\frac12$ on $\mathcal{Y}$. Finally:

\begin{align*}
J^\lambda(\bar{f}_0,\bar{g}_0)&:=\int_{\mathcal{X}_0}\bar{f}d\mu_0+\int_{\mathcal{Y}_0}\bar{g}_0d\nu\\
&=\sum_i\gamma_i^*(\mathcal{X}_0\times\mathcal{Y}_0)\int_{\mathcal{X}_0\times\mathcal{Y}_0}(\bar{f}_0(x)+\bar{g}_0(y))d\left(\sum_i\gamma^{*0}_i(x,y)\right)\\
&+\int_{(\mathcal{X}_0\times\mathcal{Y}_0)^c}\bar{f}_0(x)+\bar{g}_0(y)d\left(\sum_i\gamma^{*}_i(x,y)\right)\\
&\geq (1-2\delta)\left(\int_{\mathcal{X}_0}\bar{f}_0d\mu_0+\int_{\mathcal{Y}_0}\bar{g}_0d\nu_0\right)-(2\lVert\mathbf{c}\rVert_\infty+1)\sum_i\gamma^*((\mathcal{X}_0\times\mathcal{Y}_0)^c)\\
&\geq (1-2\delta)J_0^\lambda(\bar{f}_0,\bar{g}_0)-2\sum |\lambda_i|(2\lVert\mathbf{c}\rVert_\infty+1)\delta\\
&\geq (1-2\delta)J_0^\lambda(\tilde{f}_0,\tilde{g}_0)-2\sum |\lambda_i|(2\lVert\mathbf{c}\rVert_\infty+1)\delta\\
&\geq (1-2\delta)(\inf I^\lambda_0-\delta)-2\sum |\lambda_i|(2\lVert\mathbf{c}\rVert_\infty+1)\delta\\
&\geq  (1-2\delta)(\inf I^\lambda-(2\sum |\lambda_i|\lVert\mathbf{c}\rVert_\infty+1)\delta)-2\sum |\lambda_i|(2\lVert\mathbf{c}\rVert_\infty+1)\delta
\end{align*}

This being true for arbitrary small $\delta$, we get $\sup J^\lambda\geq\inf I^\lambda$. The other sens is always true then:
\begin{align*}
\sup\limits_{(f,g)\in\mathcal{F}^\lambda_\mathbf{c}}\int fd\mu+\int gd\nu = \inf\limits_{\gamma\in\Gamma^N_{\mu,\nu}}\sum_i\lambda_i \int c_id\gamma_i
\end{align*}

for $c_i$ uniformly continuous and $\mathcal{X}$ and $\mathcal{Y}$ non necessarily compact.

\medskip
Let now prove that the result holds for lower semi-continuous costs. Let $\mathbf{c}:=(c_i)_i$ be a collection of lower semi-continuous costs. Let $(c^n_i)_n$ be non-decreasing sequences of bounded below cost functions such that $c_i=\sup_n c^n_i$. Let fix $\lambda\in\Delta^+_N$. From last step, we have shown that for all $n$:
\begin{align}
    \label{eq:eq_on_cont}
    \inf_{\gamma\in \Gamma^N_{\mu,\nu}} I^\lambda_n(\gamma)= \sup\limits_{(f,g)\in\mathcal{F}^\lambda_{\mathbf{c}^n}}\int fd\mu+\int gd\nu
\end{align}
where $I^\lambda_n(\gamma)=\sum_i\lambda_i \int c^n_id\gamma_i$. First it is clear that:
\begin{align}
\label{eq:ineq_sup}
\sup\limits_{(f,g)\in\mathcal{F}^\lambda_\mathbf{c}}\int fd\mu+\int gd\nu\leq \sup\limits_{(f,g)\in\mathcal{F}^{\lambda}_{\mathbf{c}^n}}\int fd\mu+\int gd\nu
\end{align}
Let show that:
\begin{align*}
\inf_{\gamma\in \Gamma^N_{\mu,\nu}} I^\lambda(\gamma)=\sup_n\inf_{\gamma\in \Gamma^N_{\mu,\nu}} I^\lambda_n(\gamma) = \lim_n\inf_{\gamma\in \Gamma^N_{\mu,\nu}} I^\lambda_n(\gamma)
\end{align*}
where $I^\lambda(\gamma) = \sum_i\lambda_i \int c_id\gamma_i$. 

Let $(\gamma^{n,k})_k$ a minimizing sequence of $\Gamma^N_{\mu,\nu}$ for the problem $\inf_{\gamma\in \Gamma^N_{\mu,\nu}} \sum_i\lambda_i \int c^n_id\gamma_i$. By Lemma~\ref{lem:compact-weak}, up to an extraction, there exists  $\gamma^n\in \Gamma^N_{\mu,\nu}$ such that $(\gamma^{n,k})_k$ converges weakly to $\gamma^n$. Then:
\begin{align*}
\inf_{\gamma\in \Gamma^N_{\mu,\nu}} I_n^\lambda(\gamma) =I^\lambda_n(\gamma^n)
\end{align*}
Up to an extraction, there also exists $\gamma^*\in\Gamma^N_{\mu,\nu}$ such that $\gamma^n$ converges weakly to $\gamma^*$. For $n\geq m$, $I^\lambda_n(\gamma^n) \geq I^\lambda_m(\gamma^n)\geq I^\lambda_m(\gamma^m) $, so by continuity of $I^\lambda_m$:
\begin{align*}
\lim_n I^\lambda_n(\gamma^n) \geq\limsup_n I^\lambda_m(\gamma^n)\geq I^\lambda_m(\gamma^*)
\end{align*}
By monotone convergence, $I^\lambda_m(\gamma^*)\rightarrow I^\lambda(\gamma^*)$ and $\lim_nI^\lambda_n(\gamma_n) \geq I^\lambda(\gamma^*)\geq\inf_{\gamma\in\Gamma^N_{\mu,\nu}}I^\lambda(\gamma)$.

Along with Eqs.~\ref{eq:eq_on_cont} and \ref{eq:ineq_sup}, we get that:
\begin{align*}
\inf_{\gamma\in \Gamma^N_{\mu,\nu}}I^\lambda(\gamma)\leq\sup\limits_{(f,g)\in\mathcal{F}^\lambda_\mathbf{c}}\int fd\mu+\int gd\nu
\end{align*}
The other sens being always true, we have then  shown that, in the general case we still have:
\begin{align*}
    \inf_{\gamma\in \Gamma^N_{\mu,\nu}}I^\lambda(\gamma)=\sup\limits_{(f,g)\in\mathcal{F}^\lambda_\mathbf{c}}\int fd\mu+\int gd\nu
\end{align*}
To conclude, we apply Lemma~\ref{lem:technical-lemma-primal}, and we get: 
\begin{align*}
\sup_{\lambda\in\Delta^+_N} \sup\limits_{(f,g)\in\mathcal{F}^\lambda_\mathbf{c}}\int fd\mu+\int gd\nu &=\sup_{\lambda\in\Delta^+_N} \inf_{\gamma\in \Gamma^N_{\mu,\nu}}I^\lambda(\gamma)\\
&= \MOT_\mathbf{c}(\mu,\nu)\\
\end{align*}

\end{prv*}

\subsection{Proof of Proposition~\ref{prop:optimality-cond}}
\label{prv:optimality-cond}
\begin{prv}
Let recall that, from standard optimal transport results:
\begin{align*}
    \MOT_\mathbf{c}(\mu,\nu) = \sup_{u\in\Phi_\mathbf{c}}\int u d\mu d\nu
\end{align*}
with $\Phi_\mathbf{c}:=\left\{u\in\mathcal{C}^b(\mathcal{\mathcal{X}\times\mathcal{Y}}) ~\mathrm{s.t.}~ \exists\lambda \in \Delta_N^+,~\exists\phi\in\mathcal{C}^b(\mathcal{X}),~ u =\phi^{cc}\oplus\phi^c~\mathrm{with}~c=\min_i\lambda_ic_i\right\}$ where $\phi^c$ is the $c$-transform of $\phi$, i.e. for $y\in\mathcal{Y}$, $\phi^c(y)=\inf_{x\in\mathcal{X}}c(x,y)-\phi(x)$. 

Let denote $\omega_1,\dots,\omega_N$ the continuity modulii of $c_1,...,c_N$. The existence of continuity modulii is ensured by the uniform continuity of $c_1,\dots,c_N$ on the compact sets $\mathcal{X}\times\mathcal{Y}$ (Heine's theorem). Then a modulus of continuity for $\min_i\lambda_ic_i$ is $\sum_i\lambda_i\omega_i$. As $\phi^c$ and $\phi^{cc}$ share the same modulus of continuity than $c=\min_i\lambda_ic_i$, for $u$ is $\Phi_\mathbf{c}$, a common modulus of continuity is $2\times\sum_i\omega_i$. More over, it is clear that for all $x,y$, $\{u(x,y)~\mathrm{s.t.}~u\in \Phi_c\}$ is compact. Then, applying Ascoli's theorem, we get, that $\Phi_{\mathbf{c}}$ is compact for $\lVert.\rVert_\infty$ norm. By continuity of $u\to\int u d\mu d\nu$, the supremum is attained, and we get the existence of the optimum $u^*$. The existence of optima $(\lambda^*,f^*,g^*)$ immediately follows.

Let first assume that $(\gamma_k)_{k=1}^N$ is a solution of Eq.~\eqref{eq-primal} and $(\lambda,f,g)$ is a solution of Eq.~\eqref{eq-dual}. Then it is clear that for all $i,j$, $f\oplus g\leq \lambda_ic_i$, $(\gamma_k)_{k=1}^N\in\Gamma^N_{\mu,\nu}$ and $\int c_jd\gamma_j=\int c_id\gamma_i$ (by Proposition~\ref{prop:mot-equality}). Let $k\in\{1,\dots,N\}$. Moreover, by Theorem~\ref{thm:duality-GOT}:

\begin{align*}
   0&=\int fd\mu +\int gd\nu -\int c_id\gamma_i\\
    & = \sum \int (f(x)+g(y)) d\gamma_i(x,y)-\sum_i\lambda_i\int c_i(x,y) d\gamma_i(x,y)\\
    & = \sum \int (f(x)+g(y)-\lambda_ic_i(x,y)) d\gamma_i(x,y)\\
\end{align*}

Since $f\oplus g\leq \lambda_ic_i$ and $\gamma_i$ are positive measures then $f\oplus g= \lambda_ic_i$, $\gamma_i$-almost everywhere.

Reciprocally, let assume that there exist $(\gamma_k)_{k=1}^N\in\Gamma^N_{\mu,\nu}$ and $(\lambda,f,g)\in \Delta_n^{+}\times\mathcal{C}^b(\mathcal{X})\times\mathcal{C}^b(\mathcal{Y})$ such that $\forall i\in\{1,...,N\},~ f\oplus g\leq\lambda_i c_i$,  $\forall i,j\in\{1,...,N\}~\int c_i d\gamma_i=\int c_j d\gamma_j$ and $f \oplus g= \lambda_i c_i ~~\gamma_{i}\text{-a.e.}$. Then, for any $k$:
\begin{align*}
\int c_k d\gamma_k &= \sum_i\lambda_i\int c_id\gamma_i\\
&= \sum_i\int(f(x)+g(y))d\gamma_i(x,y)\\
&= \int f(x)d\mu(x)+\int g(y)d\nu(y)\\
&\leq \MOT_{\mathbf{c}}(\mu,\nu)\text{ by Theorem~\ref{thm:duality-GOT}}
\end{align*}
then $\gamma_k$ is solution of the primal problem. We also have for any $k$:

\begin{align*}
\int fd\mu +\int gd\nu &= \sum_i\int(f(x)+g(y))d\gamma_i(x,y)\\
&= \sum_i\int\lambda_i c_id\gamma_i\\
&=\int c_k d\gamma_k\\
&\geq \MOT_{\mathbf{c}}(\mu,\nu)\\
\end{align*}
\end{prv}
then, thanks to Theorem~\ref{thm:duality-GOT}, $(\lambda,f,g)$ is solution of the dual problem.

Let now proof the result stated in Remark~\ref{rk:lambdanonzero}. Let assume the costs are strictly positive or strictly negative. If there exist $i$ such that $\lambda_i=0$, thanks to the condition $f\oplus g\leq \lambda_i c_i$, we get $f\oplus g\leq0$ and then $f\oplus g=0$ which contradicts the conditions $f\oplus g = \lambda_kc_k$ for all $k$.

\subsection{Proof of Proposition~\ref{prop:GOT-holder}}
\label{prv:GOT-holder}

Before proving the result let us first introduce the following lemma.

\begin{lemma}
\label{lem:sup_wasser}
Let $\mathcal{X}$ and $\mathcal{Y}$ be Polish spaces. Let $\mathbf{c}:=(c_i)_{1\leq i\leq N}$ a family of bounded below continuous costs. For $(x,y)\in \mathcal{X}\times \mathcal{Y}$ and $\lambda\in\Delta_N^{+}$, we define 
$$c_\lambda(x,y):=\min_{i=1,...,N}(\lambda_i c_i(x,y))$$
then for any $(\mu,\nu)\in\mathcal{M}_+^{1}(\mathcal{X})\times\mathcal{M}_+^{1}(\mathcal{Y})$  
\begin{align}
    \MOT_{\mathbf{c}}(\mu,\nu)=\sup_{\lambda\in\Delta_N^{+}} \wass_{c_\lambda}(\mu,\nu)
\end{align}
\end{lemma}

\begin{prv*}
Let $(\mu,\nu)\in\mathcal{M}_+^{1}(\mathcal{X})\times\mathcal{M}_+^{1}(\mathcal{Y})$ and $\mathbf{c}:=(c_i)_{1\leq i\leq N}$ cost functions on $\mathcal{X}\times \mathcal{Y}$. Let $\lambda\in\Delta_N^{+}$, then by Proposition~\ref{thm:duality-GOT}:
\begin{align*}
    \MOT_{\mathbf{c}}(\mu,\nu)= \sup_{\lambda\in\Delta_N^{+}} \sup_{(f,g)\in \mathcal{F}_{\mathbf{c}}^{\lambda}} \int_{\mathcal{X}} f(x)d\mu(x)+ \int_{\mathcal{Y}} g(y)d\nu(y)
\end{align*}
Therefore by denoting $c_\lambda:=\min_i(\lambda_ic_i)$ which is a continuous. The dual form of the classical Optimal Transport problem gives that:
\begin{align*}
\sup_{(f,g)\in \mathcal{F}_{\mathbf{c}}^{\lambda}} \int_{\mathcal{X}} f(x)d\mu(x)+ \int_{\mathcal{Y}} g(y)d\nu(y) =  \wass_{c_\lambda}(\mu,\nu)
\end{align*}
and the result follows.
\end{prv*}

Let us now prove the result of Proposition~\ref{prop:GOT-holder}. 

\begin{prv*} 
Let $\mu$ and $\nu$ be two probability measures. Let $\alpha\in (0,1]$. Note that if $d$ is a metric then $d^\alpha$ too. Therefore in the following we consider $d$ a general metric on $\mathcal{X}\times\mathcal{X}$. Let $c_1:(x,y)\rightarrow2\times \mathbf{1}_{x\neq y}$ and $c_2=d^{\alpha}$. For all $\lambda\in[0,1)$:
$$c_\lambda(x,y) := \min(\lambda c_1(x,y),(1-\lambda)c_2(x,y))=\min(2\lambda,(1-\lambda)d(x,y))$$
defines a distance on  $\mathcal{X}\times\mathcal{X}$. Then according to \cite[Theorem 1.14]{villani2003topics}: 
$$\wass_{c_\lambda}(\mu,\nu)=\sup_{f\text{ s.t. } f\text{ }1-c_\lambda\text{ Lipschitz}}\int fd\mu-\int fd\nu$$
Then thanks to Lemma~\ref{lem:sup_wasser} we have
$$\MOT_{(c_1,c_2)}(\mu,\nu) = \sup_{\lambda\in[0,1],f\text{ s.t. } f\text{ }1-c_\lambda\text{ Lipschitz}}\int fd\mu-\int fd\nu$$

Let now prove that in this case: $\MOT_{(c_1,c_2)}(\mu,\nu) = \beta_d(\mu,\nu)$. Let $\lambda \in [0,1)$ and $f$ a $c_\lambda$ Lipschitz function. $f$ is lower bounded: let $m = \inf f$ and $(u_n)_n$ a sequence satisfying $f(u_n)\rightarrow m$. Then for all $x,y$, $f(x)-f(y)\leq2\lambda$ and  $f(x)-f(y)\leq(1-\lambda)d(x,y)$. Let define $g=f-m-\lambda$. For $x$ fixed and for all $n$,  $f(x)-f(u_n)\leq2\lambda$, so taking the limit in $n$ we get $f(x)-m\leq2\lambda$.  So we get that for all $x,y$, $g(x)\in[-\lambda,+\lambda]$ and $g(x)-g(y)\in[-(1-\lambda)d(x,zy),(1-\lambda)d(x,y)]$. Then $||g||_\infty\leq \lambda$ and $||g||_d\leq 1-\lambda$. By construction, we also have $\int fd\mu-\int fd\nu=\int gd\mu-\int gd\nu$.Then $||g||_\infty+||g||_d\leq 1$. So we get that $\MOT_{(c_1,c_2)}(\mu,\nu) \leq \beta_d(\mu,\nu)$.\\
Reciprocally, let $g$ be a function satisfying $||g||_\infty+||g||_d\leq 1$. Let define $f=g+||g||_\infty$ and $\lambda = ||g||_\infty$. Then, for all $x,y$, $f(x)\in[0,2\lambda]$ and so $f(x)-f(y)\leq 2\lambda$. It is immediate that $f(x)-f(y)\in[-(1-\lambda)d(x,y),(1-\lambda)d(x,y)]$. Then we get $f(x)-f(y)\leq \min(\lambda,(1-\lambda)d(x,y))$. And by construction, we still have $\int fd\mu-\int fd\nu=\int gd\mu-\int gd\nu$. So $\MOT_{(c_1,c_2)}(\mu,\nu) \geq \beta_d(\mu,\nu)$.

\medskip

Finally we get $\MOT_{(c_1,c_2)}(\mu,\nu) = \beta_d(\mu,\nu)$ when $c_1:(x,y)\rightarrow2\times \mathbf{1}_{x\neq y}$ and $c_2=d$ a distance on $\mathcal{X}\times\mathcal{X}$.
\end{prv*}

\subsection{Proof of Proposition~\ref{prop:ineqharmonic}}
\label{prv:ineqharmonic}
\begin{lemma}
\label{lem:supmin}
Let $x_1,\dots,x_N\geq0$, then:
\begin{align*}
    \sup_{\lambda\in\Delta_N^{+}} \min_i\lambda_i x_i= \frac{1}{\sum_i\frac{1}{x_i}}
\end{align*}
\end{lemma}
\begin{prv*}First if there exists $i$ such that $x_i=0$, we immediately have  $\sup_{\lambda\in\Delta_N^{+}} \min_i\lambda_i x_i=0$.\\
$g:\lambda\mapsto \min_i\lambda_i x_i$ is a continuous function on the compact set $\lambda\in\Delta_N^{+}$. Let denote $\lambda^*$ the maximum of $g$. \\
Let show that for all $i,j$, $\lambda^*_ix_i=\lambda^*_jx_j$. Let denote $i_0,\dots,i_k$ the indices such that $\lambda^*_{i_l}x_{i_l}=\min_i\lambda^*_i x_i$. Let assume there exists $j_0$ such that: $\lambda^*_{j_0}x_{j_0}>\min_i\lambda^*_i x_i$, and that all  other indices $i$ have a larger $\lambda^*_i x_i\geq \lambda^*_{j_0}x_{j_0}$. Then for $\epsilon>0$ sufficiently small, let $\tilde{\lambda}$ defined as: $\tilde{\lambda}_{j_0}=\lambda
^*_{j_0}-\epsilon$, $\tilde{\lambda}_{i_l}=\lambda
^*_{i_l}+\epsilon/k$ for all $l\in\{1,\dots,k\}$ and $\tilde{\lambda}_{i}=\lambda
^*_{i}$ for all other indices. Then $\tilde{\lambda}\in\Delta_N^{+}$ and $g(\lambda^*)<g(\tilde{\lambda})$, which contradicts that $\lambda^*$ is the maximum.\\
Then at the optimum for all $i,j$, $\lambda^*_ix_i=\lambda^*_jx_j$. So $\lambda^*_ix_i=C$ for a certain constant $C$. Moreover $\sum_i\lambda^*_i=1$. Then $1/C=\sum_i1/x_i$. Finally, for all $i$, 
\begin{align*}
\lambda^*_i=\frac{1/x_i}{\sum_i1/x_i}   
\end{align*}
and then:
\begin{align*}
\sup_{\lambda\in\Delta_N^{+}} \min_i\lambda_i x_i= \frac{1}{\sum_i\frac{1}{x_i}}.
\end{align*}

\end{prv*}
\begin{prv*}
Let $\mu$ and $\nu$ be two probability measures respectively on $\mathcal{X}$ and $\mathcal{Y}$. Let $\mathbf{c}:=(c_i)_i$ be a family of cost functions. Let define for $\lambda\in\Delta_N^{+}$, $c_\lambda(x,y) := \min_i(\lambda_i c_i(x,y))$. We have, by linearity $\wass_{c_\lambda}(\mu,\nu)\leq \min_i(\lambda_i \wass_{c_i}(\mu,\nu))$. So we deduce by Lemma~\ref{lem:sup_wasser}:

\begin{align*}
    \MOT_{\mathbf{c}}(\mu,\nu)&=\sup_{\lambda\in\Delta_N^{+}} \wass_{c_\lambda}(\mu,\nu)\\
    &\leq\sup_{\lambda\in\Delta_N^{+}} \min_i\lambda_i \wass_{c_i}(\mu,\nu)\\
    &= \frac{1}{\sum_i\frac{1}{\wass_{c_i}(\mu,\nu)}}\text{ by Lemma~\ref{lem:supmin}}\\ 
\end{align*}
which concludes the proof.
\end{prv*}

\subsection{Proof of Theorem~\ref{thm:duality-entropic}}
\label{prv:duality-entropic}

\begin{prv*}
To show the strong duality of the regularized problem, we use the same sketch of proof as for the strong duality of the original problem. 
\medskip
Let first assume that, for all $i$, $c_i$ is continuous on the compact set $\mathcal{X}\times\mathcal{Y}$. Let fix $\lambda\in\Delta^+_N$. We define, for all $u\in\mathcal{C}^b(\mathcal{X}\times\mathcal{Y})$:
\begin{align*}
    V^\lambda_i(u) = \varepsilon_i\left(\int_{(x,y)\in\mathcal{X}\times\mathcal{Y}} \exp{\frac{-u(x,y)-\lambda_ic_i(x,y)}{\varepsilon_i}}d\mu(x)d\nu(y)-1\right)
\end{align*}
and:
\begin{align*}
E(u)=\left\{\begin{matrix} \int fd\mu+\int gd\nu &\quad\text{if}\quad& \exists (f,g)\in \mathcal{C}^b(\mathcal{X})\times\mathcal{C}^b(\mathcal{Y}),~ u = f+g\\
+\infty &\quad\text{else}\quad&\end{matrix}\right.
\end{align*}
Let compute the Fenchel-Legendre transform of these functions. Let $\gamma\in\mathcal{M}(\mathcal{X}\times\mathcal{Y})$:
\begin{align*}
    V^{\lambda*}_i(-\gamma) = \sup_{u\in\mathcal{C}^b(\mathcal{X}\times\mathcal{Y})}-\int ud\gamma -\varepsilon_i\left(\int_{(x,y)\in\mathcal{X}\times\mathcal{Y}} \exp{\frac{-u(x,y)-\lambda_ic_i(x,y)}{\varepsilon_i}}d\mu(x)d\nu(y)-1\right) 
\end{align*}
However, by density of $\mathcal{C}^b(\mathcal{X}\times\mathcal{Y})$ in $L^1_{d\mu\otimes\nu}(\mathcal{X}\times\mathcal{Y})$, the set of integrable functions for $\mu\otimes\nu$ measure, we deduce that
\begin{align*}
    V^{\lambda*}_i(-\gamma) = \sup_{u\in L^1_{d\mu\otimes\nu}(\mathcal{X}\times\mathcal{Y})}-\int ud\gamma -\varepsilon_i\left(\int_{(x,y)\in\mathcal{X}\times\mathcal{Y}} \exp{\frac{-u(x,y)-\lambda_ic_i(x,y)}{\varepsilon_i}}d\mu(x)d\nu(y)-1\right) 
\end{align*}
This supremum equals $+\infty$ if $\gamma$ is not positive and not absolutely continuous with regard to $\mu\otimes \nu$. Let us now denote 
$F_{\gamma,\lambda}(u):=-\int ud\gamma -\varepsilon_i\left(\int_{(x,y)\in\mathcal{X}\times\mathcal{Y}} \exp{\frac{-u(x,y)-\lambda_ic_i(x,y)}{\varepsilon_i}}d\mu(x)d\nu(y)-1\right).$
$F_{\gamma,\lambda_*}$ is Fréchet differentiable and its maximum is attained for 
$u^*=\varepsilon_i \log\left(\frac{d\gamma}{d\mu\otimes\nu}\right)+\lambda_i c_i$. Therefore we obtain that
\begin{align*}
 V^{\lambda*}_i(-\gamma)&=\varepsilon_i\left(\int \log\left(\frac{d\gamma}{d\mu\otimes\nu}\right)d\gamma +1- \gamma(\mathcal{X}\times\mathcal{Y})\right)+\lambda_i\int c_i d\gamma\\
 &=\lambda_i\int c_id\gamma+\varepsilon_i\KL(\gamma_i||\mu\times\nu)
\end{align*}   
Thanks to the compactness of $\mathcal{X}\times\mathcal{Y}$, all the $V_i^{\lambda}$ for $i\in\{1,...,N\}$ are continuous on $\mathcal{C}^b(\mathcal{X}\times\mathcal{Y})$. Therefore by applying Lemma~\ref{lem:rockafellar-gene}, we obtain that:

\begin{align*}
\inf_{u\in \mathcal{C}^b(\mathcal{X}\times\mathcal{Y})} \sum_i V_i^{\lambda}(u) + E(u) = \sup\limits_{\substack{\gamma_1...,\gamma_N,\gamma\in \mathcal{M}(\mathcal{X}\times\mathcal{Y})\\\sum_i \gamma_i = \gamma}}-\sum_i V_i^{\lambda*}(\gamma_i)-E^*(-\gamma)
\end{align*}
\begin{align*}
\sup\limits_{f\in \mathcal{C}^b(\mathcal{X}),~g\in\mathcal{C}^b(\mathcal{Y})}&\int fd\mu+\int gd\nu \\
& - \sum_{i=1}^N \varepsilon_i\left(\int_{(x,y)\in\mathcal{X}\times\mathcal{Y}} \exp{\frac{f(x)+g(y)-\lambda_ic_i(x,y)}{\varepsilon_i}}d\mu(x)d\nu(y)-1\right)\\
&= \inf_{\gamma\in\Gamma^N_{\mu,\nu}} \sum_{i=1}^N \lambda_i\int c_i d\gamma_i + \varepsilon_i\KL(\gamma_i||\mu\otimes\nu)
\end{align*}
Therefore by considering the supremum over the $\lambda\in\Delta_N$, we obtain that
\begin{align*}
 \sup_{\lambda\in\Delta^+_N}  \sup\limits_{f\in \mathcal{C}^b(\mathcal{X}),~g\in\mathcal{C}^b(\mathcal{Y})}&\int fd\mu+\int gd\nu \\
& - \sum_{i=1}^N \varepsilon_i\left(\int_{(x,y)\in\mathcal{X}\times\mathcal{Y}} \exp{\frac{f(x)+g(y)-\lambda_ic_i(x,y)}{\varepsilon_i}}d\mu(x)d\nu(y)-1\right)\\
   &=\sup_{\lambda\in\Delta^+_N}\inf_{\gamma\in\Gamma^N_{\mu,\nu}} \sum_{i=1}^N \lambda_i\int c_i d\gamma_i + \varepsilon_i \KL(\gamma_i||\mu\otimes\nu)
\end{align*}

Let $f: (\lambda,\gamma)\in\Delta^+_N\times\Gamma^N_{\mu,\nu}\mapsto\sum_{i=1}^N \lambda_i\int c_i d\gamma_i + \varepsilon_i \KL(\gamma_i||\mu\otimes\nu)$. $f$ is clearly concave and continuous in $\lambda$. Moreover $\gamma\mapsto \KL(\gamma_i||\mu\otimes\nu)$ is convex and lower semi-continuous for weak topology~\citep[Lemma 1.4.3]{dupuis2011weak}. Hence $f$ is convex and lower-semi continuous in $\gamma$. $\Delta^+_N$ is  convex, and  $\Gamma^N_{\mu,\nu}$ is compact for weak topology (see  Lemma
~\ref{lem:compact-weak}). So by Sion's theorem,  we get the expected  result:
\begin{align*}
\min_{\gamma\in\Gamma^N_{\mu,\nu}} \sup_{\lambda\in\Delta^+_N}\sum_{i}&\lambda_i \int c_i d\gamma_i + \sum_i\varepsilon_i \KL(\gamma_i||\mu\otimes\nu)\\
&=\sup_{\lambda\in\Delta^+_N} \sup_{(f,g)\in\mathcal{C}_b(\mathcal{X})\times\mathcal{C}_b(\mathcal{Y})}\int_{\mathcal{X}} f(x)d\mu(x)+ \int_{\mathcal{Y}} g(y)d\nu(y)\\
&-\sum_{i=1}^N\varepsilon_i\left( \int_{\mathcal{X}\times\mathcal{Y}} e^{\frac{f(x)+g(y)-\lambda_ic_i(x,y)}{\varepsilon_i}} d\mu(x)d\nu(y)-1\right)
\end{align*}
Moreove by fixing $\gamma\in\Gamma^N_{\mu,\nu}$, we have
\begin{align*}
\sup_{\lambda\in\Delta^+_N}\sum_{i}&\lambda_i \int c_i d\gamma_i + \sum_i\varepsilon_i \KL(\gamma_i||\mu\otimes\nu)\\
&=\max_{i}\int c_i d\gamma_i + \sum_j\varepsilon_j \KL(\gamma_j||\mu\otimes\nu)\\
\end{align*}
which concludes the proof in case of continuous costs. A similar proof as the one of the Theorem~\ref{thm:duality-entropic} allows to extend the results for lower semi-continuous cost functions.
\end{prv*}

\newpage 

\section{Discrete cases}
\subsection{Exact discrete case}
\label{dis:exact}

Let $a\in\Delta_N^{+}$ and $b\in\Delta^+_m$ and $\mathbf{C}:=(C_i)_{1\leq i\leq N}\in\left(\mathbb{R}^{n\times m}\right)^N$ be $N$ cost matrices. Let also $\mathbf{X}:=\{x_1,...,x_n\}$ and $\mathbf{Y}:=\{y_1,...,y_m\}$ two subset of $\mathcal{X}$ and $\mathcal{Y}$ respectively. Moreover we define the two following discrete measure $\mu=\sum_{i=1}^n a_i \delta_{x_i}$ and $\nu=\sum_{i=1}^n b_i \delta_{y_i}$ and for all $i$, $C_i = (c_i(x_k,y_l))_{1\leq k\leq n,1\leq l\leq m}$ where $(c_i)_{i=1}^N$ a family of cost functions. The discretized multiple cost optimal transport primal problem can be written as follows:
\begin{align*}
\MOT_{\mathbf{c}}(\mu,\nu)=\widehat{\MOT}_{\textbf{C}}(a,b) := \inf_{P\in\Gamma_{a,b}^N} \max_i~\langle P_i,C_i\rangle
\end{align*}
where $\Gamma_{a,b}^N:=\left\{(P_i)_{1\leq i\leq N}\in\left(\mathbb{R}_+^{n\times m}\right)^N\text{ s.t. } (\sum_i P_i)\mathbf{1}_m=a \text{ and } (\sum_i P_i^T)\mathbf{1}_n=b \right\}$. 
As in the continuous case, strong duality holds and we can rewrite the dual in the discrete case also.
\begin{prop}[Duality for the discrete problem]
\label{prop:discrete-dual}

Let $a\in\Delta_N^{+}$ and $b\in\Delta^+_m$ and $\mathbf{C}:=(C_i)_{1\leq i\leq N}\in\left(\mathbb{R}^{n\times m}\right)^N$ be $N$ cost matrices. Strong duality holds for the discrete problem and
\begin{align*}
\widehat{\MOT}_{\mathbf{C}}(a,b)=\sup_{\lambda\in\Delta^+_N}\sup\limits_{(f,g)\in\mathcal{F}^{\lambda}_{\mathbf{C}}} \langle f,a\rangle+\langle g,b\rangle.
\end{align*}
where $\mathcal{F}^{\lambda}_{\mathbf{C}}:=\{(f,g)\in\mathbb{R}_{+}^n\times\mathbb{R}_{+}^m\text{ s.t. }\forall i\in\{1,...,N\},~ f\mathbf{1}_m^T+\mathbf{1}_n g^T\leq\lambda_i C_i\}$.
\end{prop}

\subsection{Entropic regularized discrete case}
\label{dis:entropic}

We now extend the regularization in the discrete case. 
Let $a\in\Delta_n^{+}$ and $b\in\Delta^+_m$ and $\mathbf{C}:=(C_i)_{1\leq i\leq N}\in\left(\mathbb{R}^{n\times m}\right)^N$ be $N$ cost matrices and $\bm{\varepsilon}=(\varepsilon_i)_{1\leq i\leq N}$ be nonnegative real numbers. The discretized regularized primal problem is:
\begin{align*}
    \widehat{\MOT}^{\bm{\varepsilon}}_{\mathbf{C}}(a,b)=\inf_{P\in \Gamma_{a,b}^N} \max_i \langle P_i,C_i\rangle -\sum_{i=1}^N\varepsilon_i \ent(P_i)
\end{align*}
where $\ent(P) = \sum_{i,j}P_{i,j}(\log P_{i,j}-1)$ for $P=(P_{i,j})_{i,j}\in \mathbb{R}_+^{n\times m}$ is the discrete entropy. In the discrete case, strong duality holds thanks to Lagrangian duality and Slater sufficient conditions:
 
\begin{prop}[Duality for the discrete regularized problem]
\label{prop:discrete-reg-dual}
Let $a\in\Delta_n^{+}$ and $b\in\Delta^+_m$ and $\mathbf{C}:=(C_i)_{1\leq i\leq N}\in\left(\mathbb{R}^{n\times m}\right)^N$ be $N$ cost matrices and $\bm{\varepsilon}:=(\varepsilon_i)_{1\leq i\leq N}$ be non negative reals. Strong duality holds and by denoting $K_i^{\lambda_i} =\exp\left(-\lambda_i C_{i}/\varepsilon_i\right)$, we have
\begin{align*}
\widehat{\MOT}_{\mathbf{C}}^{\bm{\varepsilon}}(a,b)=\sup_{\lambda\in\Delta^+_N}\sup\limits_{f\in \mathbb{R}^n,~g\in \mathbb{R}^m} \langle f, a\rangle+\langle g,b\rangle-\sum_{i=1}^N\varepsilon_i\langle e^{\mathbf{f}/\varepsilon_i},K_i^{\lambda_i} e^{\mathbf{g}/\varepsilon_i}\rangle.
\end{align*}
\end{prop}
The objective function for the dual problem is strictly concave in $(\lambda,f,g)$ but is neither smooth or strongly convex. 


\begin{prv*}
The proofs in the discrete case are simpler and only involves Lagrangian duality~\citep[Chapter 5]{boyd2004convex}. Let do the  proof in the regularized case, the one for the standard problem follows exactly the same path.

Let $a\in\Delta_N^{+}$ and $b\in\Delta^+_m$ and $\mathbf{C}:=(C_i)_{1\leq i\leq N}\in\left(\mathbb{R}^{n\times m}\right)^N$ be $N$ cost matrices. 
\begin{align*}
    \widehat{\MOT}^{\bm{\varepsilon}}_{\mathbf{C}}(a,b)&=\inf_{P\in \Gamma_{a,b}^N} \max_{1\leq i\leq N}\langle P_i,C_i\rangle -\sum_{i=1}^N\varepsilon_i \ent(P_i) \\
    &=\inf\limits_{\substack{(t,P)\in \mathbb{R}\times\left(\mathbb{R}_+^{n\times m}\right)^N\\(\sum_i P_i)\mathbf{1}_m=a\\
     (\sum_i P_i^T)\mathbf{1}_n=b\\
    \forall j,~\langle P_j,C_j\rangle \leq t}}t -\sum_{i=1}^N\varepsilon_i \ent(P_i)\\
    &=\inf\limits_{\substack{(t,P)\in \mathbb{R}\times\left(\mathbb{R}_+^{n\times m}\right)^N}}\sup\limits_{\substack{f\in\mathbb{R}^n,~g\in\mathbb{R}^m,~\lambda\in\mathbb{R}_+^N}} t+\sum_{j=1}^N\lambda_j(\langle P_j,C_j\rangle- t) -\sum_{i=1}^N\varepsilon_i \ent(P_i)\\
    &+f^T\left(a-\sum_i P_i\mathbf{1}_m\right)+g^T\left(b-\sum_i P_i^T\mathbf{1}_n\right)\\
\end{align*}
The constraints are qualified for this convex problem, hence by Slater's sufficient condition~\citep[Section 5.2.3]{boyd2004convex}, strong duality holds and:
\begin{align*}
    \widehat{\MOT}^{\bm{\varepsilon}}_{\mathbf{C}}(a,b)
    &=\sup\limits_{\substack{f\in\mathbb{R}^n,~g\in\mathbb{R}^m,~\lambda\in\mathbb{R}_+^N}}\inf\limits_{\substack{(t,P)\in \mathbb{R}\times\left(\mathbb{R}_+^{n\times m}\right)^N}} t+\sum_{j=1}^N\lambda_j(\langle P_j,C_j\rangle- t) -\sum_{j=1}^N\varepsilon_j \ent(P_j)\\
    &+f^T\left(a-\sum_{j=1}^N P_i\mathbf{1}_m\right)+g^T\left(b-\sum_{j=1}^N P_i^T\mathbf{1}_n\right)\\
    & = \sup\limits_{\substack{f\in\mathbb{R}^n\\g\in\mathbb{R}^m\\\lambda\in\Delta_N^{+}}} \langle f,a \rangle + \langle g, b \rangle + \sum_{j=1}^N\inf_{P_j\in\mathbb{R}_+^{n\times m}}\left(\langle P_j,\lambda_jC_j-f\mathbf{1}_n^T - \mathbf{1}_m g^T\rangle -\varepsilon_j \ent(P_j) \right)
\end{align*}
But for every $i=1,..,N$ the solution of 
\begin{align*}
    \inf_{P_j\in\mathbb{R}_+^{n\times m}}\left(\langle P_j,\lambda_jC_j-f\mathbf{1}_n^T - \mathbf{1}_m g^T\rangle -\varepsilon_j \ent(P_j)\right)
\end{align*}
is
\begin{align*}
  P_j = \exp\left(\frac{f\mathbf{1}_n^T + \mathbf{1}_m g^T-\lambda_j C_j}{\varepsilon_i}\right)
\end{align*}
Finally we obtain that
\begin{align*}
    \widehat{\MOT}^{\bm{\varepsilon}}_{\mathbf{C}}(a,b)
    &= \sup\limits_{\substack{f\in\mathbb{R}^n,~g\in\mathbb{R}^m,~\lambda\in\Delta_N^{+}}} \langle f,a \rangle + \langle g, b \rangle  - \sum_{k=1}^N \varepsilon_k \sum_{i,j} \exp\left(\frac{f_i+g_j-\lambda_k C_k^{i,j}}{\varepsilon_k}\right)
\end{align*}
\end{prv*}


\newpage 
\section{Other results}

\subsection{Utilitarian and Optimal Transport}
\label{res:min-sum}

\begin{prop}
Let $\mathcal{X}$ and $\mathcal{Y}$ be Polish spaces. Let $\mathbf{c}:=(c_i)_{1\leq i\leq N}$ be a family of bounded below continuous cost functions on $\mathcal{X}\times \mathcal{Y}$, and $\mu\in\mathcal{M}^1_+(\mathcal{X})$ and  $\nu\in\mathcal{M}^1_+(\mathcal{Y})$. Then we have:
\begin{align}
  \inf_{\substack{(\gamma_i)_{i=1}^N\in\Gamma^N_{\mu,\nu}\
}}  \sum_i \int c_i d\gamma_i = \wass_{\min_i(c_i)}(\mu,\nu)
\end{align}
\end{prop}

\begin{prv*}
The proof is a by-product of the proof of Theorem~\ref{thm:duality-GOT}. The continuity of the costs is necessary since $\min_i(c_i)$ is not necessarily lower semi-continuous when the costs are supposed lower semi-continuous.
\end{prv*}

\begin{rmq} 
We thank an anonymous reviewer for noticing that the utilitarian problem can be written also as an Optimal Transport on the space  $\mathcal{Z} = (\mathcal{X}\times\{1,\dots,N\})\times(\mathcal{Y}\times\{1,\dots,N\})$:
\begin{align*}
    \min_{\gamma\in\Tilde{\Gamma}_{\mu,\nu}}\int_{x,i,y,j} c((x,i),(y,j))d\gamma(x,i,y,j) 
\end{align*}
where the constraint space is  $\Tilde{\Gamma}_{\mu,\nu}:=\left\{\gamma\in\mathcal{M}_1^+(\mathcal{Z})\text{ s.t. }\Pi_{\mathcal{X}}\gamma = \mu,~\Pi_{\mathcal{Y}}\gamma = \nu\right\}$.
\end{rmq}

\subsection{MOT generalizes OT}
\label{res:MOT-gene}

\begin{prop}
\label{prop:gene-GOT}
Let $\mathcal{X}$ and $\mathcal{Y}$ be Polish spaces. Let $N\geq 0$, $\mathbf{c}=(c_i)_{1\leq i\leq N}$ be a family of nonnegative lower semi-continuous costs and let us denote for all $k\in\{1,\dots,N\}$, $\mathbf{c}_k=(c_i)_{1\leq i\leq k}$. Then for all $k\in\{1,\dots,N\}$, there exists a family of costs $\mathbf{d}_k\in\text{LSC}(\mathcal{X}\times\mathcal{Y})^N$ such that  
\begin{align}
    \MOT_{\mathbf{d}_k}(\mu,\nu) = \MOT_{\mathbf{c}_k}(\mu,\nu)
\end{align}
\end{prop}
\begin{prv*}
For all $k\in\{1,...,N\}$, we define $\mathbf{d}_k:=(c_1,...,(N-k+1)\times c_k,...,(N-k+1)\times c_k)$. Therefore, thanks to Lemma \ref{lem:sup_wasser} we have
\begin{align}
\MOT_{\mathbf{d}_k}(\mu,\nu)& = \sup_{\lambda\in\Delta_N^{+}} \wass_{c_\lambda}(\mu,\nu) \\
& = \sup_{(\lambda,\gamma)\in\Delta^k_n} \inf_{\gamma\in\Gamma_{\mu,\nu}}\int_{\mathcal{X}\times\mathcal{Y}} \min(\lambda_1 c_1,..,\lambda_{k-1}c_{k-1},\lambda_k c_k) d\gamma
\end{align}
where $\Delta^k_n:=\{(\lambda,\gamma)\in \Delta_N^{+}\times\mathbb{R}_{+}\text{:\quad } \gamma=(N-k+1)\times\min(\lambda_k,...,\lambda_N)\}$.
First remarks that
\begin{align}
    \gamma = 1 - \sum_{i=1}^{k-1} \lambda_i &\iff (N-k+1)\times\min(\lambda_k,...,\lambda_N) = \sum_{i=k}^{N} \lambda_i \\
    &\iff \lambda_k=...=\lambda_N
\end{align}
But in that case $(\lambda_1,...,\lambda_{k-1},\gamma)\in\Delta_k$ and therefore we obtain that 
\begin{align*}
     \MOT_{\mathbf{d}_k}(\mu,\nu) \geq \sup_{\lambda\in\Delta_k}\inf_{\gamma\in\Gamma_{\mu,\nu}}  \int_{\mathcal{X}\times\mathcal{Y}} \min(\lambda_1 c_1,..,\lambda_{k-1}c_{k-1},\gamma c_k) d\gamma = \MOT_{\mathbf{c}_k}(\mu,\nu) 
\end{align*}
Finally by definition we have  $\gamma\leq \sum_{i=k}^{N} \lambda_i = 1 -  \sum_{i=1}^{k-1} \lambda_i $ and therefore
\begin{align*}
 \int_{\mathcal{X}\times\mathcal{Y}} \min(\lambda_1 c_1,..,\lambda_{k-1}c_{k-1},\gamma c_k) d\gamma \leq  \int_{\mathcal{X}\times\mathcal{Y}} \min\left(\lambda_1 c_1,..,\lambda_{k-1}c_{k-1},\left(1 -  \sum_{i=1}^{k-1} \lambda_i\right) c_k\right) 
\end{align*}
Then we obtain that 
\begin{align*}
     \MOT_{\mathbf{d}_k}(\mu,\nu)\leq  \MOT_{\mathbf{c}_k}(\mu,\nu) 
\end{align*}
and the result follows.
\end{prv*}

\begin{prop}
Let $\mathcal{X}$ and $\mathcal{Y}$ be Polish spaces and $\mathbf{c}:=(c_i)_{1\leq i\leq N}$ a family of nonnegative lower semi-continuous costs on $\mathcal{X}\times \mathcal{Y}$. We suppose that, for all $i$, $c_i= N\times c_1$. Then for any $(\mu,\nu)\in\mathcal{M}_+^{1}(\mathcal{X})\times\mathcal{M}_+^{1}(\mathcal{Y})$  
\begin{align}
    \MOT_{\mathbf{c}}(\mu,\nu)=\MOT_{c_1}(\mu,\nu)=\wass_{c_1}(\mu,\nu).
\end{align}
\end{prop}
\begin{prv*}
Let $c:=(c_i)_{1\leq i\leq N}$ such that for all $i$, $c_i=c_1$. for all $(x,y)\in\mathcal{X}\times \mathcal{Y}$ and $\lambda\in\Delta^+_N$, we have:
\begin{align*}
    c_\lambda(x,y):=\min_i(\lambda_i c_i(x,y)) = \min_i(\lambda_i)c_1(x,y)
\end{align*}
Therefore we obtain from Lemma~\ref{lem:sup_wasser} that
\begin{align}
    \MOT_{c}(\mu,\nu)=\sup_{\lambda\in\Delta^+_N} \wass_{c_\lambda}(\mu,\nu)
\end{align}
But we also have that:
\begin{align*}
    \wass_{c_\lambda}(\mu,\nu)&=\inf_{\gamma\in\Gamma(\mu,\nu)}\int_{\mathcal{X}\times \mathcal{Y}} \min_i(\lambda_i c_i(x,y))d\gamma(x,y)\\
    &=\min_i(\lambda_i )\inf_{\gamma\in\Gamma(\mu,\nu)}\int_{\mathcal{X}\times \mathcal{Y}} c_1(x,y) d\gamma(x,y)\\
    &=\min_i(\lambda_i )  \wass_{c_1}(\mu,\nu)
\end{align*}
Finally by taking the supremum over $\lambda\in\Delta^+_N$ we conclude the proof.
\end{prv*}

\subsection{Regularized EOT tends to EOT}
\label{res:epsto0}

\begin{prop}
\label{prop:epsto0}
For $(\mu,\nu)\in\mathcal{M}_+^{1}(\mathcal{X})\times\mathcal{M}_+^{1}(\mathcal{Y})$ we have $  \lim\limits_{\bm{\varepsilon}\to0} \MOT_{\mathbf{c}}^{\bm{\varepsilon}}(\mu,\nu) = \MOT_{\mathbf{c}}(\mu,\nu)$.
\end{prop}

\begin{prv*}
Let $(\bm{\varepsilon}_l=(\varepsilon_{l,1},\dots,\varepsilon_{l,N}))_l$ a sequence converging to $0$. Let $\gamma_l = (\gamma_{l,1},\dots,\gamma_{l,N})$ be the optimum of $\MOT^{\bm{\varepsilon_l}}_\mathbf{c}(\mu,\nu)$. By Lemma~\ref{lem:compact-weak}, up to an extraction, $\gamma_l\rightarrow \gamma^\star=(\gamma^\star_{1},\dots,\gamma^\star_{N})\in\Gamma^N_{\mu,\nu}$. Let now $\gamma=(\gamma_{1},\dots,\gamma_{N})$ be the optimum of $\MOT_\mathbf{c}(\mu,\nu)$. By optimality of $\gamma$ and $\gamma_l$, for all $i$: \begin{align*}
    0\leq  \int c_id\gamma_{l,i}-\int c_id\gamma_{i}\leq\sum_i\varepsilon_{l,i}\left(\KL(\gamma_{i}||\mu\otimes\nu)-\KL(\gamma_{l,i}||\mu\otimes\nu)\right)
\end{align*}
By lower semi continuity of $\KL(.||\mu\otimes\nu)$ and by taking the limit inferior as $l\to\infty$, we get for all $i$, $\liminf_{\ell\rightarrow\infty} \int c_id\gamma_{l,i}=\int c_id\gamma_{i}$. Moreover by continuity of $\gamma\rightarrow \int c_i d\gamma_i$  we therefore obtain that for all $i$, $\int c_id\gamma^\star_{i}\leq \int c_id\gamma_{i}$. Then by optimality of $\gamma$ the result follows.
\end{prv*}

\subsection{Projected Accelerated Gradient Descent}
\label{res:pgd}

\begin{prop}
\label{prop:algo-dual}
Let $a\in\Delta_N^{+}$ and $b\in\Delta^+_m$ and $\mathbf{C}:=(C_i)_{1\leq i\leq N}\in\left(\mathbb{R}^{n\times m}\right)^N$ be $N$ cost matrices and $\bm{\varepsilon}:=(\varepsilon,...,\varepsilon)$ where $\varepsilon>0$. Then by denoting $K_i^{\lambda_i} =\exp\left(-\lambda_i C_{i}/\varepsilon\right)$, we have
\begin{align*}
\widehat{\MOT}_{\mathbf{C}}^{\bm{\varepsilon}}(a,b)=\sup_{\lambda\in\Delta^+_N}\sup\limits_{f\in \mathbb{R}^n,~g\in \mathbb{R}^m} F_{\mathbf{C}}^{\varepsilon}(\lambda,f,g):= \langle f, a\rangle+\langle g,b\rangle -\varepsilon\left[\log\left(\sum_{i=1}^N\langle e^{\mathbf{f}/\varepsilon},K_i^{\lambda_i} e^{\mathbf{g}/\varepsilon}\rangle \right) + 1\right].
\end{align*}
Moreover, $F_{\mathbf{C}}^{\varepsilon}$ is concave, differentiable and $\nabla F$ is $\frac{\max\left(\max\limits_{1\leq i\leq N}\Vert C_i\Vert_{\infty}^2,2N\right)}{\varepsilon}$ Lipschitz-continuous on $\mathbb{R}^N\times \mathbb{R}^n \times\mathbb{R}^m$.
\end{prop}

\begin{prv*}
Let $\mathcal{Q}:=\left\{P:=(P_1,...,P_N)\in(\mathbb{R}_{+}^{n\times m})^N \text{:\quad} \sum_{k=1}^N \sum_{i,j} P_k^{i,j}=1 \right\}$. Note that $\Gamma_{a,b}^N\subset \mathcal{Q}$, therefore from the primal formulation of the problem we have that 
\begin{align*}
    \widehat{\MOT}^{\bm{\varepsilon}}_{\mathbf{C}}(a,b)&=\sup_{\lambda\in\Delta_N^{+}}\inf_{P\in \Gamma_{a,b}^N}  \sum_{i=1}^N \lambda_i \langle P_i,C_i\rangle -\varepsilon \ent(P_i) \\
   &= \sup_{\lambda\in\Delta_N^{+}} \inf_{P\in\mathcal{Q}}  \sup_{f\in\mathbb{R}^n,~g\in\mathbb{R}^m} \sum_{i=1}^N \lambda_i \langle P_i,C_i\rangle -\varepsilon \ent(P_i)\\
   & +f^T\left(a-\sum_i P_i\mathbf{1}_m\right)+g^T\left(b-\sum_i P_i^T\mathbf{1}_n\right)\\
\end{align*}
The constraints are qualified for this convex problem, hence by Slater's sufficient condition~\citep[Section 5.2.3]{boyd2004convex}, strong duality holds. Therefore we have
\begin{align*}
    \widehat{\MOT}^{\bm{\varepsilon}}_{\mathbf{C}}(a,b)&=\sup_{\lambda\in\Delta_N^{+}}\sup_{f\in\mathbb{R}^n,~g\in\mathbb{R}^m} \inf_{P\in \mathcal{Q}} \sum_{i=1}^N \lambda_i \langle P_i,C_i\rangle -\varepsilon \ent(P_i) \\
    &+ f^T\left(a-\sum_i P_i\mathbf{1}_m\right)+g^T\left(b-\sum_i P_i^T\mathbf{1}_n\right)\\
   &= \sup_{\lambda\in\Delta_N^{+}} \sup_{f\in\mathbb{R}^n,~g\in\mathbb{R}^m} \langle f,a \rangle + \langle g, b \rangle \\
   &+ \inf_{P\in\mathcal{Q}}
   \sum_{k=1}^N \sum_{i,j} P_k^{i,j} \left(\lambda_k C_k^{i,j} +\varepsilon\left(\log(P_k^{i,j})-1\right) -f_i - g_j \right)  \\
\end{align*}
Let us now focus on the following problem:
\begin{align*}
\inf_{P\in\mathcal{Q}}
   \sum_{k=1}^N \sum_{i,j} P_k^{i,j} \left(\lambda_k C_k^{i,j} +\varepsilon\left(\log(P_k^{i,j})-1\right) -f_i - g_j \right)
\end{align*}
Note that for all $i, j,k$ and some small $\delta$,
$$ P_k^{i,j}\left(\lambda_k C_k^{i,j} -\varepsilon\left(\log(P_k^{i,j})-1\right) -f_i - g_j \right)<0$$
if $P_k^{i,j}\in(0,\delta)$ and this quantity goes to 0 as $P_k^{i,j}$ goes to 0. Therefore $P_k^{i,j}>0$ and the problem becomes
\begin{align*}
\inf_{P>0}\sup_{\nu \in\mathbb{R}}
   \sum_{k=1}^N \sum_{i,j} P_k^{i,j} \left(\lambda_k C_k^{i,j} +\varepsilon\left(\log(P_k^{i,j})-1\right) -f_i - g_j \right) +\nu\left(\sum_{k=1}^N \sum_{i,j} P_k^{i,j}-1 \right).
\end{align*}
The solution to this problem is for all $k\in\{1,..,N\}$,
\begin{align*}
     P_k = \frac{\exp\left(\frac{f\mathbf{1}_n^T + \mathbf{1}_m g^T-\lambda_k C_k}{\varepsilon}\right)}{\sum_{k=1}^N \sum_{i,j} \exp\left(\frac{f_i + g_j-\lambda_k C_k^{i,j}}{\varepsilon}\right)}
\end{align*}
Therefore we obtain that 
\begin{align*}
    \widehat{\MOT}^{\bm{\varepsilon}}_{\mathbf{C}}(a,b)&=\sup_{\lambda\in\Delta_N^{+}}\sup_{f\in\mathbb{R}^n,~g\in\mathbb{R}^m} \langle f,a \rangle + \langle g, b \rangle\\
    &-  \varepsilon \sum_{k=1}^N \sum_{i,j} P_k^{i,j} \left[\log\left(\sum_{k=1}^N \sum_{i,j} \exp\left(\frac{f_i + g_j-\lambda_k C_k^{i,j}}{\varepsilon}\right)\right) + 1\right]\\
    &= \sup_{\lambda\in\Delta_N^{+}}\sup_{f\in\mathbb{R}^n,~g\in\mathbb{R}^m} \langle f,a \rangle + \langle g, b \rangle -  \varepsilon \left[\log\left(\sum_{k=1}^N \sum_{i,j} \exp\left(\frac{f_i + g_j-\lambda_k C_k^{i,j}}{\varepsilon}\right)\right) + 1\right].
\end{align*}
From now on, we denote for all $\lambda\in\Delta_N^{+}$ 
\begin{align*}
     \widehat{\MOT}^{\bm{\varepsilon},\lambda}_{\mathbf{C}}(a,b)&:= \inf_{P\in \Gamma_{a,b}^N} \sum_{i=1}^N \lambda_i \langle P_i,C_i\rangle -\varepsilon \ent(P_i) \\
      \widehat{\MOT}^{\bm{\varepsilon},\lambda}_{\mathbf{C}}(a,b)&:= \sup_{f\in\mathbb{R}^n,~g\in\mathbb{R}^m} \langle f,a \rangle + \langle g, b \rangle -  \varepsilon \left[\log\left(\sum_{k=1}^N \sum_{i,j} \exp\left(\frac{f_i + g_j-\lambda_k C_k^{i,j}}{\varepsilon}\right)\right) + 1\right] 
\end{align*}
which has just been shown to be dual and equal. Thanks to \citep[Theorem 1]{nesterov2005smooth}, as for all $\lambda\in\mathbb{R}^N$, $P\in\Gamma_{a,b}^N\rightarrow \sum_{i=1}^N \lambda_i \langle P_i,C_i\rangle -\varepsilon \ent(P_i)$ is $\varepsilon$-strongly convex, then for all $\lambda\in\mathbb{R}^N$, $(f,g)\rightarrow \nabla_{(f,g)} F(\lambda,f,g)$ is $\frac{\Vert A\Vert_{1\rightarrow 2}^2}{\varepsilon}$ Lipschitz-continuous where $A$ is the linear operator of the equality constraints of the primal problem. Moreover this norm is equal to the maximum Euclidean norm of a column of A. By definition, each
column of A contains only $2N$ non-zero elements, which
are equal to one. Hence, $\Vert A\Vert_{1\rightarrow 2} = \sqrt{2N}$. Let us now show that for all $(f,g)\in\mathbb{R}^n\times\mathbb{R}^m $ $\lambda\in\mathbb{R}^N \rightarrow \nabla_{\lambda} F(\lambda,f,g)$ is also Lipschitz-continuous. Indeed we remarks that 
\begin{align*}
    \frac{\partial^2 F}{\partial\lambda_q\partial\lambda_k} = \frac{1}{\varepsilon\nu^2}\left[\sigma_{q,1}(\lambda)\sigma_{k,1}(\lambda) - \nu (\sigma_{k,2}(\lambda)1\!\!1_{k=q})\right]
\end{align*}
where $1\!\!1_{k=q}=1$ iff $k=q$ and 0 otherwise, for all $k\in\{1,...,N\}$ and $p\geq 1$
\begin{align*}
    \sigma_{k,p}(\lambda) &= \sum_{i,j} (C_k^{i,j})^p \exp\left(\frac{f_i + g_j-\lambda_k C_k^{i,j}}{\varepsilon}\right) \\
    \nu &= \sum_{k=1}^N \sum_{i,j} \exp\left(\frac{f_i + g_j-\lambda_k C_k^{i,j}}{\varepsilon}\right).\\
\end{align*}
Let $v\in\mathbb{R}^N$, and by denoting $\nabla^2_{\lambda}F$ the Hessian of $F$ with respect to $\lambda$ for fixed $f,g$ we obtain first that 
\begin{align*}
  v^T \nabla^2_{\lambda}F v &=\frac{1}{\varepsilon\nu^2} \left[ \left(\sum_{k=1}^N v_k \sigma_{q,1}(\lambda)\right)^2 -\nu \sum_{k=1}^N v_k^2 \sigma_{k,2}\right]\\
  &\leq \frac{1}{\varepsilon\nu^2}\left(\sum_{k=1}^N v_k \sigma_{q,1}(\lambda)\right)^2 \\
  &-\frac{1}{\varepsilon\nu^2} \left(\sum_{k=1}^N |v_k| \sqrt{\sum_{i,j} \exp\left(\frac{f_i + g_j-\lambda_k C_k^{i,j}}{\varepsilon}\right)}  \sqrt{\sum_{i,j} (C_k^{i,j})^2 \exp\left(\frac{f_i + g_j-\lambda_k C_k^{i,j}}{\varepsilon}\right)} \right)^2 \\
  &\leq \frac{1}{\varepsilon\nu^2}\left[\left(\sum_{k=1}^N v_k \sigma_{q,1}(\lambda)\right)^2- \left(\sum_{k=1}^N |v_k| \sum_{i,j}  |C_k^{i,j}| \exp\left(\frac{f_i + g_j-\lambda_k C_k^{i,j}}{\varepsilon}\right)\right)^2\right]\\
  &\leq 0
 \end{align*}
Indeed the last two inequalities come from Cauchy Schwartz. Moreover we have
\begin{align*}
 \frac{1}{\varepsilon\nu^2} \left[ \left(\sum_{k=1}^N v_k \sigma_{q,1}(\lambda)\right)^2 -\nu \sum_{k=1}^N v_k^2 \sigma_{k,2}\right] & = v^T \nabla^2_{\lambda}F v \leq 0   \\
 - \frac{\sum_{k=1}^N v_k^2 \sigma_{k,2}}{\varepsilon\nu}  & \leq \\
  - \frac{\sum_{k=1}^N v_k^2 \max\limits_{1\leq i\leq N}(\Vert C_i\Vert_{\infty}^2) }{\varepsilon}  & \leq 
 \end{align*}
 Therefore we deduce that $\lambda\in\mathbb{R}^N \rightarrow \nabla_{\lambda} F(\lambda,f,g)$ is $\frac{\max\limits_{1\leq i\leq N}(\Vert C_i\Vert_{\infty}^2)}{\varepsilon}$ Lipschitz-continuous, hence $\nabla F(\lambda,f,g)$ is $\frac{\max\left(\max\limits_{1\leq i\leq N}\Vert C_i\Vert_{\infty}^2,2N\right)}{\varepsilon}$ Lipschitz-continuous on $\mathbb{R}^N\times \mathbb{R}^n \times\mathbb{R}^m$. 
\end{prv*}

Denote $L:= \frac{\max\left(\max\limits_{1\leq i\leq N}\Vert C_i\Vert_{\infty}^2,2N\right)}{\varepsilon}$ the  Lipschitz constant of $F_{\mathbf{C}}^{\varepsilon}$. Moreover for all $\lambda\in\mathbb{R}^N$, let $\text{Proj}_{\Delta_N^{+}}(\lambda)$ the unique solution of the following optimization problem
\begin{align}
\label{prob:proj}
    \min_{x\in\Delta_N^{+}} \Vert x - \lambda\Vert_2^2.
\end{align}
Let us now introduce the following algorithm.




\begin{algorithm}[H]
\SetAlgoLined
\textbf{Input:} $\mathbf{C}=(C_i)_{1\leq i\leq N}$, $a$, $b$, $\varepsilon$, $L$\\
\textbf{Init:} $f^{-1}=f^0 \leftarrow \mathbf{0}_n\text{;  }$ $g^{-1} = g^0 \leftarrow \mathbf{0}_m\text{;  }$ $\lambda^{-1} = \lambda^0 \leftarrow (1/N,...,1/N)\in\mathbb{R}^N$\\
\For{$k=1,2,...$}{
$(v,w,z)^T \leftarrow (\lambda^{k-1},f^{k-1},g^{k-1})^T+\frac{k-2}{k+1}\left((\lambda^{k-1},f^{k-1},g^{k-1})^T- (\lambda^{k-2},f^{k-2},g^{k-2})^T\right);\\
\lambda^k \leftarrow \text{Proj}_{\Delta_N^{+}}\left( v + \frac{1}{L}\nabla_{\lambda} F_{\mathbf{C}}^{\varepsilon}(v,w,z)\right);
\\ (g^k, f^k)^T \leftarrow (w,z)^T + \frac{1}{L}\nabla_{(f,g)} F_{\mathbf{C}}^{\varepsilon}(v,w,z).$}
\caption{Accelerated Projected Gradient Ascent Algorithm\label{algo:Proj-grad}}
\textbf{Result}: $\lambda,f,g$
\end{algorithm}

\cite{beck2009fast,tseng2008accelerated} give us that the accelerated projected gradient ascent algorithm achieves the optimal rate for first order methods of $\mathcal{O}(1/k^2)$ for smooth functions. To perform the projection we use the algorithm proposed in~\cite{shalev2006efficient} which finds the solution of (\ref{prob:proj}) after $\mathcal{O}(N\log(N))$ algebraic operations \citep{wang2013projection}.

\subsection{Fair cutting cake problem}

Let $\mathcal{X}$, be a set representing a cake. The aim of the cutting cake problem is to divide it in $\mathcal{X}_1,\dots,\mathcal{X}_N$ disjoint sets among the $N$ individuals. The utility for a single individual $i$ for a slice $S$ is denoted $V_i(S)$. It is often assumed that $V_i(\mathcal{X}) = 1$ and that $V_i$ is additive for disjoint sets. There exists many criteria to assess fairness for a partition $\mathcal{X}_1,\dots,\mathcal{X}_N$ such as proportionality ($V_i(\mathcal{X}_i)\geq 1/N$), envy-freeness ($V_i(\mathcal{X}_i)\geq V_i(\mathcal{X}_j)$) or equitability ($V_i(\mathcal{X}_i)= V_j(\mathcal{X}_j)$). A possible problem to solve equitability and proportionality in the cutting cake problem is the following:
\begin{align}
    \inf_{\substack{\mathcal{X}_1,...,\mathcal{X}_N\\  \sqcup_{i=1}^N\mathcal{X}_i=\mathcal{X}}}\max_i V_i(\mathcal{X}_i)
\end{align}

Note that here we do not want to solve the problem under equality constraints since the problem might not be well defined. Moreover the existence of the optimum is not immediate. A natural relaxation of this problem is when there is a divisible quantity of each element of the cake ($x\in\mathcal{X}$). In that case, the cake is no more a set but rather a distribution on this set $\mu$. Following the primal formulation of $\MOT$, it is clear that it is a relaxation of the cutting cake problem where the goal is to divide the cake viewed as a distribution. For the cutting cake problem with two cakes $\mathcal{X}$ and $\mathcal{Y}$, the problem can be cast as follows:
\begin{align}
    \inf_{\substack{\mathcal{X}_1,...,\mathcal{X}_N~\text{s.t.}~\sqcup_{i=1}^N\mathcal{X}_i=\mathcal{X}\\\mathcal{Y}_1,...,\mathcal{Y}_N~\text{s.t.}~\sqcup_{i=1}^N\mathcal{Y}_i=\mathcal{Y}}}\max_i  V_i(\mathcal{X}_i,\mathcal{Y}_i)
\end{align}
Here $\MOT$ is the relaxation of this problem where we split the cakes viewed as distributions instead of sets themselves. Note that in this problem, the utility of the agents are coupled.
\newpage

\section{Illustrations and Experiments}
\label{appendix-illustrations}

\subsection{Primal Formulation}
Here we show the couplings obtained when we consider three negative costs $\tilde{c}_i$ which corresponds to the situation where we aim to obtain a fair division of goods between three agents. Moreover we show the couplings obtained according to the transport viewpoint where we consider the opposite of these three negative cost functions, i.e. $c_i:=-\tilde{c}_i$. We can see that the couplings obtained in the two situations are completely different, which is expected. Indeed in the fair division problem, we aim at finding couplings which maximize the total utility of each agent ($\int c_id\gamma_i^{1}$) while ensuring that their are equal while in the other case, we aim at finding couplings which minimize the total transportation cost of each agent ($\int c_id\gamma_i^{2}$) while ensuring that their are equal. Obviously we always have 
that  $$\forall i~~\int c_id\gamma_i^{2}\leq \int c_id\gamma_i^{1}.$$

\begin{figure*}[h!]
\begin{tabular}{@{}c@{}c@{}c@{}c@{}}
\includegraphics[width=0.25\textwidth]{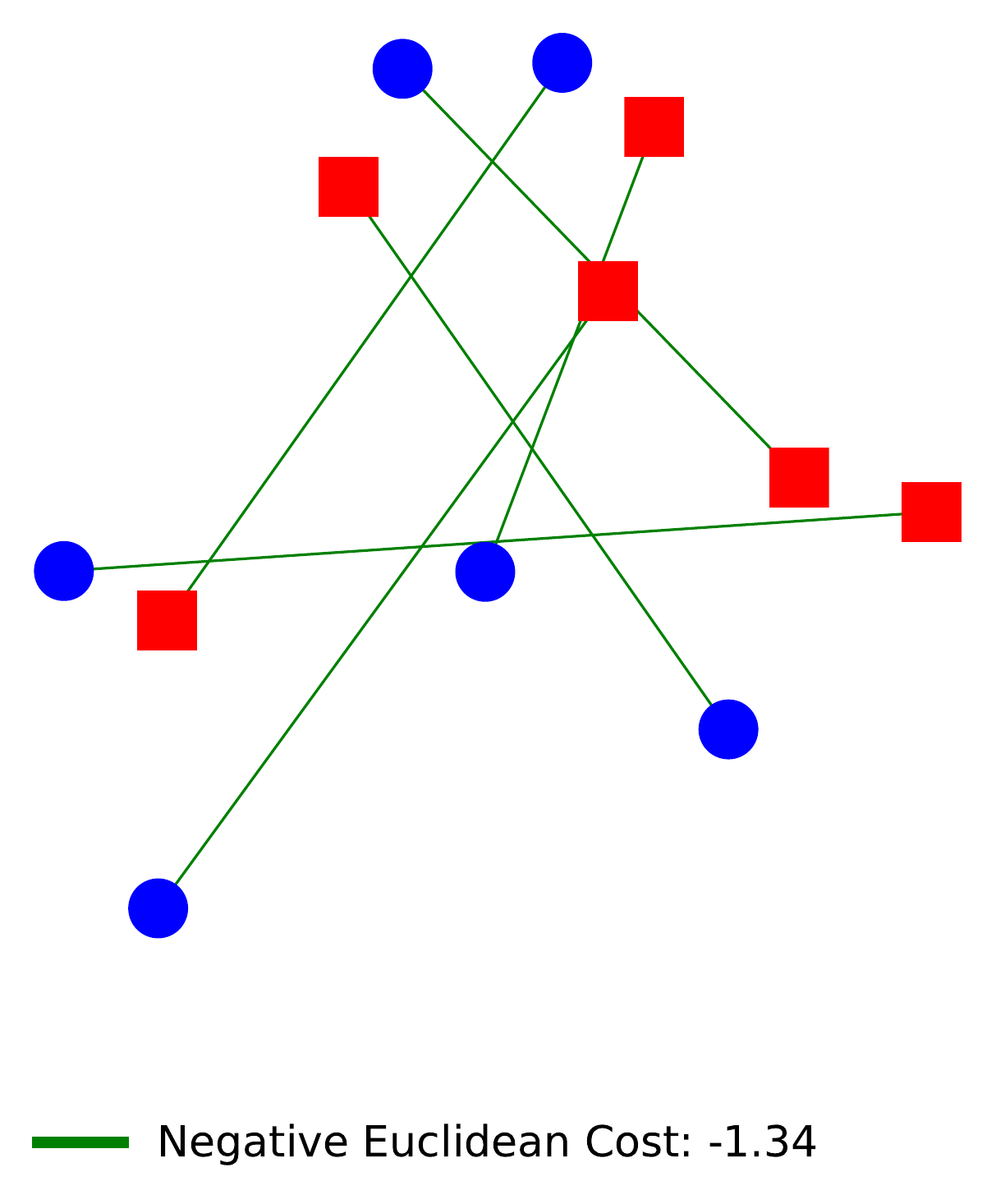}&
\includegraphics[width=0.25\textwidth]{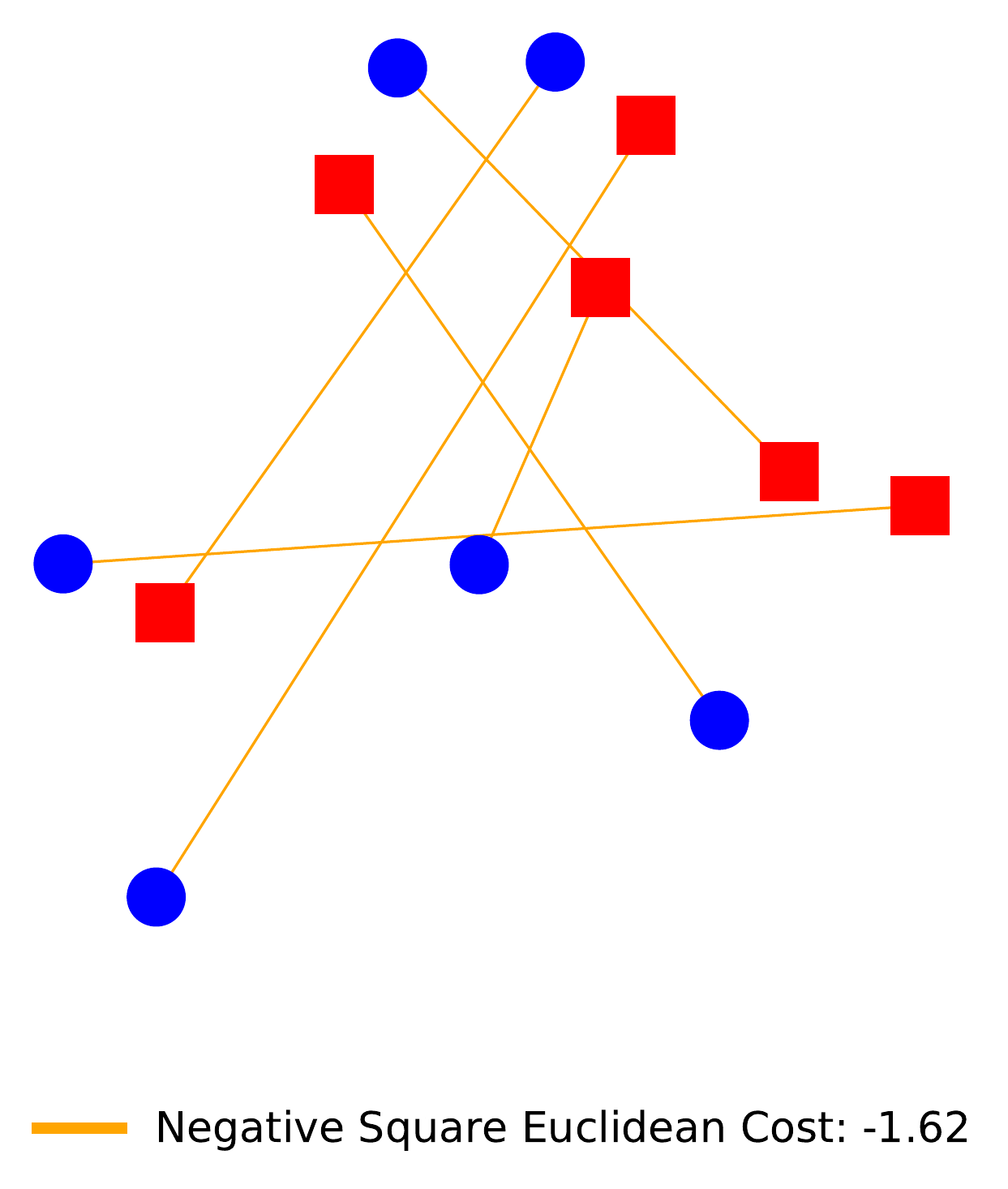}&
\includegraphics[width=0.25\textwidth]{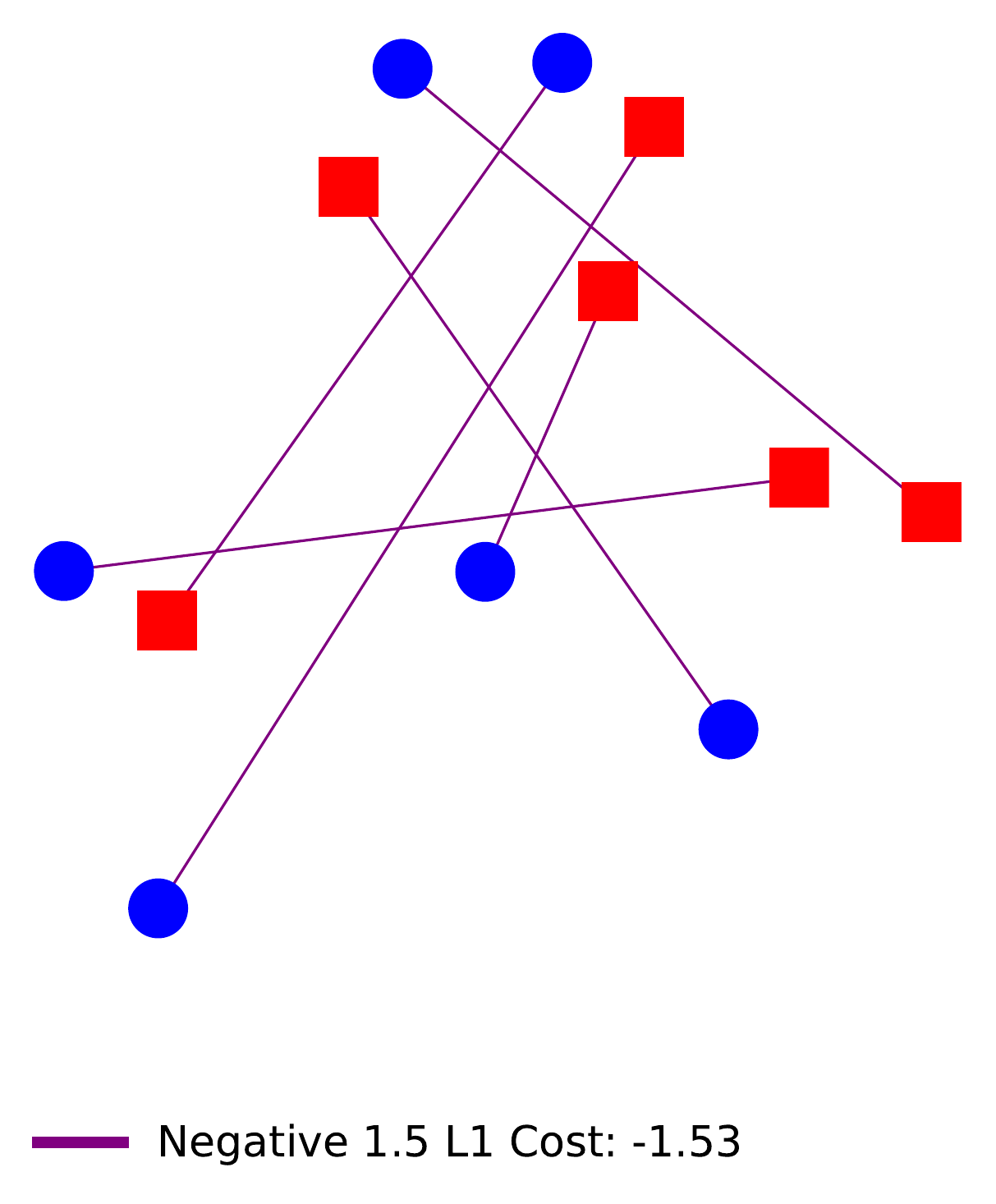}&
\includegraphics[width=0.25\textwidth]{figures/primal_W_1_2_3_neg_norm.pdf}
\end{tabular}
\caption{Comparison of the optimal couplings obtained from standard OT for three different costs and $\MOT$ in case of negative costs (i.e. utilities). Blue dots and red squares represent the locations of two discrete uniform measures. \emph{Left, middle left, middle right}: Kantorovich couplings between the two measures for negative Euclidean cost ($-\Vert\cdot\Vert_2$), negative square Euclidean cost ($-\Vert\cdot\Vert_2^{2}$) and negative 1.5 L1 norm ($-\Vert\cdot\Vert_1^{1.5}$) respectively. \emph{Right}: Equitable and optimal division of the resources between the $N=3$ different negative costs (i.e. utilities) given by $\MOT$.  Note that the partition between the agents is equitable (i.e. utilities are equal) and proportional (i.e. utilities are larger than $1/N$.}
\label{fig-primal-fair-appendix}
\end{figure*}

\begin{figure*}[h!]
\begin{tabular}{@{}c@{}c@{}c@{}c@{}}
\includegraphics[width=0.25\textwidth]{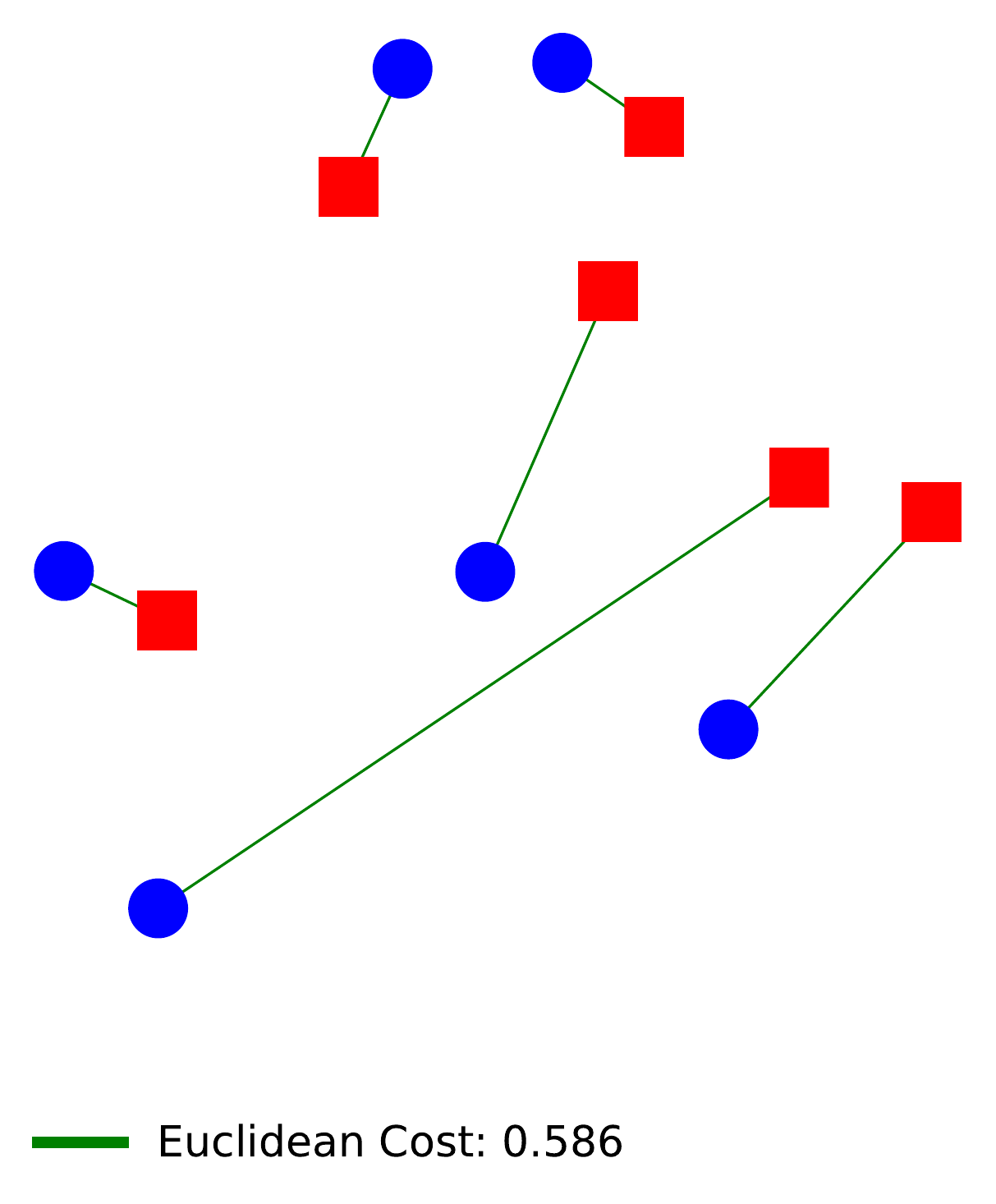}&
\includegraphics[width=0.25\textwidth]{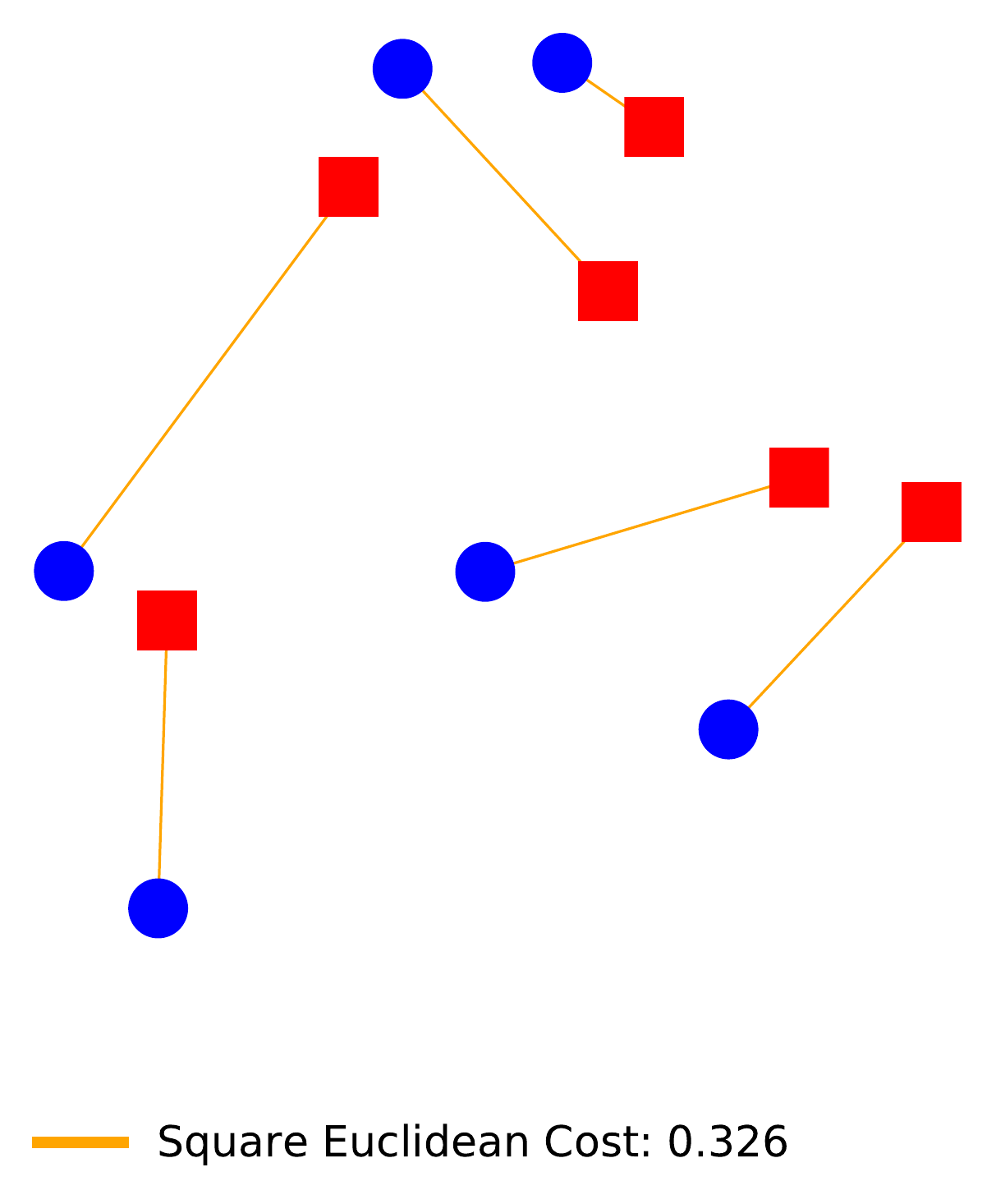}&
\includegraphics[width=0.25\textwidth]{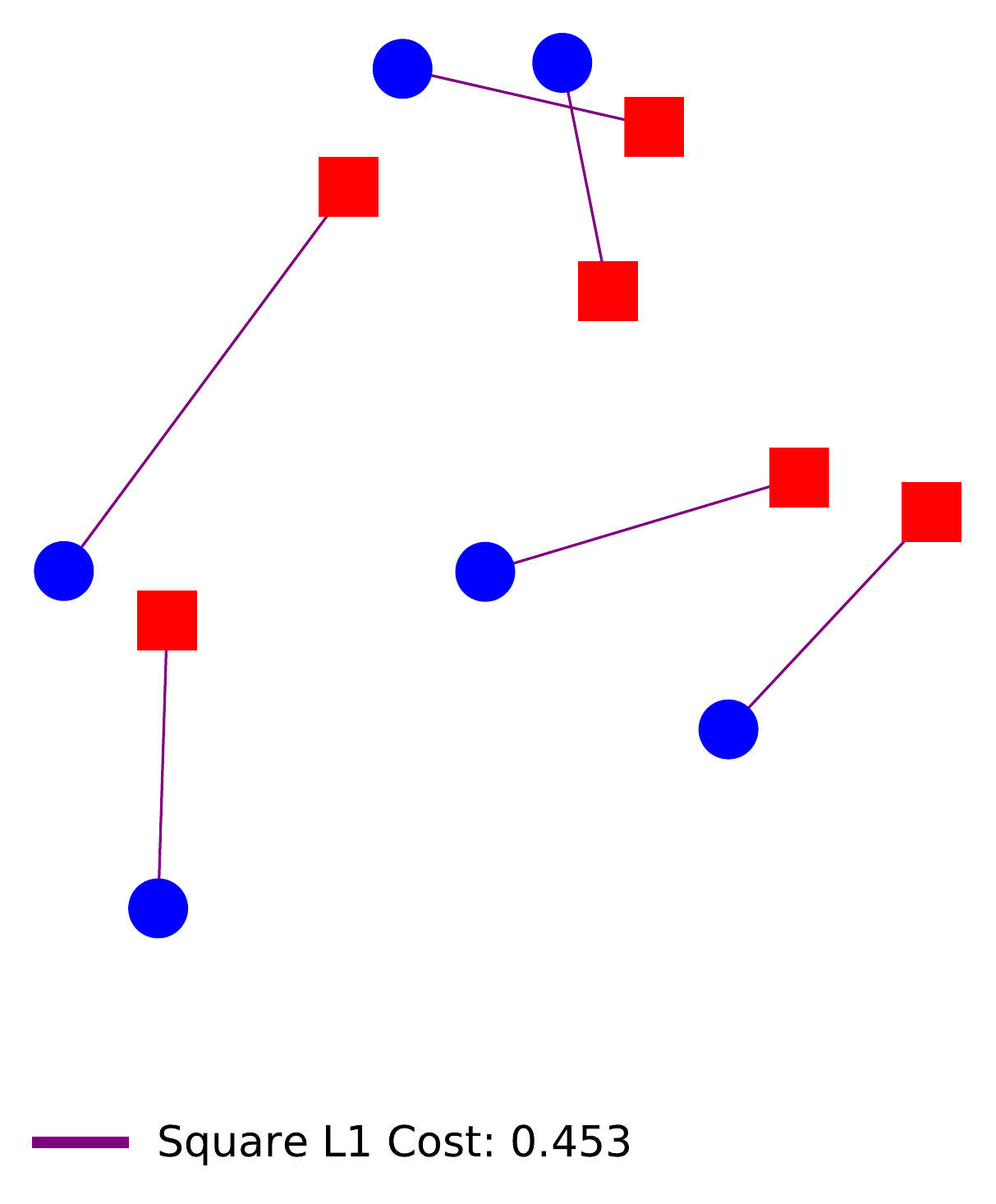}&
\includegraphics[width=0.25\textwidth]{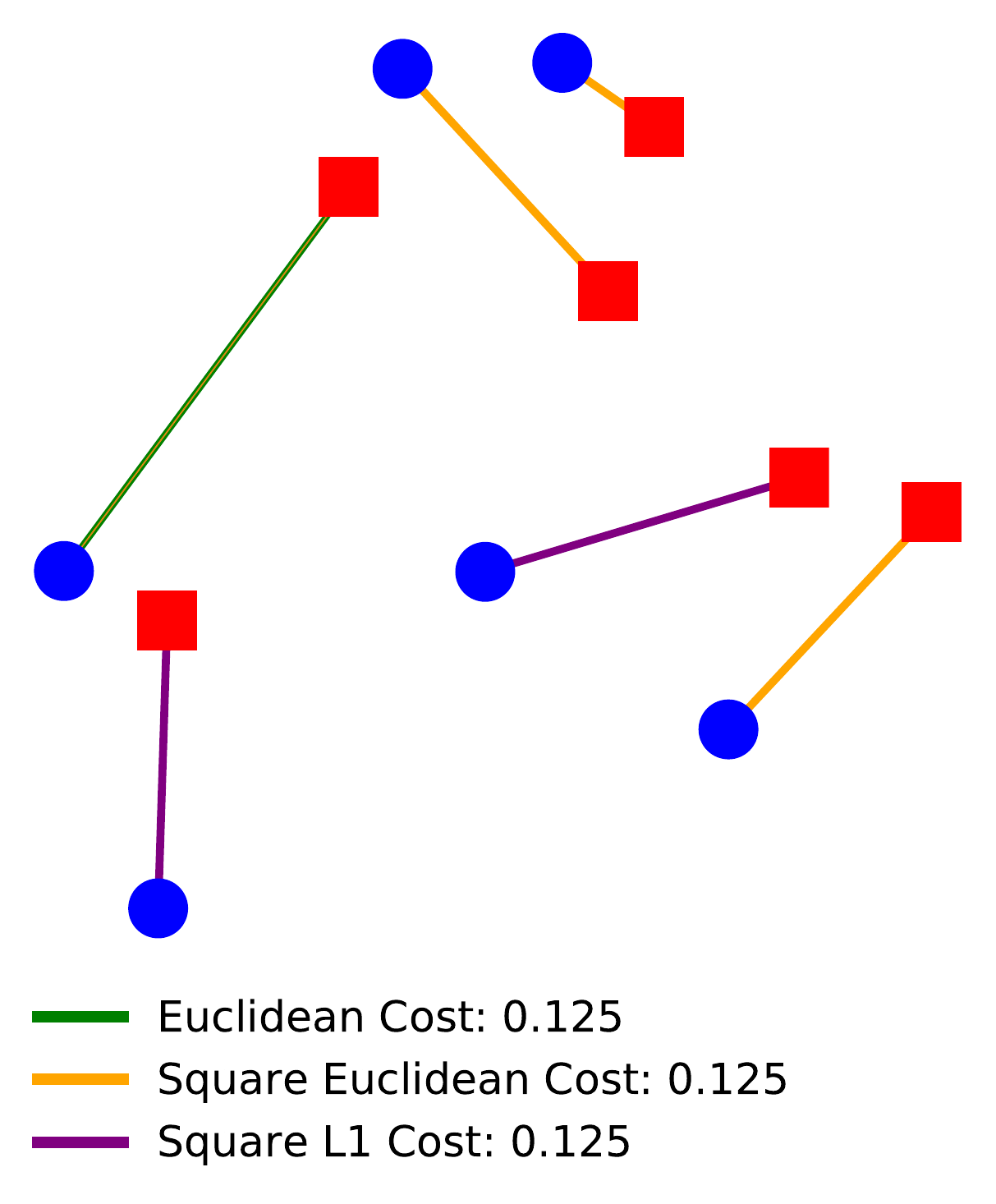}
\end{tabular}
\caption{Comparison of the optimal couplings obtained from standard OT for three different costs and $\MOT$ in case of postive costs. Blue dots and red squares represent the locations of two discrete uniform measures. \emph{Left, middle left, middle right}: Kantorovich couplings between the two measures for Euclidean cost  ($\Vert\cdot\Vert_2$), square Euclidean cost ($\Vert\cdot\Vert_2^{2}$) and 1.5 L1 norm ($\Vert\cdot\Vert_1^{1.5}$) respectively. \emph{Right}: transport couplings of $\MOT$ solving Eq.~\eqref{eq-primal}. Note that each cost contributes equally and its contribution is lower than the smallest OT cost.}
\label{fig:transport-map-ot-view}
\end{figure*}

\newpage
\subsection{Dual Formulation}
\begin{figure*}[h!]
\begin{tabular}{@{}c@{}c@{}c@{}c@{}}
\includegraphics[width=0.23\textwidth]{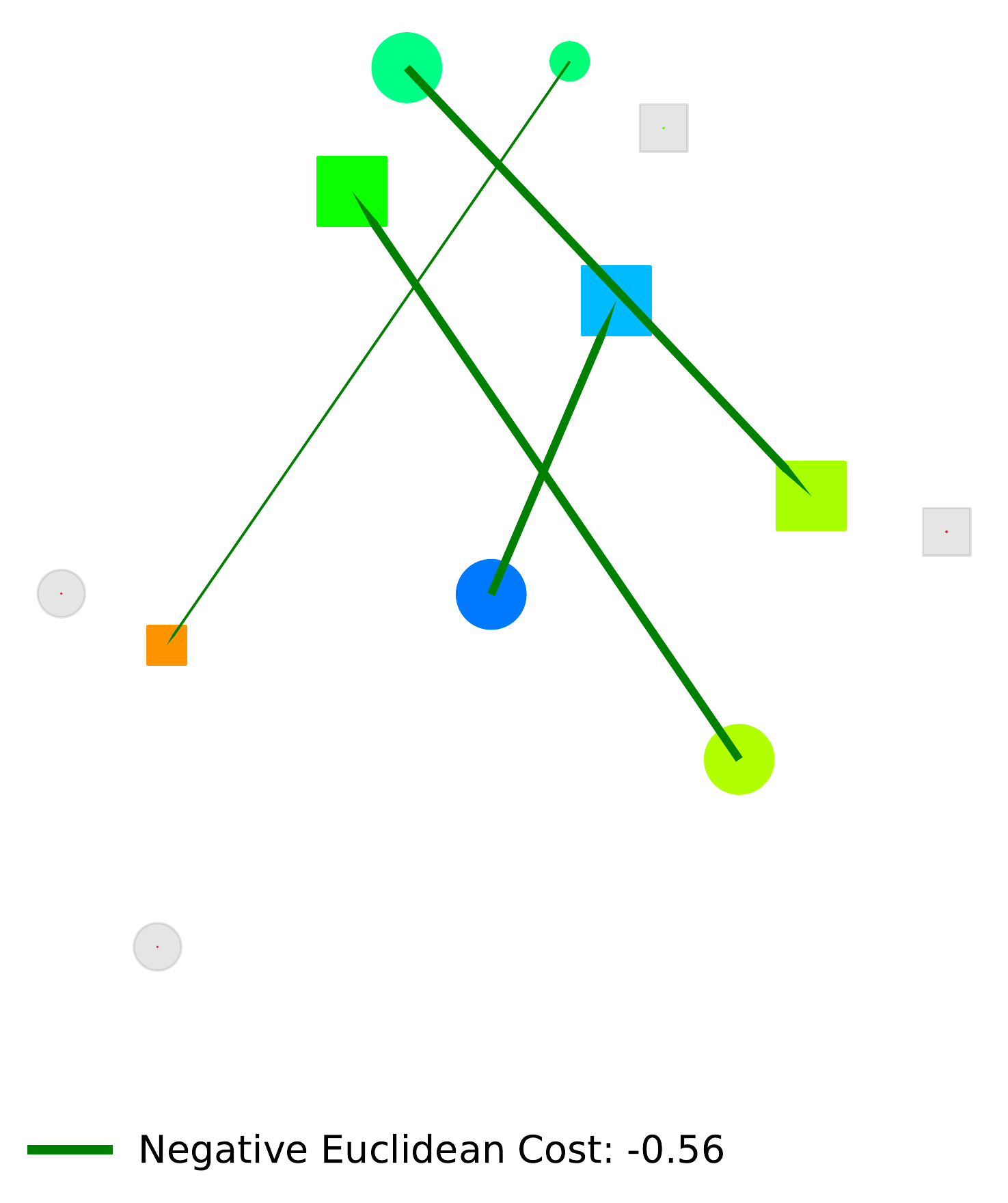}&
\includegraphics[width=0.23\textwidth]{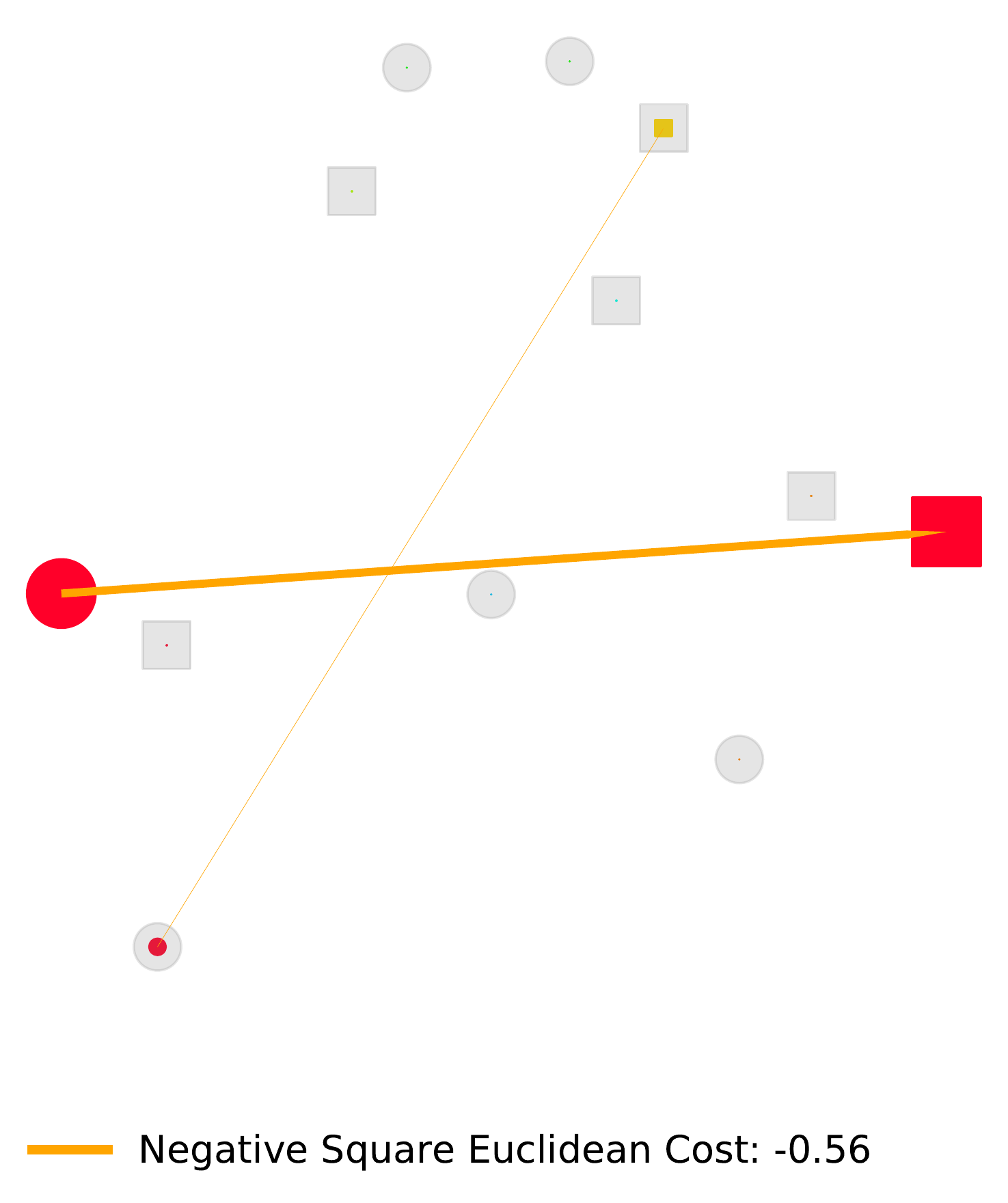}&
\includegraphics[width=0.23\textwidth]{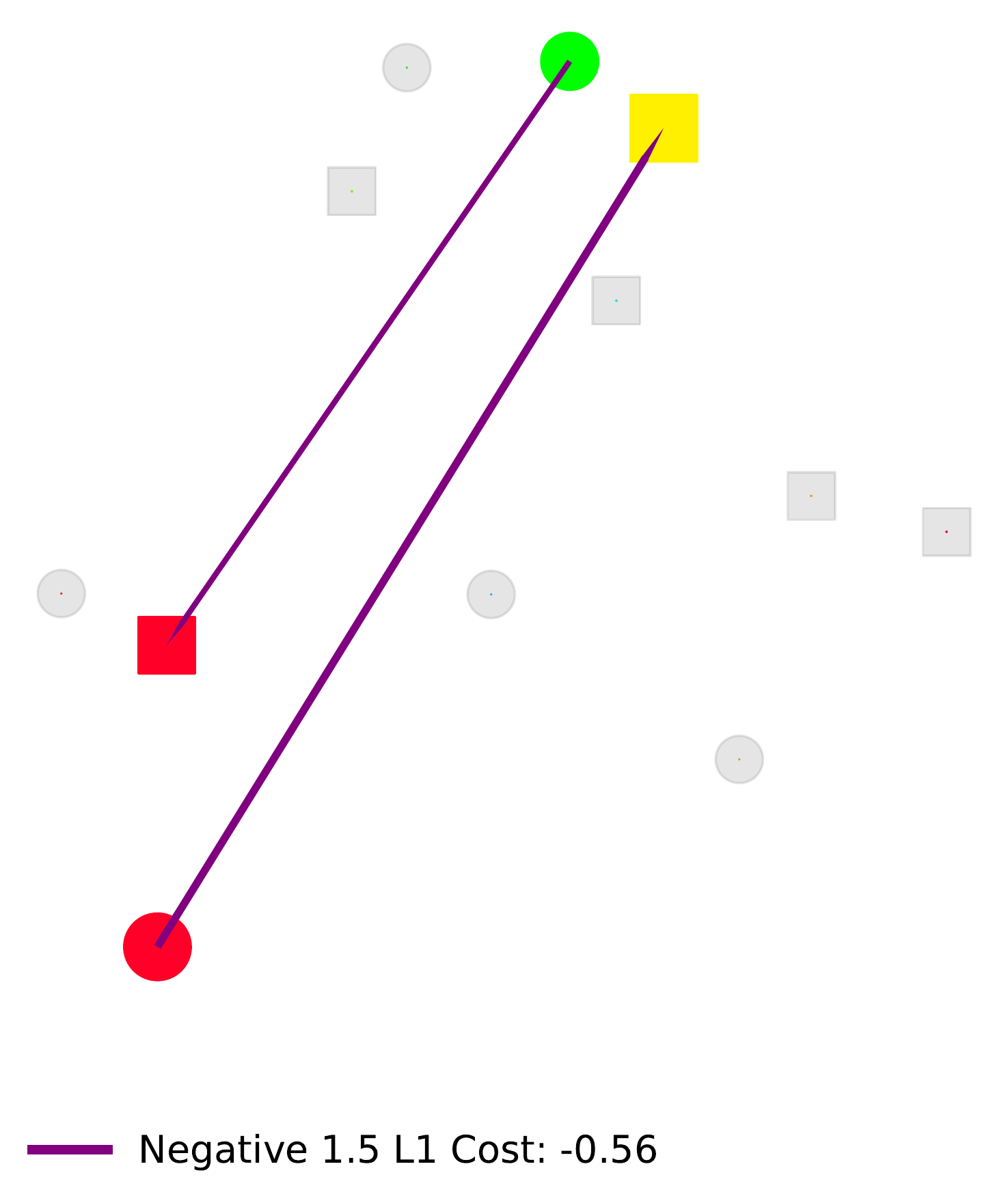}&
\includegraphics[width=0.268\textwidth]{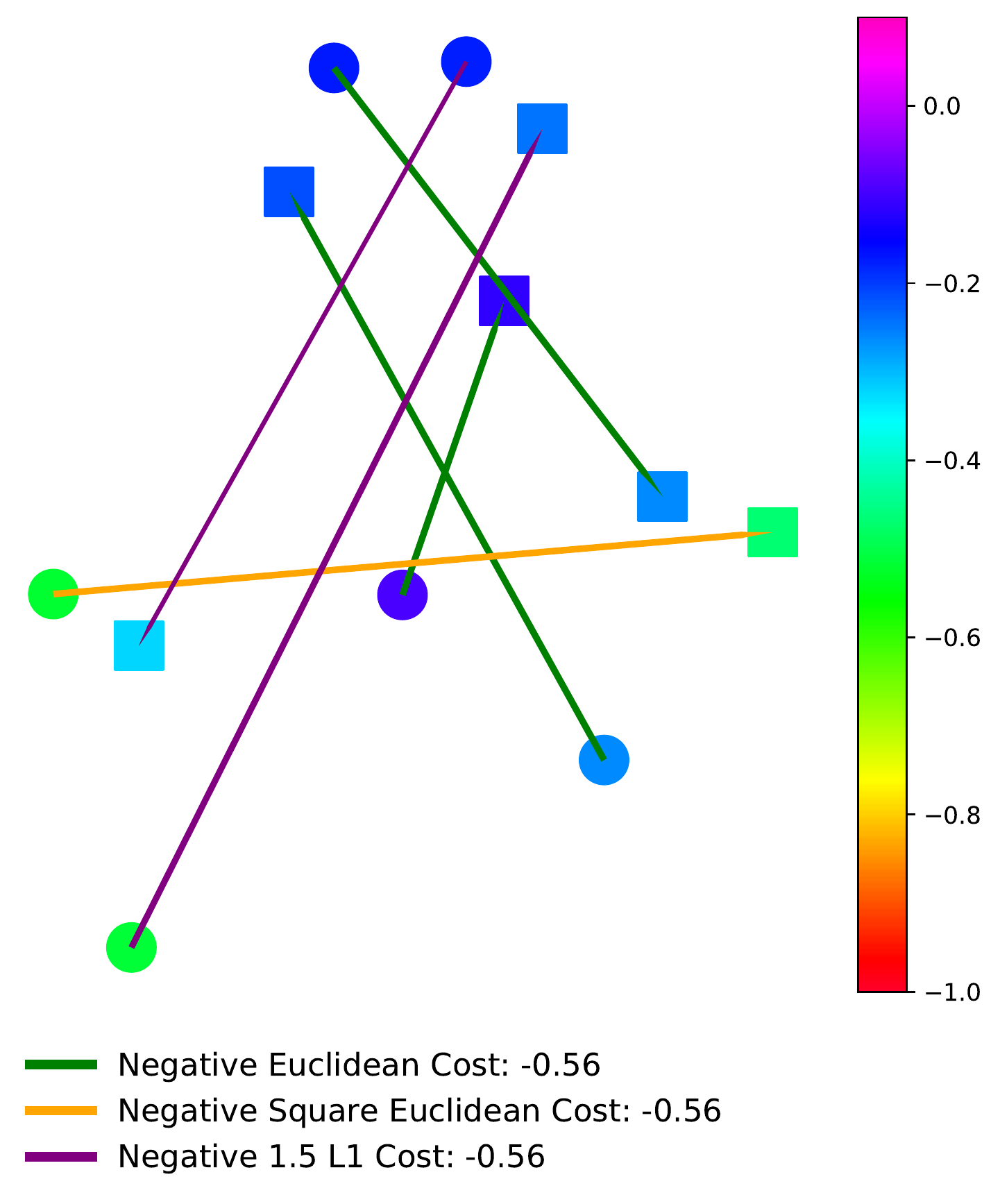}
\end{tabular}
\caption{\emph{Left, middle left, middle right}: the size of dots and squares is proportional to the weight of their representing atom in the distributions $\mu_k^{*}$ and $\nu_k^{*}$ respectively. 
The utilities $f_k^{*}$ and $g_k^{*}$ for each point in respectively $\mu_k^{*}$ and $\nu_k^{*}$ are represented by the color of dots and squares according to the color scale on the right hand side. The gray dots and squares correspond to the points that are ignored by agent $k$ in the sense that there is no mass or almost no mass in distributions $\mu^*_k$ or $\nu^*_k$. \emph{Right}: the size of dots and squares are uniform since they correspond to the weights of uniform distributions $\mu$ and $\nu$ respectively. The values of $f^*$ and $g^*$ are given also by the color at each point. Note that each agent gets exactly the same total utility, corresponding exactly to $\MOT$. This value can be computed using dual formulation~\eqref{eq-dual} and for each figure it equals the sum of the values (encoded with colors) multiplied by the weight of each point (encoded with sizes).\label{fig-dual-appendix-fair}}
\end{figure*}

\begin{figure*}[h!]
\begin{tabular}{@{}c@{}c@{}c@{}c@{}}
\includegraphics[width=0.23\textwidth]{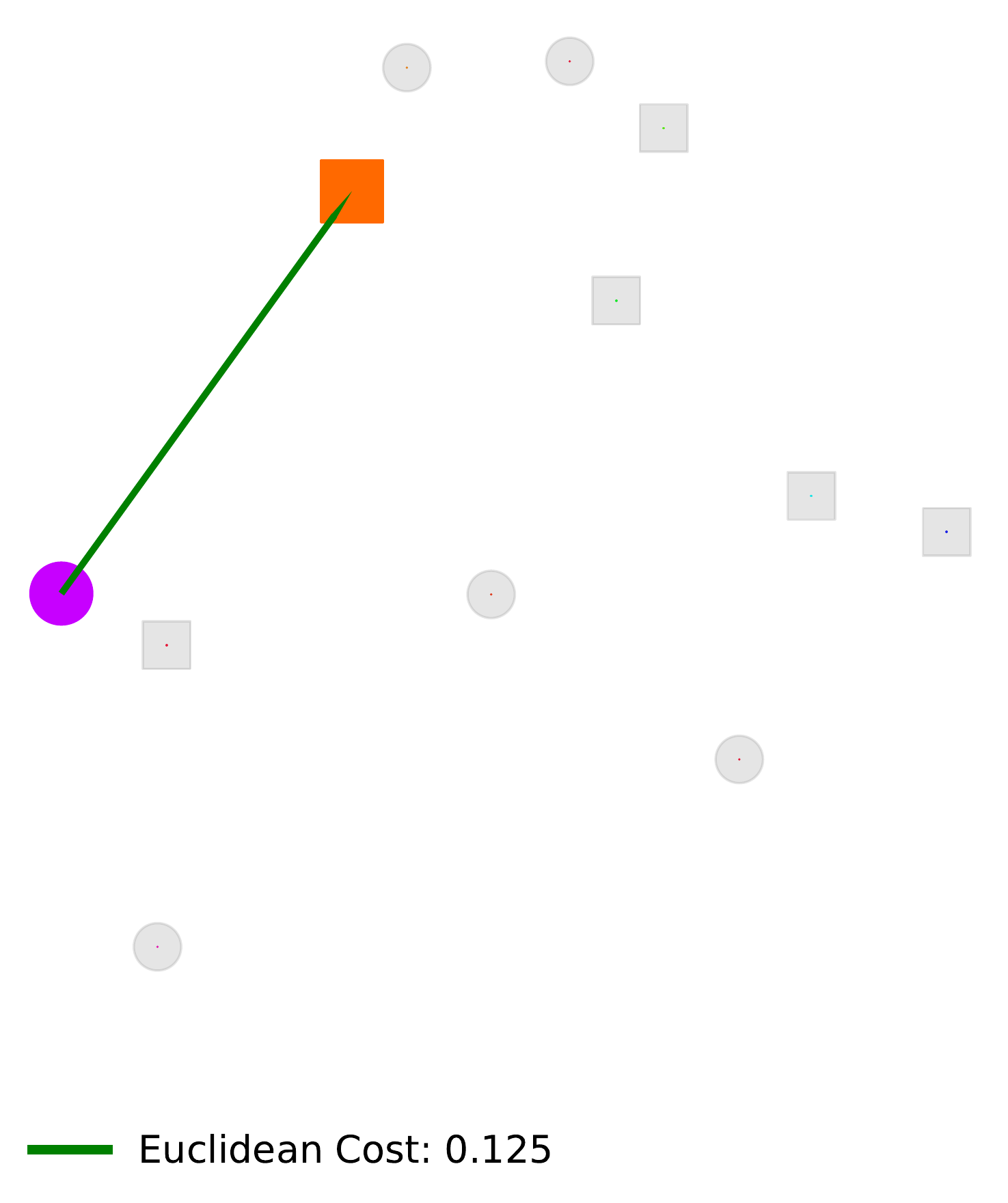}&
\includegraphics[width=0.23\textwidth]{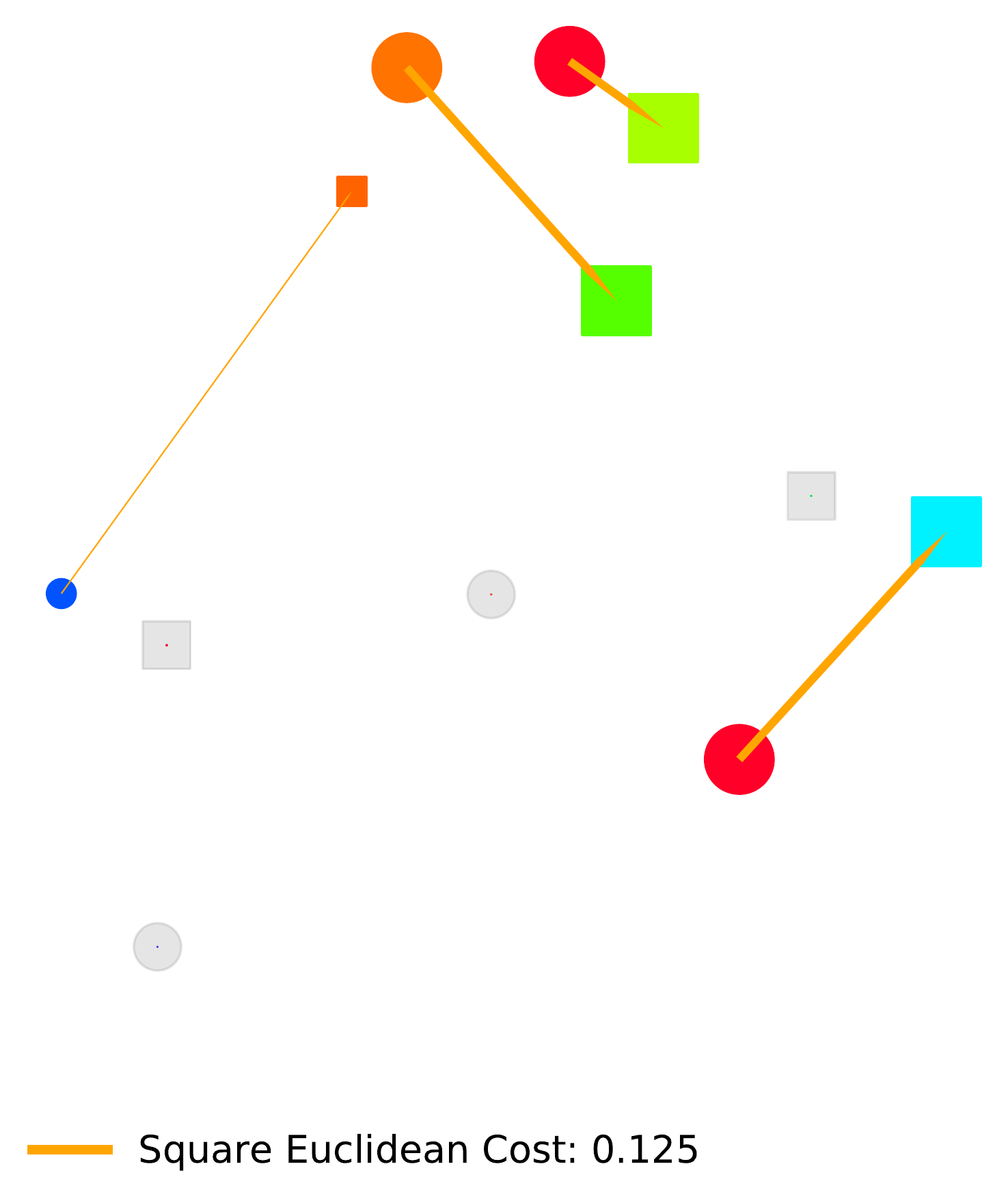}&
\includegraphics[width=0.23\textwidth]{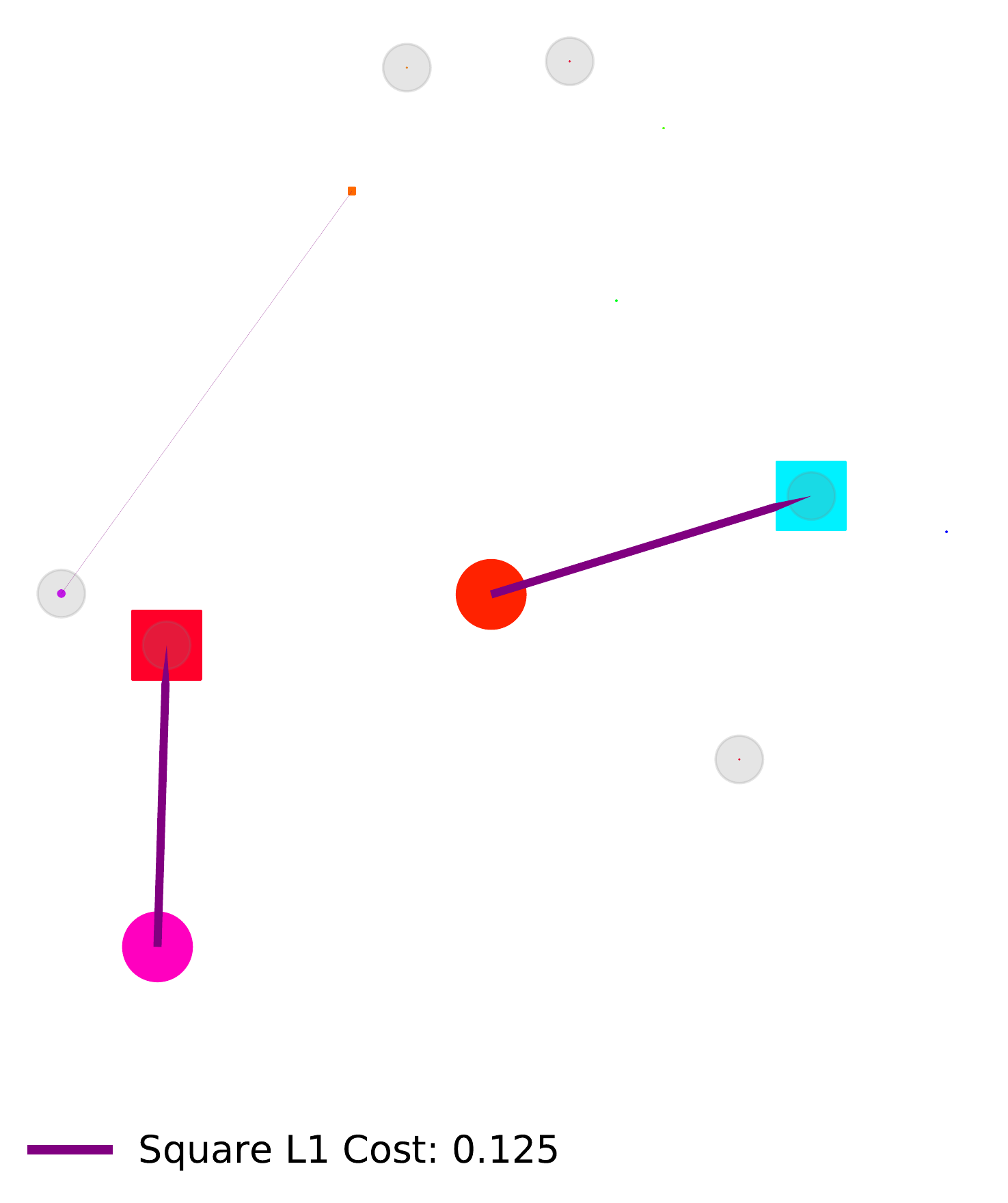}&
\includegraphics[width=0.268\textwidth]{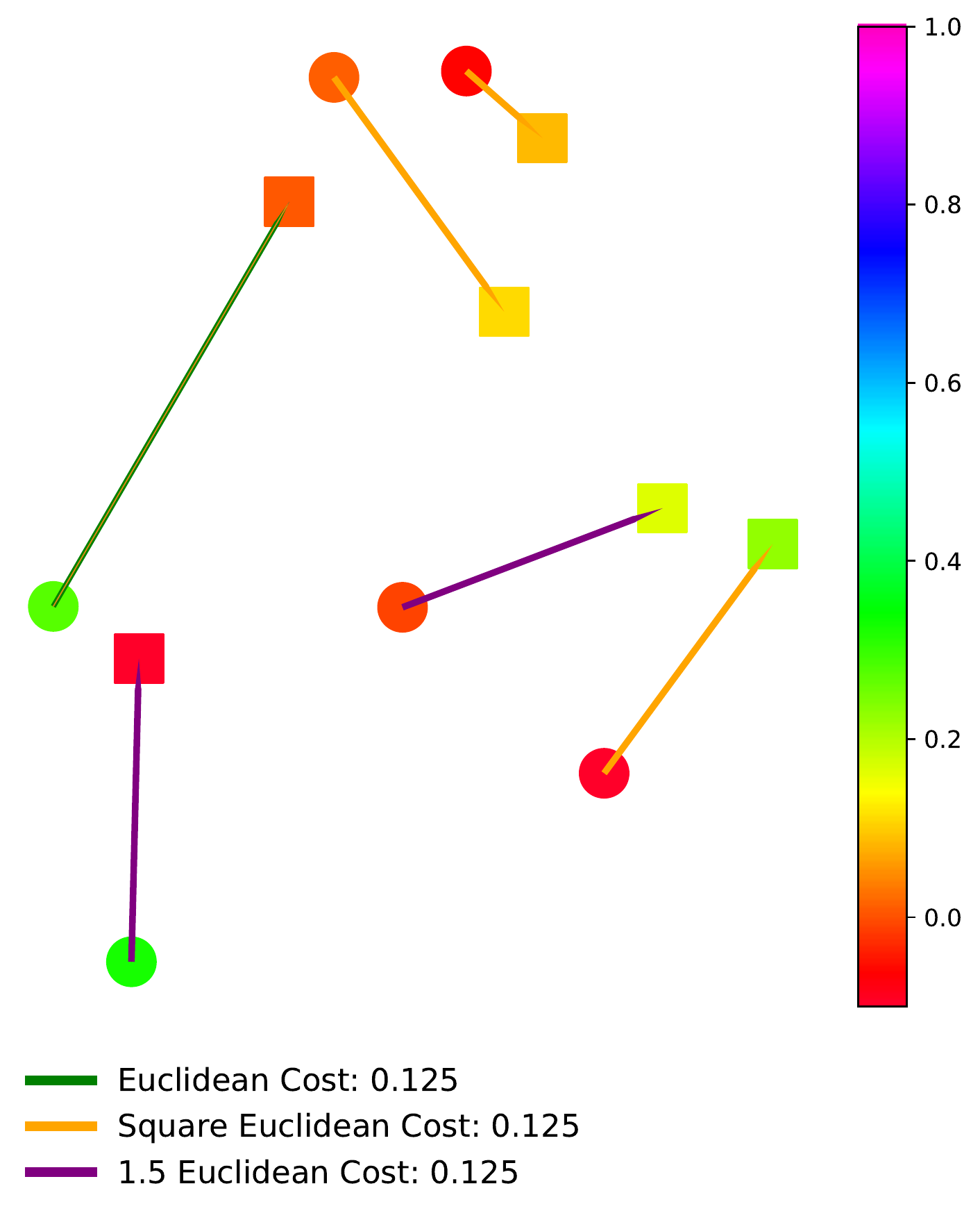}
\end{tabular}
\caption{\emph{Left, middle left, middle right}: the size of dots and squares is proportional to the weight of their representing atom in the distributions $\mu_k^{*}$ and $\nu_k^{*}$ respectively. 
The collection ``cost'' $f_k^{*}$ for each point in $\mu_k^{*}$, and its delivery counterpart $g_k^{*}$ in $\nu_k^{*}$ are represented by the color of dots and squares according to the color scale on the right hand side. The gray dots and squares correspond to the points that are ignored by agent $k$ in the sense that there is no mass or almost no mass in distributions $\mu^*_k$ or $\nu^*_k$. \emph{Right}: the size of dots and squares are uniform since they corresponds to the weights of uniform distributions $\mu$ and $\nu$ respectively. The values of $f^*$ and $g^*$ are given also by the color at each point. Note that each agent earns exactly the same amount of money, corresponding exactly \MOT cost. This value can be computed using dual formulation~\eqref{eq-dual} or its reformulation~\eqref{eq:dual_interp} and for each figure it equals the sum of the values (encoded with colors) multiplied by the weight of each point (encoded with sizes).}
\label{fig:potential-dual-ot-viewpoint}
\end{figure*}

Here we show the dual variables obtained in the exact same settings as in the primal illustrations. Figure~\ref{fig-dual-appendix-fair} shows the dual associated to the primal problem exposed in Figure~\ref{fig-primal-fair-appendix} and  Figure~\ref{fig:potential-dual-ot-viewpoint} shows the dual associated to the primal problem exposed in Figure~\ref{fig:transport-map-ot-view}.

\paragraph{Transport viewpoint of the Dual Formulation.} Assume that the $N$ agents are not able to solve the primal problem~(\ref{eq-primal}) which aims at finding the cheapest equitable partition of the work among the $N$ agents for transporting the distributions of goods $\mu$ to the distributions of stores $\nu$. Moreover assume that there is an external agent who can do the transportation work for them with the following pricing scheme: he or she splits the logistic task into that of collecting and then delivering the goods, and will apply a collection price $\tilde{f}(x)$ for one unit of good located at $x$ (no matter where that unit is sent to), and a delivery price $\tilde{g}(y)$ for one unit to the location $y$ (no matter from which place that unit comes from). Then the external agent for transporting some goods $\mu$ to some stores $\nu$ will charge $\int_{x\in\mathcal{X}} \tilde{f}(x)d\mu(x)+ \int_{y\in\mathcal{Y}} \tilde{g}(y)d\nu(y)$. However he or she has the constraint that the pricing must be equitable among the agents and therefore wants to ensure that each agent will pay exactly $\frac{1}{N}\int_{x\in\mathcal{X}} \tilde{f}(x)d\mu(x)+ \int_{y\in\mathcal{Y}} \tilde{g}(y)d\nu(y)$. Denote $f=\frac{\tilde{f}}{N}$, $g=\frac{\tilde{g}}{N}$ and therefore the price paid by each agent becomes  $\int_{x\in\mathcal{X}} f(x)d\mu(x)+ \int_{y\in\mathcal{Y}} g(y)d\nu(y)$. Moreover, to ensure that each agent will not pay more than he would if he was doing the job himself or herself, he or she must guarantee that for all $\lambda\in\Delta_N^{+}$, the pricing scheme ($f$,$g$) satisfies: 
$$f\oplus g\leq \min(\lambda_i c_i).$$
Indeed under this constraint, it is easy for the agents to check that they will never pay more than what they would pay if they were doing the transportation task as we have
$$\int_{x\in\mathcal{X}} f(x)d\mu(x)+ \int_{y\in\mathcal{Y}} g(y)d\nu(y)\leq \int_{\mathcal{X}\times\mathcal{Y}} \min_i(\lambda_i c_i) d\gamma $$ 
which holds for every $\gamma$ in particular for $\gamma^{*}=\sum_{i=1}^N\gamma_i^{*}$ optimal solution of the primal problem~(\ref{eq-primal}) from which follows
\begin{align*}
    \int_{x\in\mathcal{X}} f(x)d\mu(x)+ \int_{y\in\mathcal{Y}} g(y)d\nu(y)&\leq\sum_{i=1}^N \int_{\mathcal{X}\times\mathcal{Y}} \min_i(\lambda_i c_i) d\gamma_i^{*}\\
    &\leq \sum_{i=1}^N \lambda_i \int_{\mathcal{X}\times\mathcal{Y}} c_i d\gamma_i^{*}\\
    &=\MOT_{\mathbf{c}}(\mu,\nu)
\end{align*}
Therefore the external agent aims to maximise his or her selling price under the above constraints which is exactly the dual formulation of our problem.

Another interpretation of the dual problem when the cost are non-negative can be expressed as follows. Let us introduce the subset of $(\mathcal{C}^b(\mathcal{X})\times\mathcal{C}^b(\mathcal{Y}))^N$:
$$  
\mathcal{G}_{\mathbf{c}}^N := \left\{ (f_k,g_k)_{k=1}^N ~\mathrm{  s.t.  }~
     \forall k,~ f_k\oplus g_k\leq c_k\right\}
     $$

Let us now show the  following reformulation of the problem. See Appendix~\ref{prv:dual-reformulation} for the proof.
\begin{prop}
\label{prop:dual-reformulation}
Under the same assumptions of Proposition~\ref{prop:mot-equality},  we have
\begin{align}
\label{eq:dual_interp}
  \MOT_{\mathbf{c}}(\mu,\nu) = & \sup\limits_{  (f_k,g_k)_{k=1}^N
 \in\mathcal{G}^N_{\mathbf{c}}}\inf_{\substack{t\in  \mathbb{R}\\(\mu_k,\nu_k)_{k=1}^N\in\Upsilon_{\mu,\nu}^N }}
 \hspace{0.3em}t\\
 &\nonumber\mathrm{ s.t.}~\forall k, ~ \int f_kd\mu_k+ \int g_kd\nu_k = t 
\end{align} 
\end{prop}

\begin{prv*}
\label{prv:dual-reformulation}
Let us first introduce the following Lemma which guarantees that compacity of  $\Upsilon^N_{\mu,\nu}$ for the weak topology.
\begin{lemma}
\label{lemma:weak-topo-dual-dual}
Let $\mathcal{X}$ and $\mathcal{Y}$ be Polish spaces, and $\mu$ and $\nu$ two probability measures respectively on  $\mathcal{X}$ and $\mathcal{Y}$. Then $\Upsilon^N_{\mu,\nu}$  is sequentially compact for the weak topology induced by $\Vert \gamma \Vert = \max\limits_{i=1,..,N} \Vert \mu_i\Vert_{\TV} + \Vert \nu_i\Vert_{\TV} $. 
\end{lemma}

\begin{prv*}
Let $(\gamma^n)_{n\geq 0}$ a sequence in $\Upsilon^N_{\mu,\nu}$, and let us denote for all $n\geq 0$, $\gamma^n=(\mu^n_i,\nu^n_i)_{i=1}^N$. We first remarks that for all $i\in\{1,...,N\}$ and $n\geq 0$, $\Vert \mu_i^n\Vert_{\TV} \leq 1$ and $\Vert \nu_i^n\Vert_{\TV} \leq 1$ therefore for all $i\in\{1,...,N\}$, $(\mu^n_i)_{n\geq 0}$ and $(\nu^n_i)_{n\geq 0}$  are uniformly bounded. Moreover as $\{\mu\}$ and $\{\nu\}$ are tight, for any $\delta>0$, there exists $K\subset \mathcal{X} $ and $L\subset \mathcal{Y}$ compact such that $\mu(K^c)\leq \delta \text{\quad and\quad }  \nu(L^c)\leq \delta$. Then, we obtain that for any for all $i\in\{1,...,N\}$, $\mu_i^n(K^c)\leq \delta \text{\quad and\quad }  \nu_i^n(L^c)\leq \delta$.
Therefore, for all $i\in\{1,...,N\}$,  $(\mu_i^n)_{n\geq 0}$ and $(\nu_i^n)_{n\geq 0}$ are tight and uniformly bounded and Prokhorov's theorem~\citep[Theorem A.3.15]{dupuis2011weak} guarantees for all $i\in\{1,...,N\}$,  $(\mu_i^n)_{n\geq 0}$ and $(\nu_i^n)_{n\geq 0}$ admit a weakly convergent subsequence. By extracting a common convergent subsequence, we obtain that $(\gamma^n)_{n\geq 0}$ admits a weakly convergent subsequence. By continuity of the projection, the limit also lives in $\Upsilon
^N_{\mu,\nu}$ and the result follows.
\end{prv*}
We can now prove the Proposition. We have that for any $\lambda\in\Delta_N$
\begin{align*}
      &\sup\limits_{(f,g)\in\mathcal{F}_{\mathbf{c}}^{\lambda}} \int_{x\in\mathcal{X}} f(x)d\mu(x)+ \int_{y\in\mathcal{Y}} g(y)d\nu(y)\\
  &\leq \sup\limits_{(f_k,g_k)_{k=1}^N\in  \mathcal{G}^{N}_{\mathbf{c}}} \inf_{(\mu_k,\nu_k)_{i=1}^N\in\Upsilon_{\mu,\nu}^N } \sum_{k=1}^N \lambda_k \left[\int_{x\in\mathcal{X}} f_k(x)d\mu_k(x)+ \int_{y\in\mathcal{Y}} g_k(y)d\nu_k(y)\right]\\
  &\leq \MOT_{\mathbf{c}}(\mu,\nu)
\end{align*}
Then by taking the supremum over $\lambda\in\Delta_N$, and by applying Theorem~\ref{thm:duality-GOT} we obtain that
\begin{align*}
\MOT_{\mathbf{c}}(\mu,\nu)
 =\sup_{\lambda\in \Delta_N} \sup\limits_{(f_k,g_k)_{k=1}^N\in  \mathcal{G}^{N}_{\mathbf{c}}} \inf_{(\mu_k,\nu_k)_{k=1}^N\in\Upsilon_{\mu,\nu}^N } \sum_{k=1}^N \lambda_k \left[\int_{x\in\mathcal{X}} f_k(x)d\mu_k(x)+ \int_{y\in\mathcal{Y}} g_k(y)d\nu_k(y)\right]
\end{align*}
Let $\mathcal{G}^{N}_{\mathbf{c}}$ and $\Upsilon_{\mu,\nu}^N$ be endowed respectively with the uniform norm and the norm defined in Lemma~\ref{lemma:weak-topo-dual-dual}. Note that the objective is linear and continuous with respect to $(\mu_k,\nu_k)_{k=1}^N$ and also $(f_k,g_k)_{k=1}^N$. Moreover the spaces $\mathcal{G}^{N}_{\mathbf{c}}$ and $\Upsilon_{\mu,\nu}^N$ are clearly convex. Finally thanks to Lemma \ref{lemma:weak-topo-dual-dual}, $\Upsilon_{\mu,\nu}^N$ is compact with respect to the weak topology we can apply Sion's theorem \cite{sion1958} and we obtain that
\begin{align*}
\MOT_{\mathbf{c}}(\mu,\nu)
 = \sup\limits_{(f_k,g_k)_{k=1}^N\in  \mathcal{G}^{N}_{\mathbf{c}}} \inf_{(\mu_k,\nu_k)_{k=1}^N\in\Upsilon_{\mu,\nu}^N }\sup_{\lambda\in \Delta_N}  \sum_{k=1}^N \lambda_k \left[\int_{x\in\mathcal{X}} f_k(x)d\mu_k(x)+ \int_{y\in\mathcal{Y}} g_k(y)d\nu_k(y)\right]
\end{align*}
Let us now fix $(f_k,g_k)_{k=1}^N\in  \mathcal{G}^{N}_{\mathbf{c}}$ and $(\mu_k,\nu_k)_{k=1}^N\in\Upsilon_{\mu,\nu}^N$, therefore we have:
\begin{align*}
    &\sup_{\lambda\in \Delta_N}  \sum_{k=1}^N \lambda_k \left[\int_{x\in\mathcal{X}} f_k(x)d\mu_k(x)+ \int_{y\in\mathcal{Y}} g_k(y)d\nu_k(y)\right] \\
    &=\sup_{\lambda}\inf_{t}  t\times\left(1-\sum_{i=1}^N \lambda_i \right) + \sum_{k=1}^N \lambda_k \left[\int_{x\in\mathcal{X}} f_k(x)d\mu_k(x)+ \int_{y\in\mathcal{Y}} g_k(y)d\nu_k(y)\right]\\
     &=\inf_{t} \sup_{\lambda} t + \sum_{k=1}^N \lambda_k \left[\int_{x\in\mathcal{X}} f_k(x)d\mu_k(x)+ \int_{y\in\mathcal{Y}} g_k(y)d\nu_k(y)- t\right]\\
     &= \inf_{t} \left\{t~\mathrm{ s.t.}~\forall k, ~ \int f_kd\mu_k+ \int g_kd\nu_k = t \right\}
\end{align*}
where the inversion is possible as the Slater's conditions are satisfied  and the result follows.
\end{prv*}

\newpage

\subsection{Approximation of the Dudley Metric}

\begin{figure}[h!]
\centering
\includegraphics[width=0.6\linewidth]{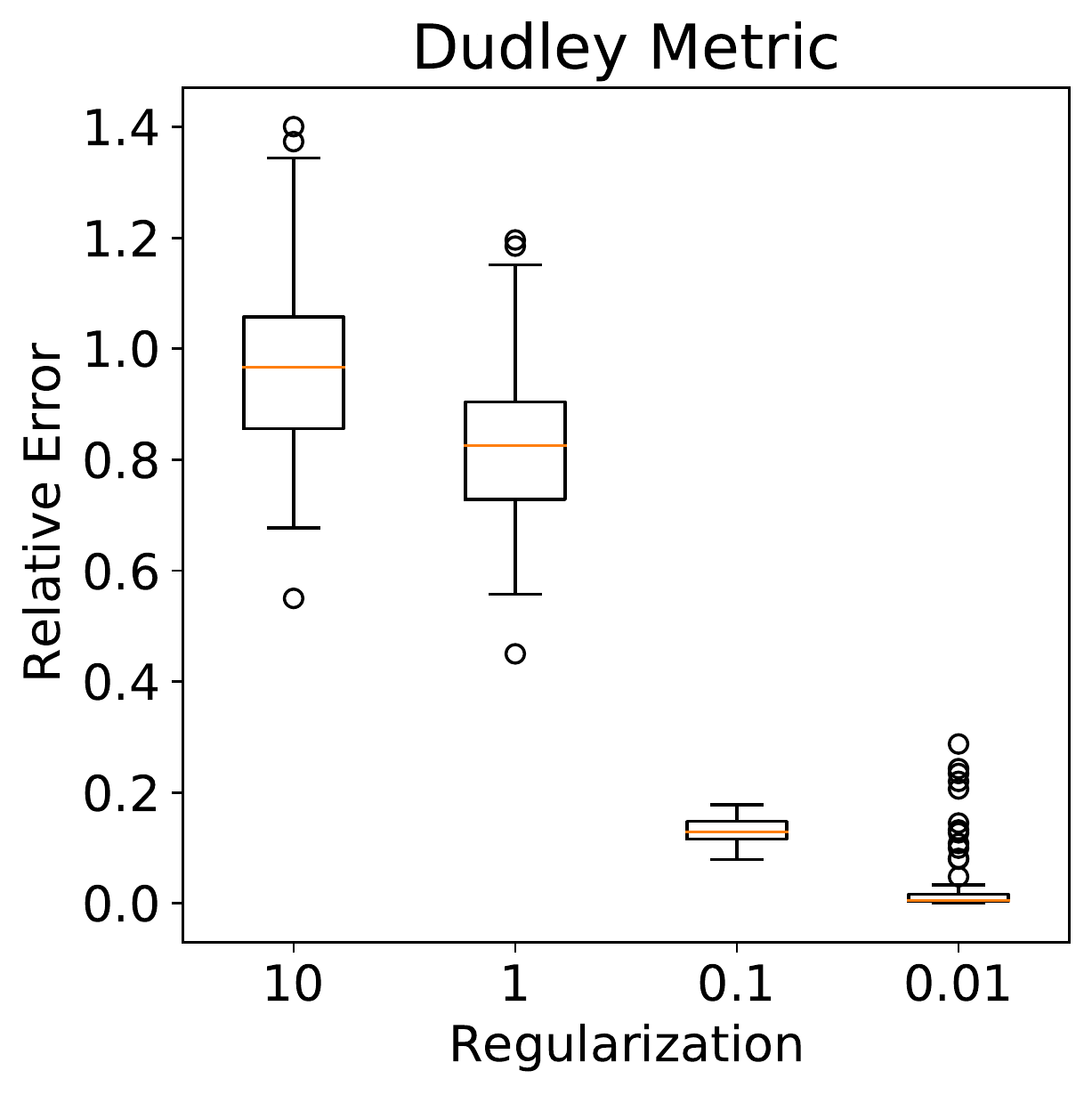}
\caption{In this experiment, we draw 100 samples from two normal distributions and we plot the relative error from ground truth for different regularizations. We consider the case where two costs are involved: $c_1= 2\times\mathbf{1}_{x\neq y}$, and $c_2=d$ where $d$ is the Euclidean distance. This case corresponds exactly to the Dudley metric (see Proposition~\ref{prop:GOT-holder}). We remark that as $\varepsilon\to 0$, the approximation error goes also to $0$.\label{fig:result_acc}}
\end{figure}

Figure~\ref{fig:result_acc} illustrates the convergence of the entropic regularization approximation when $\epsilon\to 0$. To do so  we plot the relative error from the ground truth defined as $\text{RE}:= \frac{ \MOT_{\mathbf{c}}^{\bm{\varepsilon}}-\beta_d}{\beta_d}\ $ for different regularizations where $\beta_d$ is obtained by solving the exact linear program and $\MOT_{\mathbf{c}}^{\bm{\varepsilon}}$ is obtained by our proposed Alg.~\ref{algo:Proj-Sinkhorn}.

\end{document}